\documentclass[twoside,11pt]{article}

\usepackage{blindtext}

\usepackage{jmlr2e}
\usepackage[american]{babel}

\usepackage{natbib} 
\usepackage{mathtools} 
\usepackage{booktabs} 
\usepackage{tikz,ifthen} 
\usepackage{amsmath, eufrak}
\usepackage{algorithm}
\usepackage{algpseudocode}
\usepackage{bm}
\usepackage{cancel}
\usepackage{cleveref}
\usepackage[mathcal]{eucal}
\usepackage{subfig}
\usepackage{xr}

\makeatletter
\let\oldr@@t\r@@t
\def\r@@t#1#2{%
\setbox0=\hbox{$\oldr@@t#1{#2\,}$}\dimen0=\ht0
\advance\dimen0-0.2\ht0
\setbox2=\hbox{\vrule height\ht0 depth -\dimen0}%
{\box0\lower0.4pt\box2}}
\makeatother
\usepackage{letltxmacro}
\LetLtxMacro{\oldsqrt}{\sqrt}
\renewcommand*{\sqrt}[2][\ ]{\oldsqrt[#1]{#2}}

\DeclareMathOperator*{\argmin}{argmin}

\DeclareMathOperator{\arccosh}{arccosh}
\DeclareMathOperator{\arctanh}{arctanh}
\DeclareMathOperator{\dprime}{\prime \prime}

\newcommand{\Real}{\mathbb{R}}

\newcommand{\graph}{\mathbf{G}}
\newcommand{\setofnodes}{\mathbf{X}}

\newcommand{\setofedges}{\mathbf{E}}
\newcommand{\alphabet}{\mathcal{X}}
\newcommand{\RV}[1]{X_{#1}}

\newcommand{\nbhi}{\mathcal{N}(i)}

\newcommand{\degi}{d_{i}}
\newcommand{\degj}{d_{j}}

\newcommand{\msg}[4][]{\ifthenelse{\equal{#4}{}} {\mu^{(#1)}_{#2 #3}(x_{#3})} {\mu^{(#1)}_{#2 #3}(\RV{#3}=#4)}}
\newcommand{\msgShort}[4][]{\ifthenelse{\equal{#4}{}} {\mu^{#1}_{#2 #3}(x_{#3})} {\mu^{#1}_{#2 #3}(#4)}}

\newcommand{\mse}[1][]{\text{MSE}}

\newcommand{\FG}{\mathcal{F}}
\newcommand{\UG}{\mathcal{U}}
\newcommand{\ent}{\mathcal{S}}

\newcommand{\FB}{\mathcal{F}_B}
\newcommand{\HB}{\textbf{H}_B}
\newcommand{\hii}{h_{ii}}
\newcommand{\hij}{h_{ij}}

\newcommand{\ci}{c_{1}}
\newcommand{\cii}{c_{2}}
\newcommand{\qprod}{q_{\pi}}
\newcommand{\Lp}{L^{\prime}}
\newcommand{\Lpp}{L^{\dprime}}
\newcommand{\HBij}{\textbf{H}_B^{(i,j)}}
\newcommand{\HBijsub}{\textbf{H}_{2\times2}^{(i,j)}}
\newcommand{\HBijsubdet}{\det \textbf{H}_{2\times2}^{(i,j)}}
\newcommand{\betacrit}{\beta^{\ast}}

\newcommand{\UB}{\mathcal{U}_{{B}}}
\newcommand{\SB}{\mathcal{S}_{{B}}}

\newcommand{\polytopeLocal}{\mathbb{L}}

\newcommand{\Bbox}{\mathbb{B}}
\newcommand{\qvec}{\bm{q}}
\newcommand{\qi}{q_i}
\newcommand{\qj}{q_j}

\newcommand{\thi}{\theta_i}

\newcommand{\Jij}{J_{ij}}
\newcommand{\Qij}{Q_{ij}}
\newcommand{\aij}{\alpha_{ij}}
\newcommand{\aijp}{\alpha_{ij}^{+}}
\newcommand{\aijm}{\alpha_{ij}^{-}}
\newcommand{\xij}{\xi_{ij}}
\newcommand{\xijopt}{\xi_{ij}^{\ast}}
\newcommand{\xijp}{\xi_{ij}^{+}}
\newcommand{\xijm}{\xi_{ij}^{-}}
\newcommand{\Tij}{T_{ij}}
\newcommand{\Tijp}{T_{ij}^{+}}
\newcommand{\Tijm}{T_{ij}^{-}}

\newcommand{\Ri}{R_i}
\newcommand{\ri}{r_i}
\newcommand{\rijp}{r_{ij}^{+}}
\newcommand{\rijm}{r_{ij}^{-}}
\newcommand{\Psii}{\Psi_i}

\newtheorem{thm}{Theorem}
\newtheorem{Lemma}[thm]{Lemma}

\newtheorem{cor}[thm]{Corollary}

\newtheorem{propos}[thm]{Proposition}
\newtheorem{conj}[thm]{Conjecture}

\usepackage{lastpage}

\ShortHeadings{On the Convexity and Reliability of the Bethe Free Energy Approximation}{Leisenberger, Knoll, and Pernkopf}
\firstpageno{1}

\begin{document}

\title{On the Convexity and Reliability of \\ the Bethe Free Energy Approximation} 

\author{\name Harald Leisenberger \email harald.leisenberger@tugraz.at \\
       \addr Signal Processing and Speech Communication Laboratory\\
       Graz University of Technology\\
       Graz, Austria
       \AND
       \name Christian Knoll \email christian.knoll@tugraz.at \\
       \addr Signal Processing and Speech Communication Laboratory\\
       Graz University of Technology\\
       Graz, Austria
       \AND
       \name Franz Pernkopf \email pernkopf@tugraz.at \\
       \addr Signal Processing and Speech Communication Laboratory\\
       Graz University of Technology\\
       Graz, Austria}

\editor{}

\maketitle

\vspace{-0.7cm}

\begin{abstract}
The Bethe free energy approximation provides an effective way for relaxing NP-hard problems of probabilistic inference. However, its accuracy depends on the model parameters and particularly degrades if a phase transition in the model occurs. In this work, we ana\-lyze when the Bethe approximation is reliable and how this can be verified. We argue and show by experiment that it is mostly accurate if it is convex on a submanifold of its domain, the 'Bethe box'. For verifying its convexity, we derive two sufficient conditions that are based on the definiteness properties of the Bethe Hessian matrix: the first uses the concept of diagonal dominance, and the second decomposes the Bethe Hessian matrix into a sum of sparse matrices and characterizes the definiteness properties of the individual matrices in that sum.
These theoretical results provide a simple way to estimate the critical phase transition temperature of a model.
As a practical contribution we propose \texttt{BETHE-MIN}, a projected quasi-Newton method to efficiently find a minimum of the Bethe free energy. 
\end{abstract}


\section{Introduction} \label{sec:introduction}

Statistical models from quantum mechanics have experienced an increasing popularity in real-world applications~\citep{mackay2003information,zdeborova2016statistical}. Many of the practical challenges can be modeled by a graph and require the solution of fundamental probabilistic problems, such as the computation of the partition function or marginal probabilities. However, an exact solution to these problems is only available for restricted model classes; e.g., the one- and two-dimensional Ising model~\citep{ising1925ising,onsager1944crystal} or small systems~\citep{lauritzen1988junctiontree}. Generally, one needs to apply approximation methods whose specific choice is often a compromise between efficiency and accuracy. \\

A large class of approximation algorithms relies on variational inference and the Gibbs variational principle~\citep{mezard2009information,wainwright2008exponential}. First, probabilistic inference is formulated in terms of an optimization problem with respect to the so-called variational Gibbs free energy; then, to relax the problem complexity, one optimizes a simpler objective that approximates the solutions to the inference problem. This framework has its origin in quantum mechanics and mean-field theory~\citep{parisi1988field}, and has been successfully adopted to computer science~\citep{jordan1999variational,yedidia2005constructing}. \\

One important aspect is the choice of the auxilary objective to be optimized. On the one hand, it should be simple enough to entail a true relaxation of the NP-hard inference problem. On the other hand, a too simple objective might not capture the global behavior of the model well enough. Over time, different types of free energy approximations have been proposed, with the Bethe (or Bethe-Peierls) approximation being one popular representative that often proves to be superior to other constructions in terms of a tradeoff between efficiency and accuracy~\citep{bethe1935superlattices,peierls1936superlattices}. It only considers effects between interacting variables, while ignoring long range correlations at the cost of an information loss; still, it often yields accurate results \footnote{Often, the quality of the Bethe approximation is analyzed in terms of the loopy belief propagation (LBP) algorithm; a so-called message passing algorithm whose relationship to statistical mechanics and the Bethe free energy has been established by~\citet{yedidia2001bethe}.}~\citep{frey1997revolution,murphy1999empirical}. \\

The quality of the Bethe approximation strongly varies between different models and particularly suffers from multiple and strong interactions in the graph. 
From a physisist's perspective this is due to a low model temperature; we may thus interpret a sudden decrease in its accuracy at some critical temperature as a phase transition in the model.
This unwanted behavior of the Bethe approximation is well known and has often been observed in the literature~\citep{meshi2009convexifying,weller2014understanding,leisenberger2022fixing}; however, a precise quantification of its reliability (e.g., in terms of some critical temperature) is an open problem \footnote{Sometimes the success of the Bethe approximation is attributed to the uniqueness of its minimum~\citep{mooij2004validity}. However, we will show that this can be an inaccurate assumption.}. Another prohibitive factor is the practical minimization of the Bethe free energy which -- despite being a relaxation to the Gibbs free energy --  can be difficult. \\

In this work, we analyze the reliability of the Bethe approximation. We show when can expect its provided estimates of the partition function and marginals to be accurate. For that purpose, we distinguish between different 'stages' of the Bethe free energy: convexity, non-convexity but uniqueness of a minimum, and multiple minima. Based on this differentiation, we assess the quality of the Bethe approximation with respect to its stage. \\

The intuition for this split lies in the Gibbs variational principle: While the
Gibbs free energy is convex on its domain, this is not guaranteed for the Bethe free energy. 
We aim to show that the Bethe approximation is accurate, whenever it is convex as well. Moreover, we exa\-mine if it is still accurate if the model temperature decreases, i.e., if the Bethe free energy changes its stage -- from convex to non-convex (but uniqueness of a minimum), and later to multiple minima. 
To perform this analysis, we must determine the stage of the Bethe free energy in a certain state of the model. While there exist methods to prove the uniqueness of a minimum (e.g.,~\citet{mooij2007sufficient}), this is not the case for its convexity. We address this open issue by deriving two sufficient conditions for convexity of the Bethe free energy on an appropriately defined submanifold of its domain, the 'Bethe box'. Both results rely on a profound second-order analysis of the Bethe free energy, by characterizing the definiteness properties of its Hessian. They also provide estimates for a phase transition (i.e., the temperature at which the accuracy of the Bethe approximation begins to degrade). \\

Our first result includes conditions under which the Bethe Hessian is diagonally dominant and hence positive definite (Sec.~\ref{subsec:Bethe_Diagonal_Dominance}). The starting point of this analysis is to reformulate diagonal dominance as a set of optimization problems, each defined on a local neighborhood of the graph. By analyzing these problems, we reduce the question of diagonal dominance of the Bethe Hessian to the positivity of a certain set of one-dimensional polynomials. \\

Our second result decomposes the Bethe Hessian into a sum of simpler matrices and cha\-racterizes the definiteness properties of the individual matrices in that sum (Sec.~\ref{subsec:Bethe_Hessian_Decomposition}). We consider a natural decomposition into 'edge-specific Hessians' that represent the curvature of the Bethe free energy on individual edges and their end nodes. This specific choice induces a closed-form charac\-terization of positive definiteness that is based on their determinant. \\

We conclude the theoretical part of this work by comparing our two conditions to similar results from the literature (Sec.~\ref{subsec:Theoretical_Discussion}). Specifically, we consider a complete graph with a perfectly symmetric potential configuration; a well understood model whose simplicity facilitates such a direct comparison. On this model, our second condition is exact in the sense that it correctly predicts the critical temperature at which a phase transition occurs. \\

After having established the theoretical framework, we perform an experimental analysis in which we evaluate the performance of the Bethe approximation on various graphical models. Specifically, we consider spin glasses (also known as general or mixed models) on graphs that either have a lattice or a random structure. For minimizing the Bethe free energy, we propose a 'projected quasi-Newton' algorithm (named \texttt{BETHE-MIN}, Sec.~\ref{sec:Bethe_experiment}). \\

Our experiments show that the Bethe approximation is mostly accurate if it is convex. If the temperature decreases, the Bethe free energy changes its state while it also becomes much less accurate at some point. We demonstrate that the estimated temperature at which the Bethe free energy becomes non-convex often provides a good estimate of this phase transition. These experiments help to assess, whether the Bethe approximation is reliable or not. \\

This paper is structured as follows: Sec.~\ref{sec:background} provides all relevant notations and background. Sec.~\ref{sec:Bethe_theory} includes the theoretical part of this work in which introduce all details on the Bethe free energy and discuss its characteristics that are known from the literature. Afterwards, we derive our two sufficient conditions for convexity of the Bethe free energy, before we discuss their properties on the symmetric model. Sec.~\ref{sec:Bethe_experiment} includes the practical part of this work. We first explain our algorithm \texttt{BETHE-MIN} for minimizing the Bethe free energy, which we then use in our experiments to analyze the quality of the Bethe approximation with respect to its stage. In Sec.~\ref{sec:conclusion} we conclude our work.

\section{Background} \label{sec:background}

In this section, we introduce the physical and mathematical concepts
and notation
that are relevant for this work.
This includes matrix algebra and convex functions (Sec.~\ref{subsec:mathematical}), models on graphs and the Boltzmann distribution (Sec.~\ref{subsec:modelspec}), and the Gibbs and Bethe variational principle for probabilistic inference (Sec.~\ref{subsec:variational}).
We also give an overview of the literature related to our specific research questions (Sec.~\ref{subsec:related}).

\subsection{Mathematical Preliminaries} \label{subsec:mathematical}

A real matrix $\textbf{M} = (m_{ij})_{i,j = 1,\dots,n}$ of size $n \times n$ is positive definite if $\bm{y}^T \textbf{M} \bm{y} > 0$ for all vectors $\bm{y}$ in $\Real^{n}$ whenever $\bm{y} \neq \bm{0}$, and positive semidefinite if $\bm{y}^T \textbf{M} \bm{y} \geq 0$ for all $\bm{y}$ in $\Real^{n}$.
Equivalently, $\textbf{M}$ is positive definite if all its leading principal minors \footnote{A leading principal minor is the determinant of a $(k \times k)$ - submatrix of $\textbf{M}$, consisting of the first $k$ rows and columns of $\textbf{M}$ (for $k$ in $\{1,\dots,n\}$).} are positive (Sylvester's Criterion).
The sum of positive definite matrices is positive definite. 
Furthermore, $\textbf{M}$ is diagonally dominant if, for each row $i$, the magnitude of the corresponding diagonal entry is larger than the sum of the magnitudes of all other entries in that row, i.e., 
\begin{align} \label{eq:diagonal_dominance}
 |m_{ii}| > \sum\limits_{j \neq i} |m_{ij}| .
\end{align}
A diagonally dominant matrix with positive diagonal entries is positive defi\-nite (e.g.,~\citep{golubloan2013matrix}). \\

Let $f$ be a real-valued and twice continuously differentiable function defined on an open, convex set $\mathcal{D} \subseteq \Real^n$.
Let $\nabla f(\bm{p}) = \big(\frac{\partial f}{\partial p_1}(\bm{p}), \dots, \frac{\partial f}{\partial p_n}(\bm{p})\big)^T $ be the gradient of $f$ and let $\textbf{H}_f(\bm{p}) \coloneqq \nabla^2 f(\bm{p}) = \big(\frac{\partial^2 f}{\partial p_i p_j}(\bm{p})\big)_{i,j = 1,\dots,n}$ be the Hessian matrix of $f$, each evaluated in some point $\bm{p}$ of $\mathcal{D}$.
If $\nabla f(\bm{p}) = \bm{0}$, then $\bm{p}$ is a stationary point of $f$.
A stationary point $\bm{p}$ is a local minimum (resp., maximum) of $f$ if $\textbf{H}_f(\bm{p})$ (resp., $- \textbf{H}_f(\bm{p})$ ) is positive definite, and it is a saddle point if neither $\textbf{H}_f(\bm{p})$ nor $- \textbf{H}_f(\bm{p})$ are positive definite.
We define $f$ to be strictly convex on $\mathcal{D}$ if $\textbf{H}_f$ is positive definite \footnote{This criterion is often easier to work with than the more common yet equivalent definition of strict convexity: $f$ is strictly convex on $\mathcal{D}$ if, for any pair of distinct points $\bm{p_1},\bm{p_2}$ in $\mathcal{D}$, the straight line that connects $f(\bm{p_1})$ and $f(\bm{p_2})$ lies above the curve of $f$ between $\bm{p_1}$ and $\bm{p_2}$, i.e., $\gamma f(\bm{p_1}) + (1-\gamma) f(\bm{p_2}) > f(\gamma f(\bm{p_1}) + (1-\gamma) f(\bm{p_2}))$ for $0 < \gamma < 1$.} for all $\bm{p}$ in $\mathcal{D}$.
In that case, $f$ has at most one local (and thus global) minimum in $\mathcal{D}$ (e.g.,~\citet{rockafellar1970convex}).

\subsection{Model Specification} \label{subsec:modelspec}

We consider a finite system of $N$ variables $X_1, \dots, X_N$ with an identical configuration space $\alphabet_i = \alphabet$ describing the possible states of individual variables.
A configuration $\bm{x}=(x_1, \dots, x_N)$ in the product space $\alphabet^N = \alphabet \times \dots \times \alphabet$ thus represents one possible state of the entire system.
A real-valued energy function $E(\bm{x})$ characterizes the behavior of the system by assigning each system state to its energy.
Then the Boltzmann distribution is defined as
\begin{align} \label{eq:Boltzmann}
 p_{\beta}(\bm{x}) = \frac{1}{Z(\beta)} e^{-\beta E(\bm{x})},
\end{align}
where $\beta = \frac{1}{T}$ is the inverse temperature \footnote{Usually the temperature is scaled by $k_B$, Boltzmann's constant. Without loss of generality we set $k_B =1$, hence measure temperature in units of energy.} of the system and $Z(\beta)$ is the (temperature dependent) normalization constant, known as partition function.
Hence, $p_{\beta}(\bm{x})$ gives the probability that the system is found in state $\bm{x}$.
We focus on an important class of models, satisfying the following conditions:
\begin{itemize}
 \item Variables are described in terms of their spin which takes one of two states (an \emph{Ising spin}).
 The configuration space is thus $\alphabet = \{+1,-1 \}$.
 \item Variables interact in pairs only.
 Specifically, we distinguish between \emph{ferromagnetic} (the spins are aligned) and \emph{antiferromagnetic} (the spins are anti-aligned) interactions.
 \item An external (not necessarily uniform) magnetic field $\thi$ additionally influences the spin of variables.
 \item A graph $\graph=(\setofnodes,\setofedges)$ models the structure of the system.
 Its node set \footnote{With a slight abuse of notation, we usually identify a variable $X_i$ with its representing node $i$ in the graph instead of making an explicit distinction between them.} $\setofnodes=\{X_1, \dots, X_N\}$ represents the variables and its edge set $\setofedges$ represents all interacting pairs of variables.
 Depending on the kind of interaction, we call edges ferromagnetic or antiferromagnetic.
 Furthermore, let $\nbhi$ be the neighborhood of $X_i$ in the graph (i.e., the set of all variables interacting with $X_i$) and $d_i = |\nbhi|$ be the degree of $X_i$.
 \end{itemize}
 Under these assumptions, the energy function $E(\bm{x})$ decomposes into a sum of pairwise and single terms:
 \begin{align} \label{eq:energy}
 E(\bm{x}) = \sum\limits_{(i,j) \in \setofedges} \! \!  E_{ij}(x_i,x_j) + \sum\limits_{i \in \setofnodes} E_{i}(x_i) = - \! \! \sum\limits_{(i,j) \in \setofedges} \! \! \Jij x_i x_j- \sum\limits_{i \in \setofnodes} \thi x_i.
\end{align}
The \emph{couplings} $\Jij$ are positive if $X_i$ and $X_j$ share a ferromagnetic edge, and negative if they share an antiferromagnetic edge.
The \emph{field} $\thi$ biases the spin of variable $X_i$ towards $+1$ if $\thi >0$, and towards $-1$ if $\thi<0$. 
A model with purely positive (negative) couplings is ferromagnetic (antiferromagnetic); a model with both positive and negative couplings is a \emph{spin glass}. A spin glass that is defined on a graph which contains cycles with an odd number of antiferromagnetic edges (implying that the local constraints on the involved spin pairs cannot be satisfied simultanously) is called \emph{frustrated}. \\

Two fundamentally important problems of probabilistic inference \footnote{Note that these problems are closely connected, since marginals can be written as a ratio between a 'partial' and the 'true' partition function~\citep{weller2014b}.} are:
\begin{enumerate}
	\item[\textbf{(P1)}] The computation of the partition function:
	\begin{align} \label{eq:partition_function}
		Z(\beta) = \sum\limits_{\bm{x} \in \alphabet^N} e^{-\beta E(\bm{x})}
	\end{align}
	\item[\textbf{(P2)}] The computation of marginal probabilities of the Boltzmann distribution, primarily of single variables and pairs of variables:
	\begin{align}
		p_i(x_i) = & \sum\limits_{\bm{x}^{\prime} \in \alphabet^N: x_i^{\prime} = x_i} p_{\beta}(\bm{x}^{\prime}) \label{eq:singleton_marginals} \\
		p_{ij}(x_i,x_j) =  & \sum\limits_{\bm{x}^{\prime} \in \alphabet^N: x_i^{\prime} = x_i, x_j^{\prime} = x_j} p_{\beta}(\bm{x}^{\prime}) \label{eq:pairwise_marginals}
	\end{align}
\end{enumerate}

These problems can be solved efficiently on tree-structured (i.e., acyclic) graphs by using dynamical programming~\citep{gallager1962LDPC,pearl1988reasoning}, but are generally NP-hard on \emph{loopy} (i.e., cyclic) graphs~\citep{valiant1979permanent,cooper1990computational}. 
In that case one needs to approximate the solution. One large class of appro\-ximation algorithms is based on variational inference. 
This approach converts the inference problem into an (also intractable) optimization problem, and afterwards optimizes an auxiliary objective that should be close to the original one. 
The solution to the auxiliary problem is then used as an estimate of the solution to the inference problem~\citep{jordan1999variational,wainwright2008exponential}. 
In Sec.~\ref{subsec:variational}  e present further details on the (Gibbs) variational principle.

\subsection{Gibbs Variational Principle and Bethe Approximation} \label{subsec:variational}

We can address the problems \textbf{(P1)} and \textbf{(P2)} stated in Sec.~\ref{subsec:modelspec} by reformulating them in terms of the Gibbs variational principle~\citep{mezard2009information,wainwright2008exponential}. 
The idea is to interpret the partition function and marginals as minima of the variational Gibbs free energy, which is then -- due to its intractability -- approximated by an 'auxiliary free energy' that is easier to optimize. 
Examples are the Bethe free energy~\citep{bethe1935superlattices,peierls1936superlattices}, the tree-reweighted free ener\-gies~\citep{wainwright2005logpartition}, and the Kikuchi free energies~\citep{kikuchi1951cooperative,pelizzola2005cluster,yedidia2005constructing}, each being defined over different clusters of variables. 
Their minima can be obtained by numerical optimization and serve as estimates of the solutions to \textbf{(P1)} and \textbf{(P2)}. \\

Instead of directly minimizing an auxiliary free energy, one can alternatively use so-called \emph{message passing algorithms}.
These are iterative procedures on the  graph whose fixed points are identical to minima of a certain kind of free energy approximation \footnote{More precisely, each class of message passing algorithm corresponds to a specific type of variational free energy approximation and vice versa.}.
The best-known such relationship is the duality between the Bethe approximation and the \emph{loopy belief propagation} (LBP) algorithm~\citep{yedidia2001bethe}.
We do not primarily deal with LBP in this work, but draw conclusions about its behavior by analyzing the Bethe free energy (Sec.~\ref{sec:Bethe_theory}). \\

Let us introduce the Gibbs variational principle and Bethe free energy approximation: Let $q(\bm{x})$ be any probability distribution over the product space $\mathcal{X}^N$. Then we define the \emph{variational Gibbs free energy} as the functional
\begin{align} \label{eq:Gibbs_free_energy}
 \FG(q) = \UG(q) - \frac{1}{\beta} \ent(q)
\end{align}
with the average energy (or expected energy)
\begin{align} \label{eq:Gibbs_average_energy}
 \UG(q) = \mathbb{E}_q(E(\bm{x})) = \sum\limits_{\bm{x} \in \mathcal{X}^N} q(\bm{x}) E(\bm{x})
\end{align}
and the entropy
\begin{align} \label{eq:Gibbs_entropy}
 \ent(q) = -\sum\limits_{\bm{x} \in \mathcal{X}^N} q(\bm{x}) \log q(\bm{x}).
\end{align}
Splitting the Gibbs free energy into the Kullback-Leibler divergence \footnote{The Kullback-Leibler divergence~\citep{cover1991information} measures the 'distance' between two probability distributions $p_1$ and $p_2$ over the same configuration space $\mathcal{Y}$ and is -- in the discrete case -- defined as $D_{KL}(p_1 || p_2) = - \sum\limits_{\bm{y} \in \mathcal{Y}} p_1(\bm{y}) \log \Big(\frac{p_2(\bm{y})}{p_1(\bm{y})} \Big)$.} between $q$ and the Boltzmann distribution $p_{\beta}$, and the partition function $Z(\beta)$ as
\begin{align} \label{Gibbs_KL}
 \FG(q) = \frac{1}{\beta} D(q || p_{\beta}) - \frac{1}{\beta} \log Z(\beta)
\end{align} 
shows that $\FG(q)$ is convex and has a unique minimum if $q$ is chosen as $p_{\beta}$ with the functional value $\FG(p_{\beta}) = - \frac{1}{\beta} \log Z(\beta)$ (known as Helmholtz free energy). \\

While minimizing $\FG$ over all distributions $q$ would leave us with an even increased complexity, the idea is to exploit the variational framework and replace $\FG$ by a simpler objective.
One sophisticated approach is to introduce the \emph{Bethe free energy} that makes two approximations: First, it relaxes the space of feasible distributions $q$ to a larger space of 'pseudo-distributions' $\tilde{p}$ that 
must only satisfy local instead of global probability constraints.
More precisely, $\tilde{p}$ induces pairwise pseudo-marginals $\tilde{p}_{i j}$ and singleton pseudo-marginals $\tilde{p}_{i}$ (all normalized) such that, for all pairs $(i,j) \in \setofedges$, summing $\tilde{p}_{i j}$ over all states of variable $j$ (or $i$) yields $\tilde{p}_{i}$ (or $\tilde{p}_{j}$); on the other hand, global constraints of the form $\sum\limits_{\bm{x}^{\prime} \in \alphabet^N: x_i^{\prime} = x_i, x_j^{\prime} = x_j} \tilde{p}(x_1^{\prime},\dots,x_N^{\prime}) = \tilde{p}_{i j}(x_i,x_j)$ and $\sum\limits_{\bm{x} \in \alphabet^N} \tilde{p}(\bm{x}) = 1 $ need not be satisfied.
Formally, we describe the set $\polytopeLocal$ of all pseudo-distributions -- sometimes called the \emph{local polytope} \footnote{Note that $\polytopeLocal$ is a convex set that is bounded by $\mathcal{O}(|\setofnodes| + |\setofedges|)$ linear constraints~\citep{wainwright2008exponential}.} -- as
\begin{align} \label{eq:local_polytope}
  \begin{split}
  \polytopeLocal = \{ \, \tilde{p}: \alphabet^N \rightarrow \mathbb{R}_{>0} \, \, | & \, \, \,  \sum\limits_{x_j \in \alphabet} \tilde{p}_{ij}(x_i,x_j) = \tilde{p}_i(x_i), \\ & \sum\limits_{x_i,x_j \in \alphabet} \tilde{p}_{ij}(x_i,x_j) = 1, \sum\limits_{x_i \in \alphabet} \tilde{p}_i(x_i) = 1, \\ & \, \, \text{for all} \, \, \, (i,j) \in \setofedges \, \, \, \text{and} \, \, \, i \in \setofnodes \}.
  \end{split}
\end{align}
Second, the Bethe free energy approximates the entropy $\ent(q)$ by a simpler expression, the \emph{Bethe entropy}, that only takes pairwise and local statistics into account.
For $\tilde{p} \in \polytopeLocal$, we define the variational Bethe free energy as
\begin{align} \label{eq:Bethe_free_energy}
 \FB(\tilde{p}) = \UB(\tilde{p}) - \frac{1}{\beta} \SB(\tilde{p})
\end{align}
with the \emph{Bethe average energy}
\begin{align} \label{eq:Bethe_average_energy}
 \begin{split}
 \UB(\tilde{p}) = & \sum_{(i,j) \in \setofedges} \sum_{x_i, x_j \in \alphabet} \tilde{p}_{ij}(x_i,x_j) \, E_{ij}(x_i,x_j) \, + \, \sum_{i \in \setofnodes} \sum\limits_{x_i \in \alphabet} \, \tilde{p}_i(x_i) \, E_i(x_i)
 \end{split}
\end{align}
and the \emph{Bethe entropy}
\begin{align} \label{eq:Bethe_entropy}
 \begin{split}
 \SB(\tilde{p}) = & \sum_{(i,j) \in \setofedges} \overbrace{\Big( -\sum_{x_i, x_j \in \alphabet} \tilde{p}_{ij}(x_i,x_j) \log \tilde{p}_{ij}(x_i,x_j) \Big)}^{\mathcal{S}_{ij}(\tilde{p})} \\ & - \sum_{i \in \setofnodes} \,  (d_i - 1) \underbrace{\Big( - \sum_{x_i \in \alphabet} \tilde{p}_i(x_i) \log \tilde{p}_i(x_i) \Big)}_{\mathcal{S}_i(\tilde{p})}.
 \end{split}
\end{align}
The Bethe entropy $\SB$ uses the sum over all pairwise entropies $\mathcal{S}_{ij}$ to approximate $\ent(q)$. This simplified computation results in an overcounting of entropy contributions from individual nodes, which is compensated by reducing the sum by a $(d_i - 1)$-multiple of the local entropies $\mathcal{S}_i$.
Note that $\FB$ depends on $\tilde{p}$ only via the singleton and pairwise pseudo-marginals $\tilde{p}_i, \tilde{p}_{ij}$ (for each node and edge). This allows us, with a slight abuse of notation, to identify any element $\tilde{p} \in \polytopeLocal$ with the collection of all its associated pseudo-marginals $\{\tilde{p}_i, \tilde{p}_{ij}\}$. \\

By minimizing $\FB$ over $\polytopeLocal$, one usually tries to approximate the partition function $Z(\beta)$ and/or the marginals $p_i, p_{ij}$ induced by $p_{\beta}$ according to
\begin{align}
	\min\limits_{\tilde{p}_i, \tilde{p}_{ij} \, \in \, \polytopeLocal} \color{black} \FB \, \, & \approx \, \, - \frac{1}{\beta} \log Z(\beta) \label{eq:Bethe_approx_partition} \\ 
	\, \,  \argmin\limits_{\tilde{p}_i, \tilde{p}_{ij} \, \in \, \polytopeLocal} \color{black} \FB \, \, & \approx \, \, \{ \, p_i,p_{ij} \, \, | \, \, (i,j) \in \setofedges \, \, \, \text{and} \, \, \, i \in \setofnodes \}. \label{eq:Bethe_approx_marginals}
\end{align}

The expressions on the left-hand side are well-defined, because $\FB$ is bounded from below and its derivatives tend to infinity as $\FB$ approaches the boundary of $\polytopeLocal$; i.e., $\FB$ attains its global minimum in an interior point of $\polytopeLocal$~\citep{yedidia2005constructing,weller2013bethebounds}. \\

The quality of the approximations made in~\eqref{eq:Bethe_approx_partition} and~\eqref{eq:Bethe_approx_marginals} degrades with a high connectivity and strong correlations (or equivalently, a low temperature) in the graph~\citep{meshi2009convexifying, weller2014understanding,leisenberger2022fixing}. This can be explained by the shape of the Bethe free energy: Whenever the graph contains at least two cycles, $\FB$ is -- in contrast to the Gibbs free energy -- non-convex and may lose the favorable property of having a unique global minimum~\citep{pakzad2005kikuchi,heskes2006convexity,watanabe2009graphzeta}. However, this is not the complete story and we will argue in Sec.~\ref{sec:Bethe_theory} that the Bethe approximation is still accurate if it is convex on a specific submanifold of $\polytopeLocal$.

\subsection{Related Work} \label{subsec:related}

Statistical mechanics deals with infinite systems and their properties in thermodynamic equilibrium~\citep{reif1965fundamentals,huang1987statistical}. On the basis of simplified mathematical models, it explains a system's macroscopic behavior in terms of its microscopic constitutents. The statistical properties of the system are fully described by the partition function and its derivatives; however, their analytical form is only known for a few models, e.g., the one- and two-dimensional Ising model,~\citep{ising1925ising,onsager1944crystal}. Of particular interest is the phenomenon of phase transitions, i.e., sudden changes in the qualitative behavior of the system~\citep{yeomans1992phase}. These are manifested as singularities of the thermodynamic potentials (or the partition function) with respect to the temperature. \\

The theory of Gibbs measures~\citep{georgii1988gibbs} provides an elegant framework to study phase transitions directly in infinite systems, instead of computing the thermodynamic limit of finite systems. A Gibbs measure represents an equilibrium state of an infinite system and we may interpret a phase transition as the non-uniqueness of a Gibbs measure \footnote{Formally, a probability distribution $\mu$ on an infinite system $\mathcal{S}$ is a Gibbs measure if, for all finite subsystems $\Lambda \subset \mathcal{S}$, the conditional probabilities induced by $\mu$ on $\Lambda$ are consistent with the canonical ensemble in $\Lambda$, given that all degrees of freedom outside $\Lambda$ are frozen.}. Conversely, the existence of a unique Gibbs measure predicts an absence of phase transitions in the system. As a consequence, researchers have made effort to derive conditions for uniqueness of Gibbs measures~\citep{dobrushin1968uniqueness, simon1979remark}. \\

In computer science, many applications use probabilistic graphical models (PGMs) to represent a probability distribution~\citep{koller2009graphical}. These are principally identical to models from statistical mechanics, with the major difference that the involved graphs are often relatively small and do not necesarrily exhibit a lattice structure.
In particular, finite systems do not exhibit phase transitions in the proper sense, because the thermodynamic potentials are always analytical functions. \citep{brezin1985phase,binder1987phase}. \\

Researchers in this field strongly benefited from statistical mechanics and found a surprising connection between concepts that seem entirely unrelated: the loopy belief propagation (LBP) algorithm on the one hand, which is an iterative method to approximate marginal probabilities in PGMs; and the Bethe-Peierls approximation on the other hand, which has its origins in the theory of superlattices~\citep{bethe1935superlattices,peierls1936superlattices}. More precisely,~\citet{yedidia2001bethe} have proven that there exists a bijective mapping between fixed points \footnote{LBP iteratively exchanges statistical information in form of 'messages' between neighboring nodes in the graph. If this process converges, then we say that LBP has converged to a fixed point.} of LBP and stationary points of the Bethe free energy. This was later complemented by the result that each stable fixed point of LBP corresponds to a local minimum of the Bethe free energy~\citep{heskes2003stable}.
A successful application of the Bethe approximation (or loopy belief propagation) is defined in terms of its quality towards estimating marginal probabilities and the partition function. Due to the lack of general performance guarantees (Sec.~\ref{subsec:variational}), one requires alternative ways to assess the reliability of LBP and the Bethe approximation. \\

\begin{figure}[]
\centering
\includegraphics[width=0.95\linewidth]{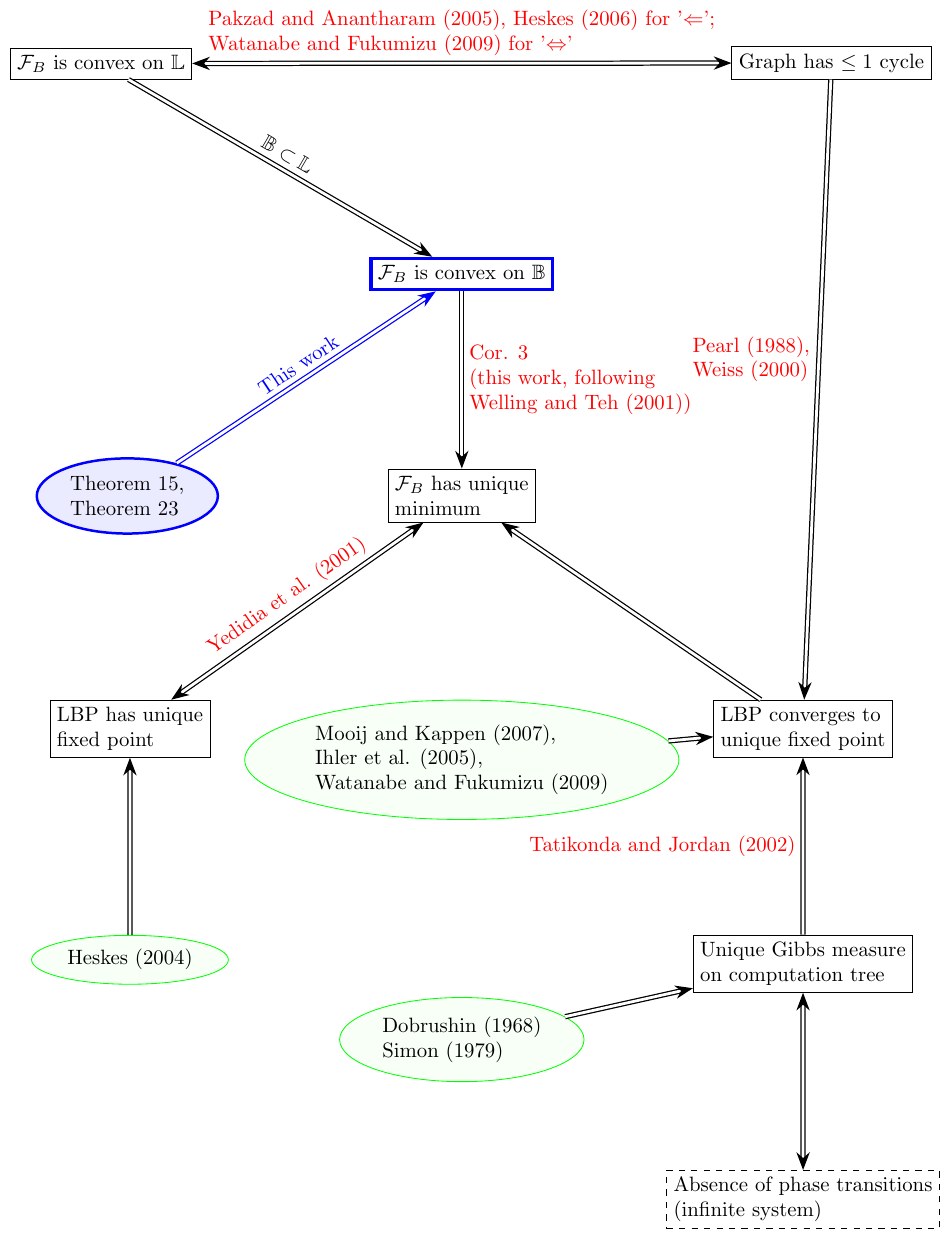}
\caption{Summary of the current state of knowledge about the relationships between the Bethe free energy and loopy belief propagation. The white rectangles contain various properties of the Bethe free energy and LBP. The main results of this work are contained in the blue ellipse and serve as sufficient conditions for the convexity of the Bethe free energy. Sufficient conditions from previous research are included in the green ellipses. Other important results are shown as red label next to the logical arrows between properties.}
\label{fig:Relationships_between_Criteria}
\end{figure}

One line of research analyzes when LBP converges to a unique fixed point. This is often considered as an indicator that the LBP solution (and thus the Bethe approxi\-mation) is accurate. As is known, LBP always converges to a unique fixed point, if the graph is a tree or contains precisely one cycle~\citep{pearl1988reasoning,weiss2000correctness}; for other graphs, the number of fixed points and the convergence behavior depend on the potential configuration (or temperature). Results on this topic include the following:~\citet{tatikonda2002gibbs} have shown that LBP converges to a unique fixed point, if there exists a unique Gibbs measure on the computation tree \footnote{The computation tree is an 'unwrapping' of the graph, starting from some node $i$ and consisting of all non-backtracking paths. E.g., the Cayley tree is the computation tree of a complete graph. The concept of Gibbs measures is well defined on the computation tree, because it consists of infinitely many nodes.}; i.e., they established a direct connection to the theory of Gibbs measures.~\citet{heskes2004uniqueness} formulated a minimax problem on the Bethe free energy whose concavity is sufficient for the uniqueness of an LBP fixed point. \citet{ihler2005message,mooij2007sufficient} consider LBP as a dynamical system and use contraction arguments to prove convergence to a unique fixed point. Further, \citet{watanabe2009graphzeta} use the Ihara zeta function from graph theory to prove uniqueness of an LBP fixed point. Figure~\ref{fig:Relationships_between_Criteria} shows the relations between the different criteria that are described above. \\

Alternatively, one can try to quantify the approximation error induced by the Bethe free energy or LBP. Various results bound its approximation error with respect to the marginal probabilities, either in closed form~\citep{tatikonda2003convergence,tagamase2006bounds} or recursively~\citep{ihler2007accuracy,mooij2008bounds,weller2013bethebounds}. Others use loop series~\citep{chertkov2006loopseries} or graph covers~\citep{vontobel2013graphcovers} to establish a relationship between the Gibbs and the Bethe minimum, which induces bounds on the approxi\-mation error with respect to the partition function.
In ferromagnetic models, the Bethe minimum bounds the Gibbs minimum from above~\citep{willsky2007variational,ruozzi2013bethebound} and can be approximated in polynomial time~\citep{weller2014a}. Unfortunately, these bounds can be arbitrarily loose and are sometimes difficult to compute. \\

Intuitively, as the Bethe free energy is used to approximate a convex functional (the Gibbs free energy, Sec.~\ref{subsec:variational}), one would expect it to provide accurate estimates whenever it is convex as well. However,~\citet{watanabe2009graphzeta} have proven that the Bethe free energy is convex, if and only if the graph is either a cycle or a tree. This might imply that the concept of convexity might not be appropriate for inferring about the quality of the Bethe approximation (because it cannot be convex except for these special graphs anyway).  \\

In this work, we will have a closer look at the convexity of the Bethe free energy. We show how it can be made a useful concept for assessing the quality of the Bethe approximation.

\section{Convexity of the Bethe Free Energy: Theory} \label{sec:Bethe_theory}

This section includes our main theoretical contributions in this work. Particularly, we address the question under which conditions $\FB$ is convex and how this can be verified.
Our theory is built on the results of previous work and exploits the current state of knowledge as a starting point.
In detail, we proceed as follows: First, we summarize the most important properties of the Bethe free energy and specify the problem to be tackled (Sec.~\ref{subsec:Bethe_Properties}).
Next, we perform a second-order analysis of the Bethe free energy and derive two sufficient conditions for its convexity, one of which is based on the criterion of diagonal dominance (Sec.~\ref{subsec:Bethe_Diagonal_Dominance}), and the other on a sum decomposition of the Hessian matrix (Sec.~\ref{subsec:Bethe_Hessian_Decomposition}). Finally, we compare our results to similar statements from the literature with respect to a perfectly symmetric model that facilitates a direct comparison (Sec.~\ref{subsec:Theoretical_Discussion}). Our notation in this section closely follows that of~\citet{welling2001belief,weller2013bethebounds,leisenberger2022fixing}.

\subsection{Properties of the Bethe Free Energy} \label{subsec:Bethe_Properties}

We first make a range of simplifications regarding the dimensionality of the local polytope~\eqref{eq:local_polytope} and the Bethe free energy~\eqref{eq:Bethe_free_energy} for binary variables. Let $X_i, X_j$ be any pair of connected variables and let $\tilde{p}_i, \tilde{p}_j, \tilde{p}_{ij}$ be a set of associated pseudo-marginal vectors \footnote{More precisely, $\tilde{p}_{i} = \begin{pmatrix}\tilde{p}_{i}(X_{i} = +1)\\\tilde{p}_{i}(X_{i} = -1)\end{pmatrix}$ and $\tilde{p}_{ij} = \begin{pmatrix}\tilde{p}_{ij}(X_i = +1,X_j = +1)\\\tilde{p}_{ij}(X_i = +1,X_j = -1)\\\tilde{p}_{ij}(X_i = -1,X_j = +1)\\\tilde{p}_{ij}(X_i = -1,X_j = -1)\end{pmatrix}$.} from the local polytope $\polytopeLocal$ (i.e., with all entries satisfying the constraints in the definition of $\polytopeLocal$). Let us denote the individual pseudo-marginal probabilities for $X_i$ and $X_j$ being in state $+1$ by $q_i \coloneqq \tilde{p}_i(X_i = +1)$ and $q_j \coloneqq \tilde{p}_j(X_j = +1)$ and let us further denote the joint pseudo-marginal probability for $X_i,X_j$ being both in state $+1$ by $\xij \coloneqq \tilde{p}_{ij}(X_i = +1,X_j = +1)$. Then, as $\tilde{p}_i, \tilde{p}_j, \tilde{p}_{ij}$ must satisfy the constraints of $\polytopeLocal$, all five remaining entries of these vectors can already be expressed in terms of $q_i, q_j, \xij$ according to the following table:

\begin{table}[h] 
\centering
\caption{Joint probability table of two binary variables $X_i$ and $X_j$.} \label{tab:prob_table}
\begin{tabular}{ c||c|c||c } 
 $\tilde{p}_{ij}(X_i, X_j)$ & $X_j = +1$ & $X_j = -1$ & \\ \hline \hline
 $X_i = +1$ & $\xi_{ij}$ & $q_i - \xi_{ij}$ & $q_i$ \\ \hline
 $X_i = -1$ & $q_j - \xi_{ij}$ & $1 + \xi_{ij} - q_i - q_j$ & $1-q_i$ \\ \hline \hline
 & $q_j$ & $1 - q_j$ &  \\
\end{tabular}
\end{table}

Additionally, we have constraints on the independent parameters $q_i, q_j, \xij$ of the form

\begin{align}
 0 < & \, \, q_i < 1, \\
 0 < & \, \, q_j < 1, \\
 \max(0,q_i + q_j - 1) < & \, \, \xi_{ij}< \min(q_i,q_j).
\end{align}

This step reparameterizes the local polytope by exploiting probabilistic dependencies between variables and removing a considerable number of redundant parameters. If we collect all node-specific variables $q_i$ in a vector $\bm{q}$ and all edge-specific parameters $\xij$ in a vector $\bm{\xi}$, we can compactly rewrite the local polytope~\eqref{eq:local_polytope} as the set

\begin{align} \label{eq:local_polytope_reparameterized}
 \begin{split}
\hspace{-0.5cm} \polytopeLocal = \{& (\bm{q}; \bm{\xi}) \in \mathbb{R}^{\lvert \setofnodes \rvert + \lvert \setofedges \rvert}: 0 <  q_i  < 1, i \in \setofnodes; \\ & \max(0, q_i + q_j - 1) <  \xi_{ij}  < \min(q_i,q_j), (i,j) \in \setofedges \}.
 \end{split}
\end{align}

\begin{figure}[]
\centering
\hspace{-1cm} \subfloat{\includegraphics[width=0.5\linewidth]{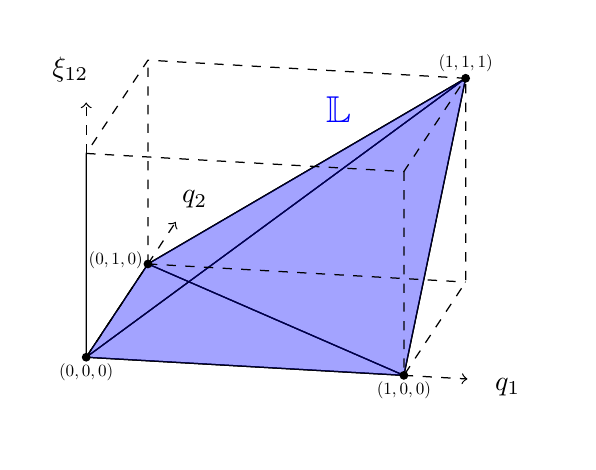}}
\hspace{0.5cm} \subfloat{\includegraphics[width=0.15\linewidth]{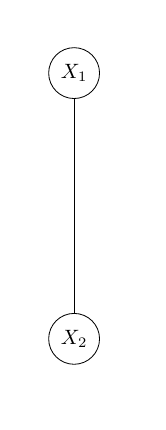}}
\caption{Local polytope $\polytopeLocal$ (left) on a single-edge graph (right).}
\label{fig:Local_Polytope}
\end{figure}

Figure~\ref{fig:Local_Polytope} shows the local polytope
on a graph with two vertices and a single edge. \\

This reparameterization of the local polytope also reparameterizes the Bethe free energy: By substituting the expressions from Table~\ref{tab:prob_table} into the corresponding terms of~\eqref{eq:Bethe_free_energy} -~\eqref{eq:Bethe_entropy}, $\FB$ takes the simpler form
\begin{align} \label{eq:Bethe_reparameterized}
 \begin{split}
 \FB(\bm{q}; \bm{\xi}) = & \overbrace{- \sum_{(i,j) \in \setofedges} \, (1+  2 \; (2 \, \xi_{ij} - q_i - q_j)) \, J_{ij} + \, \, \sum_{i \in \setofnodes} (1 - 2 q_i) \, \theta_i}^{\UB(\bm{q}; \bm{\xi})} \\  & - \frac{1}{\beta} \underbrace{\Big(\sum_{(i,j) \in \setofedges} \mathcal{S}_{ij} \; - \; \sum_{i \in \setofnodes} (d_i - 1) \, \mathcal{S}_{i} \Big)}_{\SB(\bm{q}; \bm{\xi})} ,
 \end{split}
\end{align}
with the pairwise entropies
\begin{align} \label{eq:Bethe_entropy_PW}
 \begin{split}
 \hspace{-0.15cm} \mathcal{S}_{ij} = & -\xi_{ij} \log \xi_{ij} - (1+\xi_{ij}-q_i-q_j) \log (1+\xi_{ij}-q_i-q_j) \\ & - (q_i - \xi_{ij}) \log (q_i - \xi_{ij}) - (q_j - \xi_{ij}) \log (q_j - \xi_{ij})
   \end{split}
\end{align}
and the local entropies
\begin{align} \label{eq:Bethe_entropy_Loc}
 \begin{split}
 \mathcal{S}_{i} = -q_i \log q_i - (1-q_i) \log (1-q_i).
   \end{split}
\end{align}

This and similar reparameterizations of $\polytopeLocal$ and $\FB$ are not new and have been used for diffe\-rent purposes~\citep{welling2001belief,mooij2005bethe}. One result from~\cite{watanabe2009graphzeta} that is particularly relevant for our work is the following:

\begin{propos}[\normalfont Corollary 2 in~\cite{watanabe2009graphzeta}] \label{propos:Watanabe_convexity} {\ \\}
 $\FB$ is convex on $\polytopeLocal$, if and only if the graph $\graph$ is either a tree or contains precisely one cycle.
\end{propos}

In principle, this statement fully characterizes the convexity of $\FB$. As it is only applicable for two types of graphs, convexity might not seem to be a suitable concept for explaining the success (or failure) of the Bethe approximation \footnote{In particular, there are popular applications involving graphs with multiple cycles where the Bethe approxiation (resp., LBP) is highly accurate~\citep{frey1997revolution,yanover2002protein}.}. However, we will show how we can refine Proposition~\ref{propos:Watanabe_convexity} by defining an appropriate subset $\Bbox$ of $\polytopeLocal$, on which $\FB$ can preserve its convexity for general graphs. Moreover, we claim that $\FB$ provides accurate solutions if is convex on that specific subset $\Bbox$ of $\polytopeLocal$. From this, three questions arise:
\begin{enumerate}
 \item [(i)] How can we characterize $\Bbox$ such that $\FB$ can preserve its convexity for general graphs?
 \item [(ii)] How can we determine if $\FB$ is convex on $\Bbox$?
 \item[(iii)] Can we explain the reliability of the Bethe approximation through its convexity on $\Bbox$?
\end{enumerate}
For the rest of this work, we focus on answering these three questions: (i) will be addressed in the remainder of the current subsection; (ii) will be subject of Sec.~\ref{subsec:Bethe_Diagonal_Dominance} and Sec.~\ref{subsec:Bethe_Hessian_Decomposition}; and (iii) will be theoretically discussed in Sec.~\ref{subsec:Theoretical_Discussion} and experimentally evaluated in Sec.~\ref{sec:Bethe_experiment}. \\

First we identify a subset $\Bbox$ of $\polytopeLocal$ on which $\FB$ can preserve its convexity for general graphs. For that purpose, recall that we are primarily interested in minimizing the Bethe free energy to approximate the minimum of the Gibbs free energy (Sec.~\ref{subsec:variational}). If we exclude regions of $\polytopeLocal$ where no minimum of $\FB$ can lie, then we lose no relevant information on its accuracy. Or, in other words, we expect that $\FB$ should primarily be a convex approximation to the Gibbs free energy on a subset $\Bbox$ of $\polytopeLocal$, that includes all minima of $\FB$. Following this idea, we adopt an approach made by~\citet{welling2001belief,shin2012complexity,weller2013bethebounds} -- albeit in their work, to reduce the computational complexity of Bethe minimization: \\

Let $(\bm{q}; \bm{\xi})$ be a point in $\polytopeLocal$ where all partial derivatives of $\FB$ with respect to the edge-specific variables $\xij$ (not necessarily with respect to the node-specific variables $q_i$) are zero.; i.e., the gradient of $\FB$ evaluated in that point is $\nabla \FB (\bm{q}; \bm{\xi}) = \big(\frac{\partial \FB}{\partial q_1}(\bm{q}; \bm{\xi}), \dots, \frac{\partial \FB}{\partial q_N}(\bm{q}; \bm{\xi}) ; 0,\dots,0\big)^T$. This imposes a necessary condition on any $(\bm{q}; \bm{\xi}) \in \polytopeLocal$ for being a stationary point (and thus a minimum) of $\FB$ (Sec.~\ref{subsec:mathematical}). In~\citet{welling2001belief} it has been shown that, if we choose an arbitrary (but fixed) $\bm{q}$, there exists a unique $\bm{\xi}^{\ast}(\qvec)$ -- i.e., depending on the choice of $\qvec$ -- such that $\big(\qvec;\bm{\xi}^{\ast}(\qvec)\big)$ is a candidate for a stationary point of $\FB$; moreover, the explicit form of $\bm{\xi}^{\ast}(\qvec)$ has been computed with its individual components given by
 \begin{align} \label{eq:xi_optimal}
 \begin{split}
   & \xijopt(q_i,q_j) =  \frac{1}{2\alpha_{ij}} \Big( \Qij - \sqrt{\Qij^2 - 4 \alpha_{ij}(1+\alpha_{ij}) q_i q_j  } \, \, \Big),  \\
   & \text{where} \, \,  \quad \aij = e^{4 \beta \Jij} - 1 \quad \, \, \text{and} \quad \, \, \Qij = 1 + \aij(\qi + \qj).
 \end{split}
\end{align}
In other words, for any $\bm{q}$ there exists at most one $\bm{\xi}^{\ast}(\bm{q})$ (defined in~\eqref{eq:xi_optimal}) such that $\big(\bm{q}; \bm{\xi}^{\ast}(\bm{q})\big)$ can be a minimum of $\FB$. Consequently, we address question (i) according to our previously described requirements on $\Bbox$, by defining a suitable subset $\Bbox$ of $\polytopeLocal$ as follows:

\begin{thm}[\normalfont Adopted from~\citet{welling2001belief}] \label{thm:Welling_Bethe_Box} {\ \\}
 Every minimum of $\FB$ is located on an $\lvert \setofnodes \rvert$- dimensional submanifold $\Bbox$ of $\polytopeLocal$, the \textbf{Bethe box}, which is defined as
 \begin{align} \label{eq:Bethe_box}
  \begin{split}
\Bbox \coloneqq \{ & (\bm{q}; \bm{\xi}^{\ast} (\bm{q})) \in \polytopeLocal: \, \, 0 <  q_i  < 1, i \in \setofnodes; \\
  & \, \, \xijopt(q_i,q_j) \, \, \, \text{given by} \, \, \, \eqref{eq:xi_optimal}, \, (i,j) \in \setofedges \}.
  \end{split}
 \end{align}
\end{thm}

\begin{figure}[ht]
\centering
\includegraphics[width=0.78\linewidth]{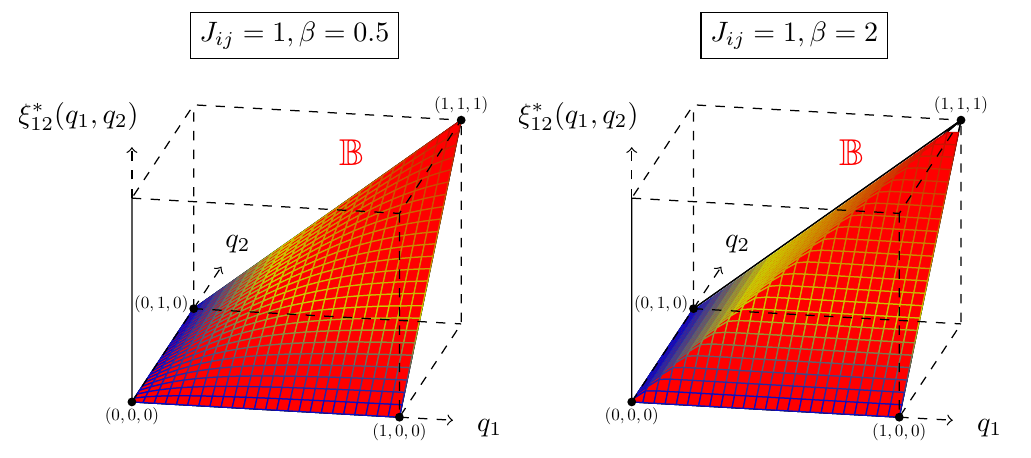}
\includegraphics[width=0.78\linewidth]{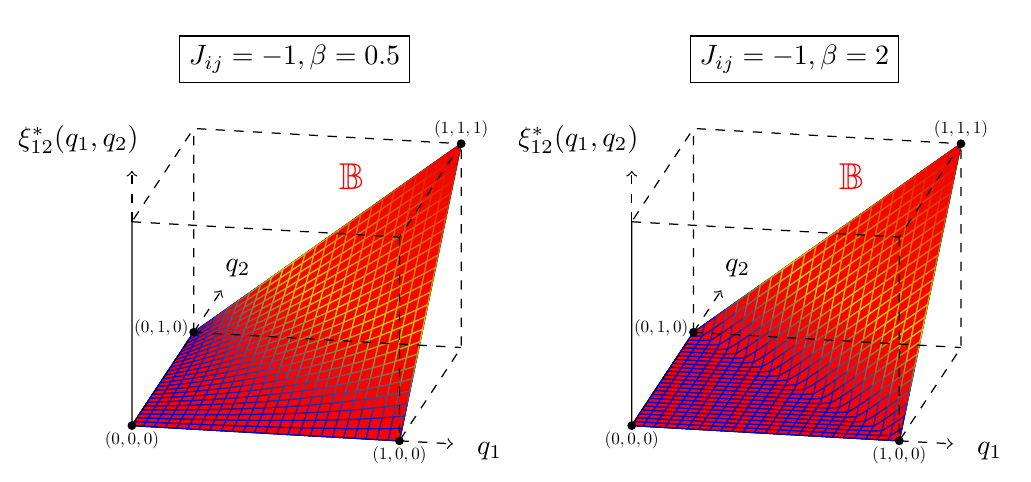}
\caption{Bethe box $\Bbox \subset \polytopeLocal$ for different values of coupling $J_{ij}$ and inverse temperature $\beta$ on a single-edge graph. For a ferromagnetic edge ($J_{ij}> 0$; first row), $\Bbox$ approaches the upper boundary of $\polytopeLocal$ as $\beta$ increases; for an antiferromagnetic edge ($J_{ij}< 0$; second row), $\Bbox$ approaches the lower boundary of $\polytopeLocal$ as $\beta$ increases.}
\label{fig:Bethe_Box}
\end{figure}

Note that $\Bbox$ is parameterized in terms of the $\qi$ only, while the $\xij$ have lost their role as independent parameters but depend on the $\qi$ through the bijective mapping~\eqref{eq:xi_optimal}. This implies that $\Bbox$ is isomorphic to the open unit cube $(0,1)^{\lvert \setofnodes \rvert}$ of dimension $\lvert \setofnodes \rvert$, which explains the naming of $\Bbox$. Figure~\ref{fig:Bethe_Box} shows the Bethe box with respect to a graph on two vertices and a single edge for different values of the associated coupling $J_{ij}$ and inverse temperature $\beta$. \\

\vspace{-0.25cm}

The Bethe box $\Bbox$ satisfies our intuitive requirements on a subset of $\polytopeLocal$, on which $\FB$ can preserve its convexity for general graphs (in fact, we prove this in Sec.~\ref{subsec:Bethe_Diagonal_Dominance} and Sec.~\ref{subsec:Bethe_Hessian_Decomposition}). Also, the definition~\eqref{eq:Bethe_box} of $\Bbox$ once again reduces the number of independent variables $\FB$, as we can define $\FB$ directly on the space of its potential minima, by substituting $\xijopt(q_i,q_j)$ from~\eqref{eq:xi_optimal} into $\xij$ in~\eqref{eq:Bethe_reparameterized}.
 An immediate implication of Theorem~\ref{thm:Welling_Bethe_Box} is:

\begin{cor} \label{cor:Unique_Bethe_Minimum} {\ \\}
 If $\FB$ is strictly convex on $\Bbox$, then it has a unique global minimum in $\polytopeLocal$.
\end{cor}

In Sec.~\ref{subsec:Theoretical_Discussion}, we will have a closer look at the connection between convexity of $\FB$ uniqueness of a minimum. Finally, we require the first-order and second-order derivatives of $\FB$ on $\Bbox$:

\begin{Lemma}[\normalfont (a): \citet{welling2001belief}; (b):
Lemma 9 in~\citet{weller2013bethebounds}] \label{lemma:Bethe_derivatives} {\ \\} \vspace{-0.6cm}
\begin{enumerate}
 \item [(a)] The first-order partial derivatives of $\FB$ on $\Bbox$ are
 \begin{align}
 \begin{split} \label{eq:Bethe_first_derivatives}
  \frac{\partial \FB}{\partial q_{i}} =  -2 \, \theta_i + 2 \sum\limits_{j \in \nbhi} J_{ij}  + \frac{1}{\beta} \log \Big( \frac{(1-q_i)^{\degi-1}} {q_i^{\degi-1}} \prod\limits_{j \in \nbhi} \frac{q_i - \xijopt }{1 + \xijopt - q_i - q_j}  \Big).
 \end{split}
 \end{align}
 \vspace{-0.70cm}
 \item [(b)] The second-order partial derivatives of $\FB$ on $\Bbox$ are
 \begin{align}
  \begin{split} \label{eq:Bethe_second_derivatives}
   \hspace{-0.75cm} & \frac{\partial^2 \FB}{\partial \qi \partial \qj} =
   \begin{dcases}
    \frac{1}{\beta} \, \Big(\! -\frac{\degi -1}{\qi (1-\qi)} + \sum\limits_{j \in \nbhi} \frac{\qj (1-\qj)}{\Tij} \Big) & \quad i = j \\
    \frac{1}{\beta} \, \Big(\frac{\qi \qj - \xijopt}{\Tij} \Big) & \quad j \in \nbhi \\
    0 & \quad \text{otherwise},
   \end{dcases}
   \end{split}
   \\  & \hspace{-0.2cm} \text{where} \quad \Tij(\qi,\qj) \coloneqq \qi \qj (1-\qi) (1-\qj) - (\xijopt(\qi,\qj) - \qi \qj)^2 > 0. \label{eq:Tij_optimal}
 \end{align}
 Also, we have that $\frac{\partial^2 \FB}{\partial \qi^2}$ is always $>0$.
\end{enumerate}
\end{Lemma}

A simple, but remarkable consequence from Lemma~\ref{lemma:Bethe_derivatives} is:

\begin{cor} \label{cor:independence_of_theta} {\ \\}
 The Hessian matrix of $\FB$ is independent from $\theta_i$. In other words, the curvature of the Bethe free energy is not influenced by the local potentials. 
\end{cor}

Next, we deal with question (ii): How can we determine, if $\FB$ is convex on the Bethe box $\Bbox$? To address this problem, we propose two different methodologies -- the first one in Sec.~\ref{subsec:Bethe_Diagonal_Dominance}, the second one in Sec.~\ref{subsec:Bethe_Hessian_Decomposition} -- that both rely on a profound analysis of the Hessian matrix, i.e., the curvature of the Bethe free energy. The basic idea of both approaches is the same: to apply the formal definition of strict convexity (Sec.~\ref{subsec:mathematical}) to the Bethe free energy: Let $\HB(\bm{q}): \Bbox \to \mathbb{R}^{\lvert \setofnodes \rvert \times \lvert \setofnodes \rvert}$ be the Hessian matrix of $\FB$ (the \emph{Bethe Hessian}), considered as a matrix-valued function on the Bethe box \footnote{To simplify the notation, we denote the Bethe Hessian by $\textbf{H}_{B}$ instead of $\textbf{H}_{\FB}$, as formally defined in Sec.~\ref{subsec:mathematical}. Also, we mostly omit the dependency of $\HB$ on $\qvec$, since this is usually clear from the context.}. Then $\FB$ is strictly convex on $\Bbox$, if $\HB(\bm{q})$ is positive definite for all $\qvec \in \Bbox$. Unfortunately, the complex analytical form of $\HB$ (Lemma~\ref{lemma:Bethe_derivatives}, (b)) prevents a direct application of this definition and requires relaxations to the problem.

\subsection{Node-Specific Convexity Criterion: Diagonal Dominance of Bethe Hessian} \label{subsec:Bethe_Diagonal_Dominance}

Our first approach to relax the convexity of $\FB$ uses the concept of diagonal dominance, a sufficient condition for positive definiteness of a matrix (Sec.~\ref{subsec:mathematical}). The main result is Theorem~\ref{thm:polynomial_characterization_of_inf_Ri}, which reduces diagonal dominance of the Bethe Hessian to the positivity of a set of one-dimensional polynomials. All proofs for this section are collected in Appendix~\ref{sec:appendix_A}. \\

Let us compactly write the $(i,j)$-th entry of $\HB$ (i.e., $\frac{\partial^2 \FB}{\partial \qi \partial \qj}$) as $\hij$. Then the diagonal dominance criterion (equation~\eqref{eq:diagonal_dominance}) says that $\HB$ is positive definite in $\qvec \in \Bbox$, if
\begin{align} \label{eq:diagonal_dominance_Bethe_first}
 \hii  - \sum\limits_{j \in \nbhi} |\hij | > 0
\end{align}
for each row of $\HB$, i.e., for all $i \in \setofnodes$. Note that $\hii > 0$, and $\hij = 0$ if neither $i = j$ nor $j \in \nbhi$. If we insert the expressions from~\eqref{eq:Bethe_second_derivatives} into~\eqref{eq:diagonal_dominance_Bethe_first}, this is equivalent to
\begin{align} \label{eq:diagonal_dominance_Bethe_second}
 \frac{1}{\beta} \, \Big(\! -\frac{\degi -1}{\qi (1-\qi)} + \sum\limits_{j \in \nbhi} \frac{\qj (1-\qj)}{\Tij} - \sum\limits_{j \in \nbhi} \, \frac{|\qi \qj - \xijopt|}{\Tij} \Big) > 0,
\end{align}
because $\Tij >  0$ (Lemma~\ref{lemma:Bethe_derivatives}, (b)). To compute the absolute values, we can use the following result:

\begin{Lemma}[\normalfont Lemma 2 in~\citet{weller2013bethebounds}] {\ \\} \label{lemma:Weller_bounds}
If $(i,j)$ is a ferromagnetic edge ($\Jij > 0$), then $\xijopt > \qi \qj$. If $(i,j)$ is an antiferromagnetic edge ($\Jij < 0$), then $\xijopt < \qi \qj$.
\end{Lemma}

If we distinguish between the two types of edges, inequality~\eqref{eq:diagonal_dominance_Bethe_second} becomes
\begin{align} \label{eq:diagonal_dominance_Bethe_third}
 -\frac{\degi -1}{\qi (1-\qi)} + \sum_{\substack{j \in \nbhi \\ \Jij > 0}} \frac{\qj (1-\qj) - \xijopt + \qi \qj}{\Tij} + \sum_{\substack{j \in \nbhi \\ \Jij < 0}} \frac{\qj (1-\qj) - \qi \qj  + \xijopt}{\Tij} > 0.
\end{align}
So far, we have only considered one specific point $\qvec \in \Bbox$; however, to guarantee the convexity of $\FB$ on the entire Bethe box $\Bbox$, inequality~\eqref{eq:diagonal_dominance_Bethe_third} must be valid for \emph{all} $\qvec \in \Bbox$ or, equivalently, for the infimum \footnote{Strictly speaking, the infimum could also be equal to zero, while positivity holds for all interior points of $\Bbox$. However, we will ignore this subtlety as it is does not affect our later results.} over all $\qvec \in \Bbox$. In summary, we arrive at the following problem:

\begin{propos} \label{prop:problem_formulation_diagonal_dominance} {\ \\}
 Let us denote the expression on the left-hand side of inequality~\eqref{eq:diagonal_dominance_Bethe_third} by $\Ri(\qvec)$. Then $\FB$ is convex on the Bethe box $\Bbox$ if, for all nodes $i \in \setofnodes$,
 \begin{align} \label{eq:diagonal_dominance_Bethe_infimum}
  \inf\limits_{\qvec \in \Bbox} \, \, \Ri(\qvec) > 0.
 \end{align}
\end{propos}
In Proposition~\ref{prop:problem_formulation_diagonal_dominance} we have reformulated diagonal dominance of the Bethe Hessian in terms of an optimization problem. The infimum of $\Ri(\qvec)$ measures to what extent the diagonal domi\-nance criterion is satisfied (or violated) in node $i$. As $\Ri$ is a nonlinear function of potentially many variables \footnote{More precisely, $\Ri$ depends on $(\degi + 1)$ variables: on $\qi$ that is associated to node $i$, and additionally on all $\qj$ that are associated to neighboring nodes $j \in \nbhi$.}, its solution (i.e., the verification of inequality~\eqref{eq:diagonal_dominance_Bethe_infimum}) is not trivial. \\

In the remainder of this section, we analyze the left-hand side of the inequality~\eqref{eq:diagonal_dominance_Bethe_infimum} and finally present a condition that is equivalent to~\eqref{eq:diagonal_dominance_Bethe_infimum} but much easier to verify (Theorem~\ref{thm:polynomial_characterization_of_inf_Ri}). Our first observation is that $\Ri(\qvec)$ has the special form
\begin{align} 
	\Ri(\qvec) & = \ri(\qi) + \sum_{\substack{j \in \nbhi \\ \Jij > 0}} \rijp(\qi,\qj) + \sum_{\substack{j \in \nbhi \\ \Jij < 0}} \rijm(\qi,\qj), \quad \text{where} \label{eq:Ri_rewritten} \\
& \ri(\qi) \coloneqq -\frac{\degi -1}{\qi (1-\qi)} \quad \text{depends only on} \, \, \qi, \label{eq:Ri_ri} \\
& \rijp(\qi,\qj) \coloneqq \frac{\qj (1-\qj) - \xijopt + \qi \qj}{\Tij} \quad \text{depends on} \, \, \qi \, \, \text{and} \, \, \qj, \, \,  \text{and} \label{eq:Ri_rijp} \\
& \rijm(\qi,\qj) \coloneqq \frac{\qj (1-\qj) - \qi \qj + \xijopt}{\Tij} \quad \text{depends on} \, \, \qi \, \, \text{and} \, \, \qj. \label{eq:Ri_rijm}
\end{align}

\begin{figure}[ht]
\centering
\includegraphics[width=0.38\linewidth]{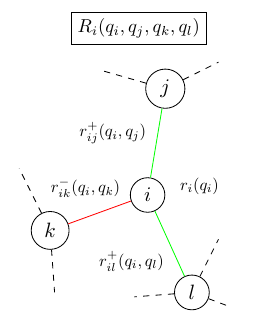}
\caption{Dependencies and components of the function $\Ri(\qvec)$ associated to some node $i$. Ferromagnetic edges are drawn in green and antiferromagnetic  edges are drawn in red.}
\label{fig:Diagonal_Dominance_Criterion}
\end{figure}

Note that, for each ferromagnetic edge we have one term of the form~\eqref{eq:Ri_rijp}, and for each antiferromagnetic edge we have one term of the form~\eqref{eq:Ri_rijm} (with $\qi$ and $\qj$ being the variables associated to the end nodes of a specific edge $(i,j)$).
In particular, $\qi$ (associated to node $i$) is the only variable that appears in each individual sum term, while any other variable $\qj$ (for $j \in \nbhi$) appears in precisely one of the pairwise terms $\rijp(\qi,\qj)$ or $\rijm(\qi,\qj)$ (depending on the type of the edge that connects $i$ and $j$). Figure~\ref{fig:Diagonal_Dominance_Criterion} illustrates this situation.

We can exploit the specific structure~\eqref{eq:Ri_rewritten} of $\Ri(\qvec)$ and decompose the computation of $\inf\limits_{\qvec \in \Bbox} \Ri(\qvec)$ into a collection of simpler sub-problems, each involving only a single variable. Concretely, we apply a two-step procedure: In the first step, we fix $\qi$ and determine
\begin{align} \label{eq:infimum_rij}
 \inf\limits_{\qj \in (0,1)} \, \, \rijp(\qi,\qj) \, \, \quad \, \, \text{and} \quad \, \, \inf\limits_{\qj \in (0,1)} \, \, \rijm(\qi,\qj)
\end{align}
for each of the pairwise terms $\rijp(\qi,\qj), \rijm(\qi,\qj)$ separately (with respect to $\qj$ as a function of $\qi$). In the second step, we fix the solutions to these sub-problems and compute the infimum of the overall function $\Ri(\qvec)$ with respect to $\qi$. Altogether, we determine
\begin{align} \label{eq:sum_of_infima}
 \tilde{R}_i \coloneqq \inf\limits_{\qi \in (0,1)} \, \, \Big[ \, \, \,  \ri(\qi) + \sum_{\substack{j \in \nbhi: \\ \Jij > 0}} \, \, \Big(\inf\limits_{\qj \in (0,1)} \, \, \rijp(\qi,\qj) \Big) + \sum_{\substack{j \in \nbhi: \\ \Jij < 0}} \, \, \Big( \inf\limits_{\qj \in (0,1)} \, \, \rijm(\qi,\qj) \Big) \, \, \, \Big] 
\end{align}
which is equal to the infimum of $\Ri(\qvec)$ as the following Lemma shows:

\begin{Lemma} \label{lemma:split_infima} {\ \\}
Let $\Ri(\qvec)$ be the expression \footnote{We remark that our proof of Lemma~\ref{lemma:split_infima} does not require the specific definition of the functions $\ri, \rijp,\rijm$, but only relies on the structure of the decomposition \eqref{eq:Ri_rewritten} of $\Ri(\qvec)$. } in~\eqref{eq:Ri_rewritten} and let $\tilde{R}_i$ be the expression in~\eqref{eq:sum_of_infima}. Then
\begin{align}
\inf\limits_{\qvec \in \Bbox} \, \, \Ri(\qvec) = \tilde{R}_i.
\end{align}
\end{Lemma}

Lemma~\ref{lemma:split_infima} justifies the described two-step procedure for evaluating $\inf\limits_{\qvec \in \Bbox} \Ri(\qvec)$ from 'inside' (fix $\qi \in (0,1)$ and compute the infima of $\rijp(\qi,\qj)$ and $\rijm(\qi,\qj)$ with respect to $\qj$ as a function of $\qi$) to 'outside' (given the solutions to these sub-problems, compute the infimum of $\Ri(\qvec)$ with respect to $\qi$). Hence, we must first solve the 'inner' sub-problems~\eqref{eq:infimum_rij} whose solutions are required in the 'outer' problem~\eqref{eq:sum_of_infima}. We have two types of sub-problems to be solved: one for ferromagnetic edges and one for antiferromagnetic edges. Instead of treating them separately, we show that they can be transformed into each other:

\begin{Lemma} \label{lemma:equivalence_rijp_rijm} {\ \\}
Let $\Jij > 0$, and let $\rijp(\qi,\qj)$ and $\rijm(\qi,\qj)$ be the two types of pairwise functions in the sum decomposition~\eqref{eq:Ri_rewritten} of $\Ri(\qvec)$, with $\rijp(\qi,\qj)$ implicitly depending on $\Jij$ (corresponding to a ferromagnetic edge) and $\rijm(\qi,\qj)$ implicitly depending on $- \Jij$ (corresponding to an antiferromagnetic edge). Then we have
\begin{align} \label{eq:equivalence_rijp_rijm}
\rijm(q_i,q_j) = \rijp(\qi,1-\qj) = \rijp(1-\qi,\qj).
\end{align}
In other words, if we reflect $\rijp(\qi,\qj)$ in its domain $(0,1) \times (0,1)$ either across the line $\qi=0.5$ or the line $\qj=0.5$, we obtain $\rijm(\qi,\qj)$ (and vice versa).
\end{Lemma}

\begin{figure}[ht]
\centering
\includegraphics[width=1.02\linewidth]{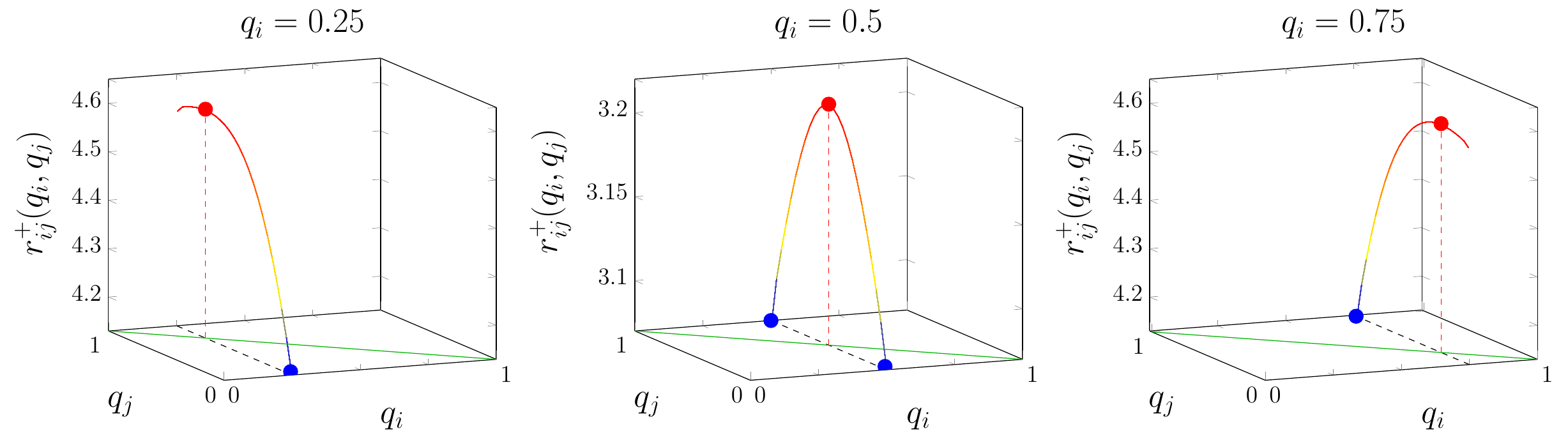}
\caption{$\rijp(\qi,\qj)$ as a function of $\qj \in (0,1)$ and for different fixed values of $\qi$. The green line $\qj = 1-\qi$ represents the space of all stationary points of $\rijp(\qi,\qj)$ (exactly one for each $\qi \in (0,1)$, Theorem~\ref{thm:unique_stationary_rijp}), which are always maxima (drawn as red points, Theorem~\ref{thm:maximum_rijp}). The blue points at the boundary represent the infima of $\rijp(\qi,\qj)$ with respect to $\qj$, which are never taken by $\rijp(\qi,\qj)$ (Lemma~\ref{lemma:rijp_critical_values} and Corollary~\ref{cor:solution_inf_rijp_qj}). Note that $\qi=0.5$ represents the only scenario where the infimum exists at both boundary points $0$ and $1$.}
\label{fig:sketch_rijp}
\end{figure}

Lemma~\ref{lemma:equivalence_rijp_rijm} establishes a direct relationship between $\rijm(\qi,\qj)$ and $\rijp(\qi,\qj)$. It therefore facilitates the determination of $\tilde{R}_i$ in~\eqref{eq:sum_of_infima} by focusing on a single edge type, instead of distinguishing between two types of edges. Without loss of generality, we may replace each pairwise term $\rijm(\qi,\qj)$ (which implicitly depends on $\Jij < 0$) in~\eqref{eq:sum_of_infima} with $\rijp(1-\qi,\qj)$ and the sign-inverted coupling $|\Jij|>0$. Under this assumption,~\eqref{eq:sum_of_infima} becomes
\begin{align} \label{eq:sum_of_infima_rewritten}
 \tilde{R}_i \coloneqq \inf\limits_{\qi \in (0,1)} \, \, \Big[ \, \, \, \ri(\qi) + \sum_{\substack{j \in \nbhi: \\ \Jij > 0}} \, \, \Big(\inf\limits_{\qj \in (0,1)} \, \, \rijp(\qi,\qj) \Big) + \sum_{\substack{j \in \nbhi: \\ \Jij < 0}} \, \, \Big( \inf\limits_{\qj \in (0,1)} \, \, \rijp(1-\qi,\qj) \Big) \, \, \, \Big].
\end{align}
We now return to the 'inner' sub-problems stated in~\eqref{eq:infimum_rij} with our focus on the first problem type: the determination of $\inf\limits_{\qj \in (0,1)} \, \, \rijp(\qi,\qj)$ with $\qi \in (0,1)$ arbitrary but fixed \footnote{By Lemma~\ref{lemma:equivalence_rijp_rijm}, the determination of $\inf\limits_{\qj \in (0,1)} \, \, \rijm(\qi,\qj)$ can be reduced to the ferromagnetic case.}. In particular, we consider $\rijp(\qi,\qj)$ as a one-dimensional function of $\qj$ defined on the interval $(0,1)$ (representing the line segment that connects $(\qi,0)$ and $(\qi,1)$ in the domain $(0,1) \times (0,1)$). To find the greatest lower bound on $\rijp$ in that interval, we first show that $\rijp$ has a unique stationary point in $(0,1)$ (Theorem~\ref{thm:unique_stationary_rijp}) which, however, is a maximum (Theorem~\ref{thm:maximum_rijp}). This implies that the infimum is not taken by $\rijp$ but exists as a limit value at the boundary of $(0,1)$. In Lemma~\ref{lemma:rijp_critical_values} and Corollary~\ref{cor:solution_inf_rijp_qj}, we derive the explicit formula for the infimum, in dependence of the chosen $\qi$. Figure~\ref{fig:sketch_rijp} shows sketches of $\rijp(\qi,\qj)$ for different values of $\qi$. \\ 

For proving Theorem~\ref{thm:unique_stationary_rijp}, we require the first-order partial derivatives of the functions $\xijopt, \Tij,$ and $\rijp$ with respect to $\qi$ and $\qj$. Moreover, we prepare the second-order partial derivatives of $\xijopt$ and $\Tij$, that we will require in the proof of Theorem~\ref{thm:HBijsubdet_at_betacrit} (Sec.~\ref{subsec:Bethe_Hessian_Decomposition}). \newpage

\begin{Lemma} \label{lemma:partial_derivatives_xij_Tij_rijp} {\ \\}
\begin{itemize}
 \item[(a)]
The first-order partial derivatives of $\xijopt(\qi,\qj), \Tij(\qi,\qj), \rijp(\qi,\qj)$ are
 \begin{align}
  \hspace{-1cm} & \frac{\partial \xijopt}{\partial q_{i}} = \frac{\aij (\qj - \xijopt) + \qj}{1 + \aij(\qi+\qj-2\xijopt)}, \label{eq:deri_1_xij_qi} \\
  \hspace{-1cm} & \frac{\partial \xijopt}{\partial q_{j}} = \frac{\aij (\qi - \xijopt) + \qi}{1 + \aij(\qi+\qj-2\xijopt)}, \label{eq:deri_1_xij_qj} \\
  \hspace{-1cm} & \frac{\partial \Tij}{\partial q_{i}} = (1-2\qi)\qj(1-\qj) - 2(\xijopt - \qi \qj) (\frac{\partial \xijopt}{\partial q_{i}} - \qj), \label{eq:deri_1_Tij_qi} \\
  \hspace{-1cm} & \frac{\partial \Tij}{\partial q_{j}} = (1-2\qj)\qi(1-\qi) - 2(\xijopt - \qi \qj) (\frac{\partial \xijopt}{\partial q_{j}} - \qi), \label{eq:deri_1_Tij_qj} \\
  \hspace{-1cm} & \frac{\partial \rijp}{\partial q_{i}} = \frac{(-\frac{\partial \xijopt}{\partial q_{i}} + \qj)\Tij - (\qj(1-\qj)-\xijopt + \qi\qj)\frac{\partial \Tij}{\partial q_{i}}}{\Tij^2}, \label{eq:deri_1_rijp_qi} \\
  \hspace{-1cm} & \frac{\partial \rijp}{\partial q_{j}} = \frac{(1-2\qj-\frac{\partial \xijopt}{\partial q_{j}} +\qi)\Tij - (\qj(1-\qj)-\xijopt + \qi\qj)\frac{\partial \Tij}{\partial q_{j}}}{\Tij^2}. \label{eq:deri_1_rijp_qj}
 \end{align}
 \item[(b)]
 The second-order partial derivatives of $\xijopt(\qi,\qj)$ and $\Tij(\qi,\qj)$ are
 \begin{align}
  \hspace{-1cm} & \frac{\partial^2 \xijopt}{\partial \qi^2} = \aij \frac{\big(-\frac{\partial \xijopt}{\partial q_{i}} \big) \cdot A_{ij} - \big(\aij(\qj-\xijopt)+\qj \big) \big(1 - 2 \frac{\partial \xijopt}{\partial q_{i}}\big)}{A_{ij}^2}, \label{eq:deri_2_xij_qi2} \\
  \hspace{-1cm} & \frac{\partial^2 \xijopt}{\partial \qi \partial \qj}  = \frac{\aij\big[\big(1-\frac{\partial \xijopt}{\partial q_{j}}+\frac{1}{\aij} \big) \cdot A_{ij} -  \big(\aij(\qj-\xijopt)+\qj \big) \big(1 - 2 \frac{\partial \xijopt}{\partial q_{j}}\big)\big]}{A_{ij}^2}, \label{eq:deri_2_xij_qi_qj} \\
  \hspace{-1cm} & \frac{\partial^2 \xijopt}{\partial \qj^2} = \aij \frac{\big(-\frac{\partial \xijopt}{\partial q_{j}} \big) \cdot A_{ij} - \big(\aij(\qi-\xijopt)+\qi \big) \big(1 - 2 \frac{\partial \xijopt}{\partial q_{j}}\big)}{A_{ij}^2}, \label{eq:deri_2_xij_qj2} \\
  \hspace{-1cm} & \frac{\partial^2 \Tij}{\partial \qi^2} = -2\qj(1-\qj) -2 \Big(\big(\frac{\partial \xijopt}{\partial q_{i}} - \qj\big)^2 + (\xijopt - \qi\qj) \big( \frac{\partial^2 \xijopt}{\partial \qi^2}\big) \Big), \label{eq:deri_2_Tij_qi2} \\
  \hspace{-1cm} & \frac{\partial^2 \Tij}{\partial \qi \partial \qj} = (1-2\qj)(1-2\qi) - 2\Big(\big(\frac{\partial \xijopt}{\partial q_{i}} - \qj\big) \big(\frac{\partial \xijopt}{\partial q_{j}} - \qi\big) +(\xijopt - \qi\qj) \big( \frac{\partial^2 \xijopt}{\partial \qi \partial \qj}-1\big)\Big), \label{eq:deri_2_Tij_qi_qj} \\
  \hspace{-1cm} & \frac{\partial^2 \Tij}{\partial \qj^2} = -2\qi(1-\qi) -2 \Big(\big(\frac{\partial \xijopt}{\partial q_{j}} - \qi\big)^2 + (\xijopt - \qi\qj) \big( \frac{\partial^2 \xijopt}{\partial \qj^2}\big) \Big), \label{eq:deri_2_Tij_qj2}
 \end{align}
 where we have defined $A_{ij} \coloneqq \big(1+\aij(\qi+\qj-2\xijopt) \big)$.
 \end{itemize}
\end{Lemma} \newpage


\begin{thm} \label{thm:unique_stationary_rijp} {\ \\}
Let $\qi$ be an arbitrary (but fixed) value in $(0,1)$ and consider $\rijp(\qi,\qj)$ as a function of $\qj$ in the interval $(0,1)$. The only stationary point of $\rijp(\qi,\qj)$ in that interval is $\qj^{\ast} = 1-\qi$. In other words, the nonlinear equation
\begin{equation} \label{eq:stationary_equation_rijp}
 \frac{\partial \rijp}{\partial q_{j}} = 0
\end{equation}
has the unique solution $\qj^{\ast} = 1-\qi$.
\end{thm}

Next, we analyze what kind of stationary point $\rijp(\qi,\qj)$ has in $\qj^\ast = 1-\qi$ and if this depends on the choice of $\qi$. We answer this in Theorem~\ref{thm:maximum_rijp}, by comparing the value of $\rijp$ in $\qj^\ast$ to its limit values at the boundary points. These are computed in the next Lemma:

\begin{Lemma} \label{lemma:rijp_critical_values} {\ \\}
Let $\qi \in (0,1)$ be arbitrary (but fixed)- and consider $\rijp(\qi,\qj)$ as a function of $\qj$ in $(0,1)$.
\begin{enumerate}
 \item [(a)] In its unique stationary point $\qj^{\ast}=1-\qi$, $\rijp$ takes the value
 \begin{align} \label{eq:rijp_value_in_stationary}
  \rijp(\qi,1-\qi) = \frac{1}{\xijopt(\qi,1-\qi)}.
 \end{align}
 \item [(b)] If $\rijp$ approaches the boundary points of $(0,1)$, its limit values are
\begin{align}
 & \lim\limits_{\qj \to 0} \rijp(\qi,\qj) = \frac{1+\aij \qi^2}{(1+\aij\qi)\qi(1-\qi)} , \quad \text{and} \label{eq:limit_values_rijp_zero} \\
 & \lim\limits_{\qj \to 1} \rijp(\qi,\qj) = \frac{1+\aij (1-\qi)^2}{\big(1+\aij(1-\qi)\big)\qi(1-\qi)}. \label{eq:limit_values_rijp_one}
\end{align}
This further implies the relation
\begin{align}
 \lim\limits_{\qj \to 1} \rijp(\qi,\qj) = \lim\limits_{\qj \to 0} \rijp(1-\qi,\qj). \label{eq:symmetry_limit_values_rijp}
\end{align}

\end{enumerate}
\end{Lemma}

For a more compact notation, let us define 
\begin{align}
 & r_{ij}^{(+,0)}(\qi) \coloneqq \lim\limits_{\qj \to 0} \rijp(\qi,\qj), \quad \text{and} \\
 & r_{ij}^{(+,1)}(\qi) \coloneqq \lim\limits_{\qj \to 1} \rijp(\qi,\qj).
\end{align}

Note that $r_{ij}^{(+,0)}$ and $r_{ij}^{(+,1)}$ are not functions of $\qj$ any longer but only depend on $\qi$. \\

The next Theorem shows that $\qj^{\ast}=1-\qi$ is always a maximum of $\rijp(\qi,\qj)$ (still considered as a function of $\qj$).

\begin{thm} \label{thm:maximum_rijp} {\ \\}
 We have the following relations between the values of $\rijp(\qi,\qj)$ in its stationary point $\qj^{\ast}=1-\qi$ and its limit values at the boundary: 
 \begin{itemize}
  \item If $0 < \qi < 0.5$, then $r_{ij}^{(+,0)}(\qi) < r_{ij}^{(+,1)}(\qi) < \rijp(\qi,1-\qi)$.
  \item If $0.5 < \qi < 1$, then $r_{ij}^{(+,1)}(\qi) < r_{ij}^{(+,0)}(\qi) < \rijp(\qi,1-\qi)$.
  \item If $\qi = 0.5$, then $r_{ij}^{(+,0)}(\qi) = r_{ij}^{(+,1)}(\qi) < \rijp(\qi,1-\qi)$.
 \end{itemize}
In particular, for any choice of $\qi \in (0,1)$, the unique stationary point $\qj^{\ast} = 1-\qi$ is a maximum of $\rijp(\qi,\qj)$.
\end{thm}

Let us recap the insights from the previous statements: To find the infimum of $\rijp(\qi,\qj)$ with respect to $\qj \in (0,1)$ as the variable and $\qi \in (0,1)$ arbitrary but fixed, we have first shown that this function has the unique stationary point $\qj^{\ast} = 1-\qi$ (Theorem~\ref{thm:unique_stationary_rijp}). We argued that this must be a maximum (Theorem~\ref{thm:maximum_rijp}), which implies that the infimum is not taken by $\rijp(\qi,\qj)$ in the interior of $(0,1)$, but approached at one of the boundary points (i.e, for $\qj \to 0$ or for $\qj \to 1$). We have derived an analytical formula for the limit values of $\rijp(\qi,\qj)$ at these boundary points as a function of $\qi$ (Lemma~\ref{lemma:rijp_critical_values}), and determined which of the two limit values is less than the other, depending on the choice of $\qi \in (0,1)$. Together with the other observations, we conclude that this must then be the infimum.

\begin{cor} \label{cor:solution_inf_rijp_qj} {\ \\}
 For $\qi \in (0,1)$ arbitrary (but fixed), we have
 \begin{align*}
  \inf\limits_{\qj \in (0,1)} \rijp(\qi,\qj) = 
  \begin{dcases}
    \frac{1+\aij \qi^2}{(1+\aij\qi)\qi(1-\qi)}, \quad & \text{if} \quad 0 < \qi \leq 0.5, \quad \text{and} \\
    \frac{1+\aij (1-\qi)^2}{\big(1+\aij(1-\qi)\big)\qi(1-\qi)}, \quad & \text{if} \quad 0.5 < \qi < 1.
  \end{dcases}
 \end{align*}
\end{cor}

\begin{figure}[h]
\centering \includegraphics[width=0.36\linewidth]{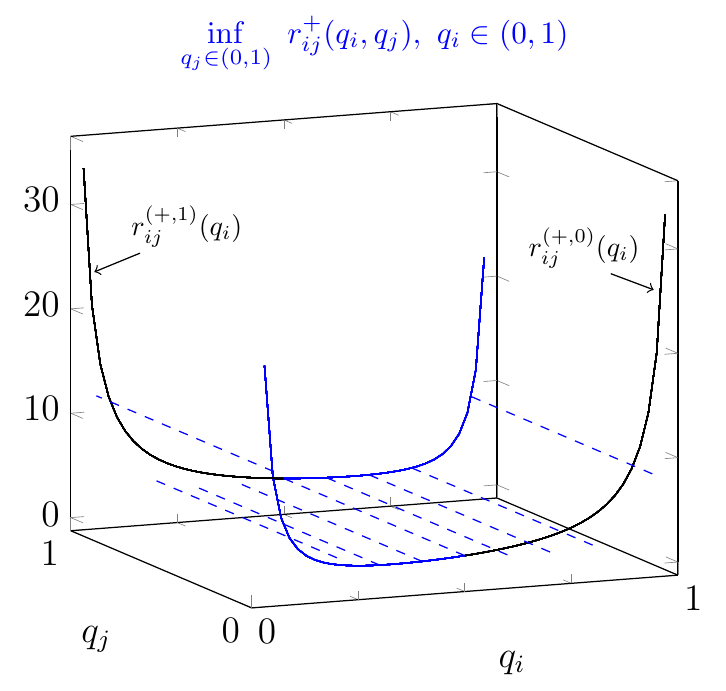} 
\caption{Visualization of the limit values of $\rijp(\qi,\qj)$ at the boundary points $\qj=0$ and $\qj=1$ as a function of $\qi$. The solid blue curves indicate the position of the infimum, which is approached for $\qj \to 0$ if $\qi \leq 0.5$, and for $\qj \to 1$ if $\qi \geq 0.5$ (Corollary~\ref{cor:solution_inf_rijp_qj}). The dashed blue lines represent the level of the infimum for different fixed values of $\qi$.}
\label{fig:rijp_infimum}
\end{figure}

Figure~\ref{fig:rijp_infimum} illustrates $\inf\limits_{\qj \in (0,1)} \rijp(\qi,\qj)$ as a function of $\qi$ (drawn in solid blue curves).

Corollary~\ref{cor:solution_inf_rijp_qj} provides the solution to the 'inner' sub-problems stated in~\eqref{eq:infimum_rij}. Hence, we return to the 'outer' problem~\eqref{eq:sum_of_infima_rewritten} which is the evaluation of $\tilde{\Ri}$. Before we substitute the formulas derived in Corollary~\ref{cor:solution_inf_rijp_qj} and the expression for $\ri(\qi)$ from~\eqref{eq:Ri_ri} into~\eqref{eq:sum_of_infima_rewritten}, we observe that, for all $\qi \in (0,1)$,
\begin{align} \label{eq:symmetry_infimum_rijp}
 \inf\limits_{\qj \in (0,1)} \rijp(\qi,\qj) = \inf\limits_{\qj \in (0,1)} \rijp(1-\qi,\qj).
\end{align}
This follows directly from the relation~\eqref{eq:symmetry_limit_values_rijp} from Lemma~\ref{lemma:rijp_critical_values}.
Consequently, $\inf\limits_{\qj \in (0,1)} \rijp(\qi,\qj)$ (as a function of $\qi$) is symmetric across $\qi=0.5$, which makes it sufficient to evaluate $\tilde{\Ri}$ for $\qi \in (0,0.5]$.
If we assume that, for each edge, we only take the coupling strength $|\Jij|$ but not the sign of $\Jij$ into account (which is justified by Lemma~\ref{lemma:equivalence_rijp_rijm}), equation~\eqref{eq:sum_of_infima_rewritten} becomes
\begin{align}
 \tilde{R}_i & = \inf\limits_{\qi \in (0,0.5)} \, \, \Big[ \frac{-(\degi-1)}{\qi(1-\qi)} +  \sum_{\substack{j \in \nbhi}} \, \, \frac{1+\aij \qi^2}{(1+\aij\qi)\qi(1-\qi)}  \Big] \\
  & \hspace{-0.5cm} = \inf\limits_{\qi \in (0,0.5)} \, \, \Big[ \frac{-(\degi-1) \prod\limits_{j \in \nbhi} (1+\aij\qi) +  \sum\limits_{j \in \nbhi} \big((1+\aij\qi^2) \prod\limits_{k \in \nbhi \setminus j} (1+\alpha_{ik}\qi)\big)}{\qi(1-\qi) \prod\limits_{j \in \nbhi} (1+\aij\qi)}  \Big]. \label{eq:quotient_Psii}
\end{align}
To check the diagonal dominance criterion~\eqref{eq:diagonal_dominance_Bethe_infimum} stated in Proposition~\ref{prop:problem_formulation_diagonal_dominance}, we do not need to compute the exact value of this infimum (which, by Lemma~\ref{lemma:split_infima}, is identical to $\inf\limits_{\qvec \in \Bbox} \, \, \Ri(\qvec)$), but only have to determine if it is positive. One simple observation further facilitate this task: All terms in the denominator of~\eqref{eq:quotient_Psii} are positive; i.e., positivity of the infimum depends only on the numerator, which is a polynomial in a single variable $\qi$ and of degree $\degi+1$. \\ 
Altogether, we present the final Theorem of this subsection:

\begin{thm} \label{thm:polynomial_characterization_of_inf_Ri}  {\ \\}
 The convexity criterion stated in Proposition~\ref{prop:problem_formulation_diagonal_dominance} is equivalent to the following condition: The Bethe free energy $\FB$ is convex on the Bethe box $\Bbox$ if, for all nodes $i \in \setofnodes$, the one-dimensional polynomial 
 \begin{align} \label{eq:diagonal_dominance_final_condition}
  \Psii(\qi) \coloneqq -(\degi-1) \prod\limits_{j \in \nbhi}(1+\aij\qi) + \sum\limits_{j \in \nbhi} \Big((1+\aij\qi^2) \prod\limits_{k \in \nbhi \setminus j}(1+\alpha_{ik}\qi) \Big)
 \end{align}
 is strictly positive on the interval $(0,0.5]$.
\end{thm}

\begin{figure}[h]
\centering
\includegraphics[width=0.4165\linewidth]{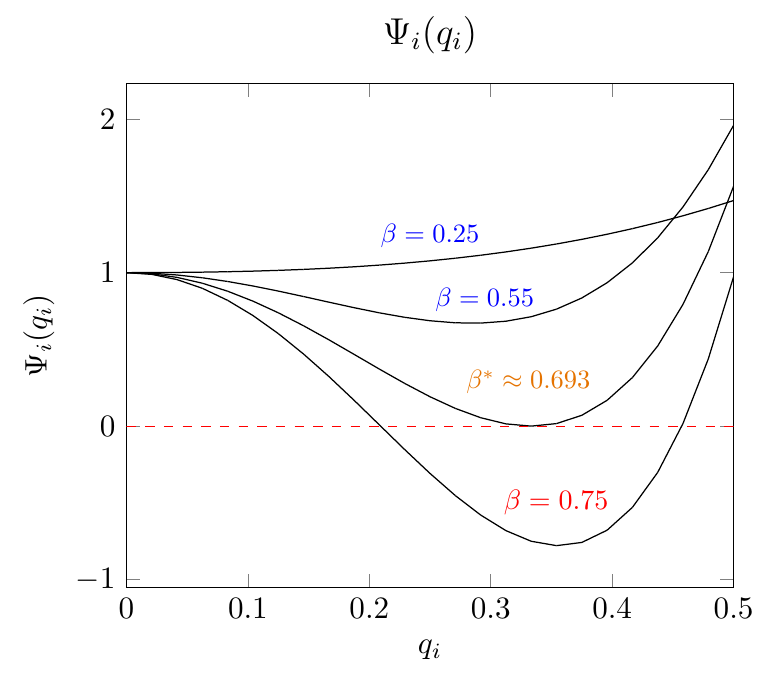}
\caption{Development of the polynomial $\Psii(\qi)$ for increasing values of the inverse tempera\-ture $\beta$, associated to a node $i$ with three neighbors ($\degi=3$), and a coupling $\Jij = J = 0.5$ that is shared by all three incident edges. For $\beta=0.25$ and $\beta=0.55$, condition~\eqref{eq:diagonal_dominance_final_condition} from Theorem~\ref{thm:polynomial_characterization_of_inf_Ri} is satisfied with respect to node $i$ (blue). $\betacrit \approx 0.693$ is the critical threshold at which~\eqref{eq:diagonal_dominance_final_condition} is violated for the first time, as $\Psii(\qi)$ touches the horizontal axis and takes root in the interval $(0,0.5)$ (orange); for higher values of the inverse temperature (e.g., $\beta=0.75$), $\Psii(\qi)$ has multiple roots in $(0,0.5)$ which causes~\eqref{eq:diagonal_dominance_final_condition} to be violated (red).}
\label{fig:Psii} 
\end{figure}

Theorem~\ref{thm:polynomial_characterization_of_inf_Ri} is the central result of this subsection. It provides a suffient condition for the convexity of the Bethe free energy $\FB$ on the Bethe box $\Bbox$. Figure~\ref{fig:Psii} shows the behavior of the polynomial $\Psii(\qi)$ in dependence of the inverse temperature $\beta$. \newpage

\subsection{Edge-Specific Convexity Criterion: Sum Decomposition of Bethe Hessian} \label{subsec:Bethe_Hessian_Decomposition}

Our second approach to relax the convexity of $\FB$ decomposes the Bethe Hessian into a sum of sparse matrices, and derives a sufficient condition for the positive definiteness of the individual matrices in that sum. The main result is Theorem~\ref{thm:determinant_characterization_of_convexity}, a closed formula for the critical threshold of the inverse temperature, at which convexity of the Bethe free energy is not guaranteed any longer. All proofs for this section are collected in Appendix~\ref{sec:appendix_B}. \\

The starting point is the same as in Sec.~\ref{subsec:Bethe_Diagonal_Dominance}, namely to ask under which conditions the Bethe Hessian $\HB(\bm{q})$ is positive definite for all $\qvec$ in the Bethe box $\Bbox$ (the definition of convexity, Sec.~\ref{subsec:mathematical}). However, we use an alternative approach to relax this problem: While, in Sec.~\ref{subsec:Bethe_Diagonal_Dominance}, we have applied the diagonal dominance criterion to the entire Bethe Hessian, we now decompose $\HB$ into a sum of sparse matrices $\HB^{\prime}$ (i.e., by writing $\HB = \sum \HB^{\prime}$) and then characterize the positive definiteness of the individual matrices in that sum. In case that all $\HB^{\prime}$ are positive definite over $\Bbox$, their sum $\HB$ is positive definite over $\Bbox$ as well. This induces a sufficient (but not necessary \footnote{Note that $\HB$ might still be positive definite if some (or even all) matrices $\HB^{\prime}$ in the sum decomposition are not. This explains why our approach is merely a relaxation to the problem of characterizing the positive definiteness of $\HB$.}) condition for the convexity of $\FB$ on the Bethe box $\Bbox$. Further differences between the our approaches will be discussed in Sec.~\ref{subsec:Theoretical_Discussion}. \\

The first problem we are facing is how to define the matrices $\HB^{\prime}$ in a sum decomposition of $\HB$. On the one hand, they should be simple enough to allow for an exact characterization of their definiteness properties. On the other hand, too many (simple) matrices in the sum may induce a relatively high gap between the 'true' and the predicted convexity (as a sufficient criterion for convexity requires positive definiteness of each individual matrix in the sum). Instead of trying to minimize this gap, we propose an edge-specific class of sum decompositions, in which the individual matrices are naturally related to the edges in the graph (i.e., for each edge $(i,j)$ we have one associated matrix $\HBij$ in the sum). This particular choice can be motivated by the properties of the Bethe free energy, which is itself a sum of edge- and node-specific statistics~\eqref{eq:Bethe_reparameterized}. While it might be tempting to write the Bethe Hessian as a sum of Hessians corresponding to these statistics (i.e., computing the Hessians of the individual terms in~\eqref{eq:Bethe_reparameterized} and summing them up), we must consider one aspect: the local entropies $\mathcal{S}_{i}$ are always concave and their Hessians can never be positive definite (i.e., considering them as separate matrices in the sum can never result in a sufficient criterion for convexity). Also, these are the only node-specific statistics contributing to the Bethe Hessian, because $\HB$ is independent from the fields $\theta_i$ (Corollary~\ref{cor:independence_of_theta}). Thus, instead of defining separate 'node-specific Hessians' $\HB^{i}$ for each node $i$, it is preferable to absorb the node-specific statistics from $\HB$ into the 'edge-specific Hessians' $\HBij$. \\

We consider the following class of sum decompositions: Let us write the Bethe Hessian as 
\begin{align} \label{eq:edge_specific_sum_decomposition}
\HB = \sum\limits_{(i,j) \in \setofedges} \HBij,
\end{align}
with the entries of $\HBij$ -- for an arbitrary edge $(i,j)$ and indices $k,l \in \{1,\dots,|\setofnodes|\}$ -- being defined as
\begin{align}
  \begin{split} \label{eq:HBij_entries}
   \hspace{-0.4cm} \big(\HBij\big)_{k,l} =
   \begin{dcases}
    \frac{1}{\beta} \, \Big(\! -s_{ij} \, \frac{(\degi -1)}{\qi (1-\qi)} + \frac{\qj (1-\qj)}{\Tij} \Big) & \quad k = l = i \\
    \frac{1}{\beta} \, \Big(\! -s_{ji} \, \frac{(\degj -1)}{\qj (1-\qj)} + \frac{\qi (1-\qi)}{\Tij} \Big) & \quad k = l = j \\
    \frac{1}{\beta} \, \Big(\frac{\qi \qj - \xijopt}{\Tij} \Big) & \quad k,j \in \{i,j\}, \, k \neq l \\
    0 & \quad \text{otherwise}.
   \end{dcases}
   \end{split}
\end{align}

This definition requires some explanation. First, note that we have introduced parameters $s_{ij}$ and $s_{ji}$ (i.e, two parameters per edge) whose purpose is to 'distribute' the second-order local entropy contributions from $\HB$ (using the equations from~\eqref{eq:Bethe_second_derivatives}, (b)) over the edge-specific Hessians $\HBij$. More precisely, $s_{ij}$ represents the rela\-tive proportion of the second-order local entropy contribution related to node $i$ (i.e, of $\frac{\partial^2}{\partial \qi^2}\big(-\frac{1}{\beta} (\degi-1)\mathcal{S}_{i}\big) = -\frac{1}{\beta} \frac{(\degi-1)}{\qi(1-\qi)}$) that we absorb from $\HB$ into the edge-specific Hessian $\HBij$. Analogously, $s_{ji}$ does the same with the second-order local entropy contribution related to node $j$ (i.e., with $-\frac{1}{\beta} \frac{(\degj-1)}{\qj(1-\qj)}$)). In other words, for each node $i$, we distribute its second-order local entropy contributions from $\HB$ over all edge-specific Hessians $\HBij$ with $j \in \nbhi$. To make this a reasonable approach, where each $s_{ij}$ represents a certain proportion of the corresponding second-order local entropy contribution, we assume that all parameters satisfy 
\begin{align} \label{eq:sij_constraint_1}
 0 \leq s_{ij}, s_{ji} \leq 1.
\end{align}
 Then, to guarantee that the sum decomposition~\eqref{eq:edge_specific_sum_decomposition} holds, we introduce linear constraints
\begin{align} \label{eq:sij_constraint_2}
 \sum\limits_{j \in \nbhi} s_{ij} = 1
\end{align}
for all nodes. By this, the second-order local entropy contributions are exactly distributed over the edge-specific Hessians.
The other terms in the definition~\eqref{eq:HBij_entries} of $\HBij$ (i.e., the main diagonal entries $\frac{1}{\beta} \big(\frac{\qj (1-\qj)}{\Tij}\big)$ resp. $\frac{1}{\beta} \big(\frac{\qi (1-\qi)}{\Tij}\big)$, and the off-diagonal entries $\frac{1}{\beta} \, \Big(\frac{\qi \qj - \xijopt}{\Tij} \Big)$) stem from the second-order pairwise entropy contributions $\frac{\partial^2}{\partial \qi^2} \big(- \frac{1}{\beta} \big(\mathcal{S}_{ij}\big)\big)$ in $\HB$. \\

One favorable property of this specific class of sum decompositions is the sparseness of its involved matrices. From~\eqref{eq:HBij_entries}, we observe that each edge-specific Hessian $\HBij$ has precisely four non-zero entries at the index positions $[i,i], [i,j], [j,i]$, and $[j,j]$, with identical entries at $[i,j]$ and $[j,i]$. More generally, we deal with a sum of quadratic matrices of the form
\begin{align} \label{eq:general_matrix_form}
\textbf{M}^{(i,j)} \coloneqq
\begin{pmatrix}
  0 & \cdots  &   & 0 & \cdots  &  & 0 \\
  \vdots & \ddots & & \vdots & & & \vdots \\
  & & a^{(i,j)} & \cdots & c^{(i,j)} & & \\
  0 & \cdots & \vdots & & \vdots & \cdots & 0 \\
  \vdots& & c^{(i,j)} & \cdots & b^{(i,j)} & &\vdots \\
   & & & \vdots  & & \ddots &  \\
  0 &\cdots  &  & 0 & \cdots &  & 0
\end{pmatrix}
\end{align}
where at most one matrix in the sum can have non-zero entries at a specific pair of off-diagonal positions $[i,j]$ and $[j,i]$ (which is precisely the case if $(i,j)$ is an edge in the graph). Here is another subtlety to be addressed: Strictly speaking, this kind of matrix can never be positive definite -- but only positive semidefinite -- as one can always find vectors $\bm{y} \neq \bm{0}$ such that $\bm{y}^T \textbf{M}^{(i,j)} \bm{y} = \bm{0}$. However, one can show that a sum of positive semidefinite matrices of the form~\eqref{eq:general_matrix_form} is positive definite, if only the $(2 \times 2)$ - submatrices
\begin{align} \label{eq:submatrix_2x2}
\textbf{M}_{2\times2}^{(i,j)} \coloneqq
\begin{pmatrix}
  a^{(i,j)} & c^{(i,j)} \\
  c^{(i,j)} & b^{(i,j)}
\end{pmatrix}
\end{align}
are positive definite for each individual matrix $\textbf{M}^{(i,j)}$ in the sum \footnote{To prove this, write $\bm{y}^T \big(\sum \textbf{M}^{(i,j)} \big) \bm{y}$ equivalently as a sum of quadratic forms $\sum \begin{pmatrix} y_i & y_j \end{pmatrix} \textbf{M}_{2\times2}^{(i,j)} \begin{pmatrix} y_i \\ y_j \end{pmatrix}$ (by exploiting the structure~\eqref{eq:general_matrix_form} of $\textbf{M}^{(i,j)}$) and use the positive definiteness of all submatrices $\textbf{M}_{2\times2}^{(i,j)}$.}. This facilitates our analysis, as we can focus on studying the definiteness properties of $(2 \times 2)$ - submatrices of the edge-specific Hessians $\HBij$ defined as
\begin{align} \label{eq:sub_Hessian}
 \HBijsub \coloneqq \frac{1}{\beta} \begin{pmatrix}
   -s_{ij} \, \frac{(\degi -1)}{\qi (1-\qi)} + \frac{\qj (1-\qj)}{\Tij} & \frac{\qi \qj - \xijopt}{\Tij} \\
   \frac{\qi \qj - \xijopt}{\Tij} & -s_{ji} \, \frac{(\degj -1)}{\qj (1-\qj)} + \frac{\qi (1-\qi)}{\Tij}
\end{pmatrix}
.
\end{align} \vspace{0.1cm}

So far, we have explained our approach to deriving a sufficient criterion for convexity of the Bethe free energy $\FB$ on the Bethe box $\Bbox$ by writing its Hessian $\HB$ (Lemma~\ref{lemma:Bethe_derivatives}, (b)) as a sum of matrices and proving positive definiteness of the individual matrices in that sum. Moreover, we have introduced a class of edge-specific sum decompositions~\eqref{eq:edge_specific_sum_decomposition} and shown that an edge-specific Hessian~\eqref{eq:HBij_entries} is positive semidefinite, if and only if a $(2 \times 2$) - submatrix $\HBijsub$ of $\HBij$, defined by~\eqref{eq:sub_Hessian}, is positive definite. We have also argued that positive semidefiniteness of all edge-specific Hessians $\HBij$ is sufficient for positive definiteness of the Bethe Hessian $\HB$.  In the remainder of this section, we analyze when the submatrices $\HBijsub$ are positive definite \footnote{Note that $\HB$ and the matrices in its sum decomposition are functions defined over the Bethe box, i.e., depend on $\qvec \in \Bbox$. This implies that the edge-specific Hessians and their submatrices must be positive (semi-)definite for all pairs $(\qi,\qj)$ in $(0,1)\times(0,1)$ (as $\HBij$ depends only on the associated variables $\qi$ and $\qj$) to guarantee convexity of $\FB$.}; i.e., in dependence of its involved parameters (which are the independent variables $\qi,\qj$; the model-specific parameters $\degi$, $\degj$, $\Jij$; the constrained parameters $s_{ij}, s_{ji}$; and the inverse temperature $\beta$). \\

According to Sylvester's criterion (Sec.~\ref{subsec:mathematical}), a quadratic real matrix is positive definite if and only if all its leading principle minors are positive. In the case of a $(2 \times 2)$ - matrix, these are the upper left element of the matrix and the determinant of the entire matrix. This implies that a submatrix $\HBijsub$ of $\HBij$ is positive definite in an arbitrary point $(\qi,\qj)$ in $(0,1) \times (0,1)$, if and only if the inequalities
\begin{align}
 & -s_{ij} \, \frac{(\degi -1)}{\qi (1-\qi)} + \frac{\qj (1-\qj)}{\Tij} >0 \quad \text{and} \label{eq:HBijsub_first_minor} \\
 \begin{split}
 & \Big( - s_{ij} \, \frac{(\degi -1)}{\qi (1-\qi)} + \frac{\qj (1-\qj)}{\Tij}\Big) \cdot \Big( - s_{ji} \, \frac{(\degj -1)}{\qj (1-\qj)} + \frac{\qi (1-\qi)}{\Tij}\Big) \\ 
 & - \Big(\frac{\qi \qj - \xijopt}{\Tij} \Big)^2 >0 \label{eq:HBijsub_second_minor}
 \end{split}
\end{align}
are satisfied (for some prespecified parameters $s_{ij},s_{ji}$). The left-hand side of~\eqref{eq:HBijsub_second_minor} is the determinant of $\HBijsub$, where we have omitted the positive scale factor $\frac{1}{\beta^2}$ as it does not influence the sign of the determinant (still, we have implicit dependence on the inverse temperature $\beta$ via $\xij$ and $\Tij$). These formulas are analytically complex, as $\xij$ and $\Tij$ are themselves functions of $\qi,\qj$ (defined in~\eqref{eq:xi_optimal} and~\eqref{eq:Tij_optimal}). Also, both inequalites must be satisfied for all $(\qi,\qj)$ in $(0,1) \times (0,1)$ (or, equivalently, for the infimum over all such pairs) which complicates their verification. To simplify the analysis, we further specify our definition of edge-specific Hessians. Recall that there is some freedom in selecting the constrained parameters $s_{ij}, s_{ji}$, as long as they satisfy the constraints~\eqref{eq:sij_constraint_1} and~\eqref{eq:sij_constraint_2}. A particularly natural choice arises if we distribute the second-order local entropy contributions related to node $i$ uniformly over the edge-speficic Hessians $\HBij$ (for $j$ in $\nbhi$); i.e., if we take $s_{ij} = \frac{1}{\degi}$ and $s_{ji} = \frac{1}{\degj}$. Clearly, this is a valid choice since $0 < \frac{1}{\degi} < 1$ and $\sum\limits_{j \in \nbhi} \frac{1}{\degi} = \degi \cdot \frac{1}{\degi}= 1$. Note that other choices of $s_{ij}, s_{ji}$ might induce more powerful criteria for convexity \footnote{One possible alternative would be to distribute the second-order local entropy contribution according to the strength of couplings related to the edges between neighboring nodes; i.e., take $s_{ij}= \frac{|\Jij|}{\sum\limits_{k \in \nbhi}|J_{ik}|}$ . Also, it remains an open problem if we can 'optimize' the choice of $s_{ij}, s_{ji}$ to minimize the gap between the convexity induced by a certain criterion and the true convexity.} (as this might be the case for different kind of sum decompositions than the edge-specific one); however, our choice entails the positive aspect that inequality~\eqref{eq:HBijsub_first_minor} is always satisfied:
\begin{align*}
 & -s_{ij} \, \frac{(\degi -1)}{\qi (1-\qi)} + \frac{\qj (1-\qj)}{\Tij} \\
 = \, &  - \frac{1}{\degi} \, \frac{(\degi -1)}{\qi (1-\qi)} + \frac{\qj (1-\qj)}{\Tij} \\ 
 \stackrel{\eqref{eq:Tij_optimal}}{=} \, & - \frac{1}{\degi} \, \frac{(\degi -1)}{\qi (1-\qi)} + \frac{\qj (1-\qj)}{\qi\qj(1-\qi)(1-\qj) - (\xijopt-\qi\qj)^2} \\
 > \, & - \frac{1}{\degi} \, \frac{(\degi -1)}{\qi (1-\qi)} + \frac{\qj (1-\qj)}{\qi\qj(1-\qi)(1-\qj)} \\
 = \, & \frac{-(\degi-1) + \degi }{\degi \qi (1-\qi)} \, = \, \frac{1}{\degi \qi (1-\qi)} \, > \, 0 \\
\end{align*}
Using this 'uniform' choice of $s_{ij},s_{ji}$, we can thus focus on inequality~\eqref{eq:HBijsub_second_minor} whose left-hand side is the determinant of $\HBijsub$, defined as
\begin{align} \label{eq:HBijsub_determinant_first_version}
\begin{split}
 & \HBijsubdet(\qi,\qj) \\
  & \coloneqq \Big(\frac{-(\degi -1)}{\degi \qi (1-\qi)} + \frac{\qj (1-\qj)}{\Tij}\Big) \cdot \Big(\frac{-(\degj-1)}{\degj \qj (1-\qj)} + \frac{\qi (1-\qi)}{\Tij}\Big) - \Big(\frac{\qi \qj - \xijopt}{\Tij} \Big)^2.
 \end{split}
\end{align}

Before we proceed, we simplify the expression in~\eqref{eq:HBijsub_determinant_first_version} and work with a more compact formula of $\HBijsubdet$. Namely,~\eqref{eq:HBijsub_determinant_first_version} is the same as

\begin{flalign}
 & \frac{\degi-1}{\degi} \cdot \frac{\degj-1}{\degj} \cdot \frac{1}{\qi\qj(1-\qi)(1-\qj)} - \frac{1}{\Tij} \cdot \Big(\frac{\degi-1}{\degi} + \frac{\degj-1}{\degj} \Big) \nonumber \\ 
 + \, & \frac{\overbrace{\qi\qj(1-\qi)(1-\qj) - (\qi\qj-\xijopt)^2}^{=\Tij}}{\Tij^2} \nonumber \\ \label{eq:HBijsub_determinant}
 = \, & \frac{\degi-1}{\degi} \cdot \frac{\degj-1}{\degj} \cdot \frac{1}{\qi\qj(1-\qi)(1-\qj)} + \frac{1}{\Tij} \Big(1 - \frac{\degi-1}{\degi} - \frac{\degj-1}{\degj} \Big).
\end{flalign}

Let us summarize the essentials of our present analysis:

\begin{propos} \label{prop:problem_formulation_sum_decomposition} {\ \\}
 Let $\HBijsub(\qi,\qj)$ be the submatrices of the edge-specific Hessians $\HBij(\qi,\qj)$ defined in~\eqref{eq:sub_Hessian} resp.~\eqref{eq:HBij_entries} together with a set of parameters $\{s_{ij}, s_{ji}\}$ for each edge $(i,j)$ satisfying the constraints~\eqref{eq:sij_constraint_1} and~\eqref{eq:sij_constraint_2}. Then the Bethe free energy $\FB$ is convex on the Bethe box $\Bbox$, if all submatrices $\HBijsub(\qi,\qj)$ are positive definite for all $(\qi,\qj)$ in $(0,1) \times (0,1)$, i.e., if the inequalities~\eqref{eq:HBijsub_first_minor} and~\eqref{eq:HBijsub_second_minor} are satisfied. If we set $s_{ij} = \frac{1}{\degi}$, then $\HBijsub(\qi,\qj)$ is positive definite if and only if the determinant of $\HBijsub(\qi,\qj)$ defined in~\eqref{eq:HBijsub_determinant} is positive.
\end{propos} \vspace{0.2cm}

To use these qualitative statements for proving the convexity of $\FB$, we need to characterize the behavior of $\HBijsubdet(\qi,\qj)$ (as a function of $\qi$ and $\qj$) in dependence of its parameters.  In particular, we must determine whether $\HBijsubdet(\qi,\qj)$ is positive for all points $(\qi,\qj)$ in $(0,1) \times (0,1)$, or if there exists a point where it is non-positive (in which case positive definiteness of $\HBijsub$ would be violated). In principle, we could once again formulate an optimization problem in which we try to compute the infimum of $\HBijsubdet$ and determine whether it is positive or not (analogously to problem~\eqref{eq:diagonal_dominance_Bethe_infimum} in Sec.~\ref{subsec:Bethe_Diagonal_Dominance}); however, we follow a slightly different path in this analysis. Let us consider an edge $(i,j)$ in the graph with fixed degrees $\degi$, $\degj$ of its end nodes and with an associated coupling $\Jij$, and let us assume that we vary the inverse temperature $\beta$. First of all, we observe that $\HBijsubdet$ is positive on its domain if $\beta$ approaches zero (or in other words, in the infinite temperature limit $T \to \infty$):

\begin{Lemma} \label{lemma:determinant_infinite_temperature} {\ \\}
 In the infinite temperature limit, we have for all $(\qi,\qj)$ in $(0,1) \times (0,1):$
 \begin{align} \label{eq:determinant_infinite_temperature}
 \lim \limits_{\beta \to \,  0} \big[ \HBijsubdet(\qi,\qj) \big] = \frac{1}{\degi \degj} \cdot \frac{1}{\qi \qj (1-\qi) (1-\qj)} > 0
 \end{align}
 \end{Lemma}

Lemma~\ref{lemma:determinant_infinite_temperature} characterizes the behavior of $\HBijsubdet$ in the infinite temperature limit. It also states that the determinant is positive on its domain, if the inverse temperature $\beta$ is sufficiently small (because $\HBijsubdet$ is a continuous function of $\qi$ and $\qj$). This proves the existence \footnote{We remark that $\betacrit$ might take the value $\infty$, i.e., $\FB$ might be convex for any temperature. Also, if we additionally require that $\betacrit$ is the greatest positive number (including $\infty$) satisfying the described property, then it is even unique.} of some critical threshold $\betacrit>0$ such that, for all values $0 < \beta < \betacrit$, Proposition~\ref{prop:problem_formulation_sum_decomposition} is satisfied \footnote{More precisely, Lemma~\ref{lemma:determinant_infinite_temperature} is only an edge-specific result, where we have considered the submatrix $\HBijsub$ associated to some arbitrary edge $(i,j)$ in the graph. However, the statement in the text follows immediately, if we have one such 'edge-specific' threshold $(\betacrit)^{(i,j)}$ for each edge $(i,j)$ and then identify their minimum as the critical threshold $\betacrit$.}, and therefore justifies our approach in this section.  Moverover, Proposition~\ref{prop:problem_formulation_sum_decomposition} and Lemma~\ref{lemma:determinant_infinite_temperature} prove a fundamental result:
\begin{cor} \label{cor:convexity_Bethe_high_temperature} {\ \\}
 For a sufficiently high temperature, the Bethe free energy $\FB$ is convex on the Bethe box $\Bbox$.
\end{cor}
Regarding our previous analysis, this statement is not surprising; however, to the best of our knowledge, this is the first time that this result has been proven in the literature. \\

The next question is how $\HBijsubdet$ behaves, if we make an infinitesimal change in the (inverse) temperature. Intuitively, we expect the determinant to decrease if we reduce the temperature, as the Bethe approximation often deteriorates with stronger interactions in the model (thus, the conditions in Proposition~\ref{prop:problem_formulation_sum_decomposition} would be expected to be violated at some point,~\citet{meshi2009convexifying, weller2014understanding,knoll2020JSTAT}). This is confirmed by the following Lemma, in which we prove an even stronger result, namely that $\HBijsubdet$ is a monotonically decreasing function of $\beta$:

\begin{Lemma} \label{lemma:determinant_monotonically_decreasing} {\ \\}
For all $(\qi,\qj)$ in $(0,1) \times (0,1)$, the determinant of $\HBijsub(\qi,\qj)$ is monotonically decreasing, as $\beta$ increases, i.e.,
\begin{align} \label{eq:HBij_determinant_monotonic}
 \frac{\partial}{\partial \beta}\big[\HBijsubdet(\qi,\qj)\big] < 0.
\end{align}
\end{Lemma}

\begin{figure}[]
\centering
\includegraphics[width=1\linewidth]{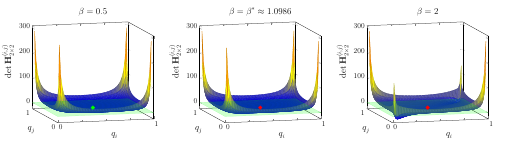}
\includegraphics[width=1\linewidth]{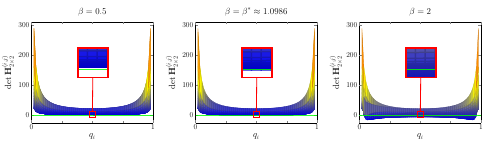}
\caption[Caption without FN]{Behavior of $\HBijsubdet$ on its domain $(0,1) \times (0,1)$ for $\degi = \degj = 3$ and $\Jij = 0.5$ (corresponding to a ferromagnetic edge whose end nodes have both three neighbors), and for varying values of the inverse temperature $\beta$. \\ 
Top row: 3D plots of $\HBijsubdet$ for increasing values of $\beta$. The green planes represent the $\qi\qj$ - plane (i.e, the zero level of the determinant). If the graph of $\HBijsubdet$ lies above this plane, then $\HBijsub$ is positive definite for the associated value of $\beta$; if the graph cuts (or touches) the plane, it is not. According to Theorem~\ref{thm:determinant_characterization_of_convexity}, $\HBijsubdet$ is positive on its domain if and only if it is positive in the center point $(0.5,0.5)$. This is, e.g., the case for $\beta=0.5$, as is indicated by the green circle in the left-hand figure. In the middle figure, the graph of $\HBijsubdet$ touches the green plane in the center point. This occurs precisely for the critical value $\betacrit$, which is the smallest $\beta$ such that $\HBijsubdet$ is not longer positive on its domain. This is because $\HBijsubdet$ takes the value zero in the center point (Lemma~\ref{lemma:determinant_HBijsub_center}), as is indicated by a red circle. Since $\HBijsubdet$ is monotonically decreasing as $\beta$ increases (Lemma~\ref{lemma:determinant_monotonically_decreasing}), it is negative in $(0.5,0.5)$ for all larger values of $\beta$. In the right-hand figure ($\beta=2$), the graph of $\HBijsubdet$ cuts the green plane in multiple points. \\
Bottom row: The same graphs as in the top row, but from a side perspective at the level of the $\qi\qj$ - plane (with the $\qi$ - axis located in front of the viewpoint and the $\qj$ - axis hidden behind). The area around the center point has been magnified, so that the described behavior is better visible\protect\footnotemark.}
\label{fig:HBijsub_determinant}
\end{figure}

Lemma~\ref{lemma:determinant_monotonically_decreasing} makes an important statement as it says the following: If the determinant of $\HBijsub$ is negative for some value of the inverse temperature $\beta$ (by which positive definiteness of $\HBijsub$ and thus the condition in Proposition~\ref{prop:problem_formulation_sum_decomposition} would be violated), then it is also negative for all larger values of $\beta$. On the other hand, we know by Lemma~\ref{lemma:determinant_infinite_temperature} that $\HBijsubdet$ is always positive if $\beta$ is sufficiently small. To complete the image, we will next show that there exists a real (and unique) value $\betacrit$, for which $\HBijsubdet$ equals zero in some point $(\qi,\qj)$ in $(0,1) \times (0,1)$, and which is consequently the smallest $\beta$ for which the determinant is not positive. Moreover, we show that the point where this scenario occurs is precisely the 'center' of the domain, i.e., the point $(0.5,0.5)$. Figure~\ref{fig:HBijsub_determinant} illustrates this behavior of the determinant. Regarding our analysis, this implies that we can explicitly compute a critical value $\betacrit$ such that Proposition~\ref{prop:problem_formulation_sum_decomposition} is either satisfied or violated. \\

 \footnotetext{Due to numerical reasons, an exact visualization of the situation for $\beta = \betacrit$ was not possible.}

Before we address the global properties of the determinant, we focus on the center point $(0.5,0.5)$. The following result provides the critical value $\betacrit$ of the inverse temperature, at which the determinant is zero in that point:

\begin{Lemma} \label{lemma:determinant_HBijsub_center} {\ \\}
 Let $\degi,\degj > 2$ and let
 \begin{align} \label{eq:critical_beta}
  \betacrit \coloneqq \frac{1}{2 |\Jij|} \arccosh\big(1 + \frac{2}{\degi \degj - \degi - \degj} \big)
 \end{align}
be the 'critical' value of the inverse temperature.
Then
\begin{align} \label{eq:determinant_critical_beta}
 \HBijsubdet(0.5,0.5)
 \begin{cases}
  \hspace{-0.3cm} & > 0 \quad \text{for} \quad \beta < \betacrit, \\
  \hspace{-0.3cm} & = 0 \quad \text{if} \quad \, \, \, \, \beta = \betacrit, \quad \text{and} \\
  \hspace{-0.3cm} & < 0  \quad \text{for} \quad \beta > \betacrit .
 \end{cases}
\end{align}
\end{Lemma}

Lemma~\ref{lemma:determinant_HBijsub_center} provides the critical threshold $\betacrit$ of the inverse temperature at which $\HBijsubdet$, evaluated in the center point $(0.5,0.5)$, changes its sign from positive to negative (i.e., at which $\HBijsubdet(0.5,0.5)$ becomes indefinite). This may not seem especially important as there might exist other points $(\qi,\qj)$ in the domain, in which the determinant is already negative for smaller values of $\beta$ than $\betacrit$ (while we are searching for the greatest value of $\beta$ at which the determinant is positive on its entire domain). However, as already indicated ahead of Lemma~\ref{lemma:determinant_HBijsub_center}, the center point is precisely the point where $\HBijsubdet$ first changes its sign if the inverse temperature steadily increases (when starting from close to zero).

\begin{thm} \label{thm:HBijsubdet_at_betacrit} {\ \\}
 Let $\degi,\degj > 2$, let $\betacrit$ be the critical threshold of the inverse temperature derived in Lemma~\ref{lemma:determinant_HBijsub_center}, and let $\HBijsubdet$ (considered as a function of $(\qi,\qj)$) implicitly depend on $\betacrit$. Then $\HBijsubdet$ has a unique root in the center point $(0.5,0.5)$ and is positive for all other points $(\qi,\qj)$ in $(0,1) \times (0,1)$.
 \end{thm}

Theorem~\ref{thm:HBijsubdet_at_betacrit} is a strong result that fully characterizes the global behavior of $\HBijsubdet$ on its domain. It says that, if the inverse temperature of the model takes the critical value $\betacrit$ (derived in Lemma~\ref{lemma:determinant_HBijsub_center}), there exists precisely one point in the domain -- the center point -- where the determinant is non-positive (while it is positive for all other points). As a consequence, the conditions for convexity of the Bethe free energy from Proposition~\ref{prop:problem_formulation_sum_decomposition} are not satisfied if $\beta = \betacrit$. If we combine this knowledge with the insights gained in Lemma~\ref{lemma:determinant_monotonically_decreasing} (the determinant monotonically decreases as $\beta$ increases), we can conclude as follows: On the one hand, $\HBijsubdet$ must be positive on the entire domain whenever the inverse temperature $\beta$ is below that critical threshold $\betacrit$; on the other hand, $\HBijsubdet$ is negative in $(0.5,0.5)$ whenever the inverse temperature $\beta$ is beyond that critical threshold $\betacrit$. Hence, the complete information about the global definiteness properties of $\HBijsub$ lies already in the center point: If $\HBijsub$ is positive definite in $(0.5,0.5)$, then it is positive definite on its entire domain. This induces a particularly simple criterion for checking the positive definiteness of $\HBijsub$ on its domain, as one simply needs to compute the critical value of the inverse temperature $\betacrit$ derived in Lemma~\ref{lemma:determinant_HBijsub_center}. If $\beta < \betacrit$, then $\HBijsub$ is positive definite; otherwise it is indefinite (because at least the main diagonal entries are always positive). Lemma~\ref{lemma:determinant_monotonically_decreasing}, Lemma~\ref{lemma:determinant_HBijsub_center}, and Theorem~\ref{thm:HBijsubdet_at_betacrit} therefore imply the following corollary:

\begin{cor} \label{cor:check_critical_beta_in_center} {\ \\}
 If $\HBijsubdet$ is positive in the center point $(0.5,0.5)$, then it is positive for all $(\qi,\qj)$ in $(0,1) \times (0,1)$. This is precisely the case, if the inverse temperature of the model lies below the critical threshold $\betacrit$. Equivalently, if $\HBijsub$ is positive definite in the center point $(0.5,0.5)$, then it is positive definite \footnote{More precisely, the second statement is true if we have defined the constrained parameters $s_{ij},s_{ji}$ in the definition of the edge-specfic Hessians~\eqref{eq:HBij_entries} as $s_{ij}=\frac{1}{\degi}, s_{ji}=\frac{1}{\degj}$; not necessarily, however, for all parameters $s_{ij},s_{ji}$ satisfying the constraints~\eqref{eq:sij_constraint_1},~\eqref{eq:sij_constraint_2}.} for all $(\qi,\qj)$ in $(0,1) \times (0,1)$.
\end{cor} 

Up to this point, we have analyzed the properties of a single edge-specific Hessian $\HBij$ in the sum decomposition~\eqref{eq:edge_specific_sum_decomposition} of the Bethe Hessian $\HB$. However, there might of course be values of the critical temperature for which certain matrices in the sum are positive definite, while others are not. Our sufficient criterion for positive definiteness of $\HB$ (and thus convexity of $\FB$) requires positive definiteness of each individual matrix in that sum. We finally arrive at the central result of this section, which is a reformulation of Proposition~\ref{prop:problem_formulation_sum_decomposition}, based on the insights that we have gained from our analysis: \newpage

 \begin{thm} \label{thm:determinant_characterization_of_convexity} {\ \\}
Let $\beta$ be the inverse model temperature. The Bethe free energy $\FB$ is convex on the Bethe box $\Bbox$ if, for all edges $(i,j)$ in the graph with degrees $\degi, \degj > 2$ of their end nodes, the following inequality is satisfied:
\begin{align} \label{eq:final_inequality_determinant_criterion}
 \beta < \frac{1}{2|\Jij|} \arccosh\Big(1 + \frac{2}{\degi \degj - \degi - \degj} \Big)
\end{align}
\end{thm}
\vspace{0.5cm}

In Sec.~\ref{subsec:Bethe_Diagonal_Dominance} and Sec.~\ref{subsec:Bethe_Hessian_Decomposition} we have derived conditions for the convexity of $\FB$ on $\Bbox$, and thus provided answers to the second of our three questions (i) - (iii) formulated in Sec.~\ref{subsec:Bethe_Properties} that were the starting point of our analysis. In the remaining part of this work, we address question (iii): Can we explain the reliability of the Bethe approximation through its convexity on the Bethe box $\Bbox$? Or in other words, is convexity of $\FB$ on $\Bbox$ a property that is necessary and/or sufficient for guaranteeing a high quality of the marginal and partition function estimates provided by the Bethe approximation (according to the approximations made in~\eqref{eq:Bethe_approx_partition} and~\eqref{eq:Bethe_approx_marginals}, Sec.~\ref{subsec:variational})? In Sec.~\ref{subsec:Theoretical_Discussion} we specify this query and discuss a few of its theoretical aspects. In Sec.~\ref{sec:Bethe_experiment} we present our experimental results.

\subsection{Theoretical Discussion} \label{subsec:Theoretical_Discussion}

The purpose of this section is to recap a few characteristics and differences of our main results from Sec.~\ref{subsec:Bethe_Diagonal_Dominance} (Theorem~\ref{thm:polynomial_characterization_of_inf_Ri}) and Sec.~\ref{subsec:Bethe_Hessian_Decomposition} (Theorem~\ref{thm:determinant_characterization_of_convexity}). Furthermore, we put them into the context of similar results from the literature (Sec.~\ref{subsec:related}), by comparing them on a perfectly symmetric model (a complete graph with homogenous potentials) whose simplicity allows for an analytical specfication of its phase transition boundaries \footnote{More precisely, this characterization is valid for the infinite version of a complete graph, which is the Cayley tree. However, we argue that at least for the case of vanishing fields (i.e., $\theta=0$) these boundaries are sharp for the finite versions as well.}~\citep{georgii1988gibbs,tagamase2006gibbs}. Then, to discuss theoretical aspects of question (iii), we analyze the behavior of the Bethe free energy in dependence of the inverse temperature, by distinguishing between its three different 'stages': convexity, non-convexity but uniqueness of a Bethe minimum, and the coexistence of multiple Bethe minima. This specification refines the conceptual view on the Bethe approximation and motivates our experimental analysis in Sec.~\ref{sec:Bethe_experiment}. \\ 

Let us once again summarize the main results from the two previous sections, both being sufficient conditions for the convexity of the Bethe free energy $\FB$ on the Bethe box $\Bbox$. According to Theorem~\ref{thm:polynomial_characterization_of_inf_Ri} (derived from diagonal dominance of the Bethe Hessian), $\FB$ is convex on $\Bbox$ if, for all nodes $i$ in the graph, the one-dimensional polynomial 
\begin{align*}
\Psii(\qi) = -(\degi-1) \prod\limits_{j \in \nbhi}(1+\aij\qi) + \sum\limits_{j \in \nbhi} \Big((1+\aij\qi^2) \prod\limits_{k \in \nbhi \setminus j}(1+\alpha_{ik}\qi) \Big)
\end{align*}
is strictly positive on the interval $(0,0.5]$. To apply this result in practice, we need to verify for all polynomials $\Psii(\qi)$ (for $i \in \setofnodes$) if none of them has a root in the interval $(0,0.5]$. This can be done efficiently by computational methods from numerical optimization \footnote{If an exact statement is desired, one can apply Sturm's theorem which calculates the total number of real roots of any one-dimensional polynomial in an arbitrary interval~\citep{thomas1941sturm}.}. The degree of $\Psii(\qi)$ is $\degi+1$ and thus increases with the number of neighbors of node $i$. This can make a numerical search for potential roots of $\Psii(\qi)$ more challenging (still, it remains a polynomial in one variable). The overall computational effort increases at least linearly with the number of nodes in the graph (since positivity must be verified for $|\setofnodes|$ polynomials). \\

Other than the above result, Theorem~\ref{thm:determinant_characterization_of_convexity} (derived from a sum decomposition of the Bethe Hessian) provides a condition in closed-form. It says that $\FB$ is convex on $\Bbox$ if, for all edges $(i,j)$ in $\setofedges$ whose end nodes $i$ and $j$ have at least three neighbors, the inverse temperature of the model satisfied the inequality
\begin{align*}
 \beta < \frac{1}{2|\Jij|} \arccosh\Big(1 + \frac{2}{\degi \degj - \degi - \degj} \Big).
\end{align*}
This condition is particularly easy to verify and additionally provides an explicit estimate of the temperature at which the Bethe free energy might become non-convex (by taking the greatest $\beta$ that simultanously satis\-fies all inequalities -- i.e., for all edges $(i,j)$ -- of the above form). Note that Theorem~\ref{thm:polynomial_characterization_of_inf_Ri} induces such an estimate only in implicit form, given as the greatest $\beta$ such that none of the polynomials $\Psii(\qi)$ has a root in $(0,0.5]$. To compute it, one can steadily increase $\beta$ and stop when the first polynomial $\Psii(\qi)$ has a root in $(0,0.5]$. The computational effort for checking Theorem~\ref{thm:determinant_characterization_of_convexity} increases linearly with the number of edges in the graph (whereby for each edge only an elementary calculation needs to be performed). \\

In summary, Theorem~\ref{thm:determinant_characterization_of_convexity} has computational advantages over Theorem~\ref{thm:polynomial_characterization_of_inf_Ri} and appears to be better interpretable. However, this does not imply that it is more powerful in the sense of inducing more accurate estimates of the critical temperature at which $\FB$ changes its stage from convex to non-convex. \\

In the sequel, we put our results in a broader context by comparing them to similar results from the literature. An overview of all important statements and their connections has been given in Fig.~\ref{fig:Relationships_between_Criteria} (Sec.~\ref{subsec:related}). First, it seems convenient to categorize them according to their purposes and properties. Subsequently, we discuss them on complete graphs with homogenous potentials. For further details, we refer to~\citet{mooij2007sufficient}.
\begin{itemize}
 \item In infinite systems, the theory of Gibbs measures interprets a phase transition as the existence of a unique Gibbs measure. The result of~\citet{dobrushin1968uniqueness} is sufficient for uniqueness of a Gibbs measure and takes the influence of the fields $\theta_i$ into account; however, its evaluation time increases exponentially with the number of nodes in the graph. The condition of \citet{simon1979remark} is sufficient for Dobrushin's condition and simpler to evaluate, but it is only based on the couplings $\Jij$ and thus weaker for models that are strongly influenced by the fields $\theta_i$.
 \item The relationship between Gibbs measure theory and approximate inference in (finite) probabilistic graphical models has been established by~\citet{tatikonda2002gibbs}. Namely, loopy belief propagation converges to a unique fixed point if there exists a unique Gibbs measure on the computation tree (which is an infinite unwrapping of the graph; in particular, Gibbs measures are well-defined on such a tree). To the best of our know\-ledge, nothing is known about the reverse direction of this statement.
 \item The results of~\citet{ihler2005message,mooij2007sufficient,watanabe2009graphzeta} are sufficient for the convergence of LBP to a unique fixed point, and thus for the uniqueness of a Bethe minimum. Their evaluation is based on closed-form expressions or eigenvalue problems, and can be done efficiently. Moreover, Mooij's criterion also takes the influence of fields $\theta_i$ into account. Experimental results indicate that Mooij's criterion is the most powerful condition \footnote{By most powerful we mean that it correctly predicts the uniqueness of an LBP fixed point for the largest variety of models, when compared to other conditions. This usually implies more accurate estimates of the temperature, at which multiple LBP fixed points (or Bethe minima) arise.} for predicting uniqueness of an LBP fixed point (or Bethe minimum). Another sufficient condition for the uniqueness of an LBP fixed point (but not convergence to it) is that of~\citet{heskes2004uniqueness}. It appears to be less powerful than other criteria, but is of theoretical importance. Its evaluation requires the solution of a linear program and only takes the couplings $\Jij$ into account.
 \item The results in this work (Theorem~\ref{thm:polynomial_characterization_of_inf_Ri} and Theorem~\ref{thm:determinant_characterization_of_convexity}) are sufficient for the convexity of the Bethe free energy $\FB$ on a submanifold $\Bbox$ (the Bethe box) of its actual domain $\polytopeLocal$ (the local polytope). As Proposition~\ref{propos:Watanabe_convexity} shows, $\FB$ can never be convex on the entire local polytope, unless the graph contains at most one cycle. However, convexity of $\FB$ on $\Bbox$ implies the uniqueness of a Bethe minimum (Cor.~\ref{cor:Unique_Bethe_Minimum}, Sec.~\ref{subsec:Bethe_Properties}). Our results are based on the couplings $\Jij$ only, because convexity of $\FB$ is a property that is not influenced by the fields $\theta_i$ (Cor.~\ref{cor:independence_of_theta}). Both conditions can be evaluated efficiently.
\end{itemize}

These results fulfill different purposes, being sufficient for convergence of LBP to a unique fixed point, uniqueness of a Gibbs measure on the computation tree, or convexi\-ty of the Bethe free energy. Still, all of them estimate a critical temperature $\betacrit$ -- either implicitly (e.g., Theorem~\ref{thm:polynomial_characterization_of_inf_Ri}) or explicitly (e.g., Theorem~\ref{thm:determinant_characterization_of_convexity}) -- at which the Bethe free energy changes its stage (from convex to non-convex, or from uniqueness of a minimum to multiple minima). \\

Ultimately, we aim to better understand the impact of the current stage of the Bethe free energy on its approximation accuracy (i.e., with respect to the marginals and partition function). For many models, the quality of the Bethe approximation significantly degrades at some critical temperature, and thus loses its reliability. This effect can be interpreted as a finite version of a 'phase transition' in the model \footnote{These are not phase transitions in the sense of infinite models, because the partition function in finite models is always a continuous function of the temperature (Sec.~\ref{subsec:related}).}. In our analysis, we aim to understand if this reliabilty loss can be attributed to a specific stage change of the Bethe free energy. \\

As we do not have an oracle to tell us in which stage $\FB$ is at a certain tempera\-ture, we must rely on the results from the literature and our work. To get a notion about their individual power, we perform an analysis on a particularly simple model whose behavior can be exactly characterized: a complete graph with homogenous potentials (i.e., each node is connected to each other with a uniform node degree $d=\degi$, and we have $\Jij=J$ for all edges and $\theta_i=\theta$ for all nodes). We discuss how powerful the above discussed results are on that simple model towards predicting a state change of the Bethe free enegry (of course, this does not imply that they are equally powerful for more complex models). \\

We first consider our condition based on diagonal dominance of the Bethe Hessian. For this model, Theorem~\ref{thm:polynomial_characterization_of_inf_Ri} says that $\FB$ is convex on $\Bbox$ if the polynomial
\begin{align*}
 \Psii(\qi) = -(d-1) (1+\aij \qi)^d + d (1+\aij\qi^2) (1+\aij\qi)^{d-1}
\end{align*}
is positive in the interval $(0,0.5]$. Note that, due to symmetries, this polynomial is the same for all nodes. Positivity of $\Psii(\qi)$  is equivalent to the positivity of
\begin{align*}
 -(d-1) (1+\aij \qi) + d (1+\aij\qi^2),
\end{align*}
which is a quadratic polynomial and has the roots $\frac{d-1}{2d} \pm \frac{1}{2d}\sqrt{(d-1)^2-\frac{4d}{\aij}}$.
If the discriminant is negative, there are only complex roots and $\Psii(\qi)$ is positive in $(0,0.5]$. This is precisely the case if $\aij < \frac{4d}{(d-1)^2}$ or, with $\aij = e^{4\beta|J|} -1$ from~\eqref{eq:xi_optimal} substituted, if
\begin{align} \label{eq:diag_dom_for_complete_graph}
\beta < \frac{1}{4|J|} \log \Big(\frac{(d+1)^2}{(d-1)^2}\Big).
\end{align}
If the discriminant is zero, $\Psii(\qi)$ has the unique root $\frac{d-1}{2d}$ which (for $d\geq 2$) is a number in $(0,0.5]$ (i.e., $\Psii(\qi)$ is not longer positive in $(0,0.5]$). If the discriminant is positive, we have two real roots. We claim that the lower root is always in $(0,0.5]$ and thus the diagonal dominance criterion is violated for all values of $\beta$ that are greater or equal to the right-hand side of~\eqref{eq:diag_dom_for_complete_graph}. This follows from the observation that the value of the lower root monotonically decreases, if $\beta$ (and thus $\aij$ and the square root term) increases. In the limit $\beta \to \infty$, the lower root approaches zero and is consequently always in $(0,\frac{d-1}{2d})$, which is a subset of $(0,0.5]$. We conclude that the diagonal dominance criterion is satisfied, if and only if $\beta$ satisfies the inequality~\eqref{eq:diag_dom_for_complete_graph}. \\

\begin{figure}[ht]
\centering
\includegraphics[width=0.56\linewidth]{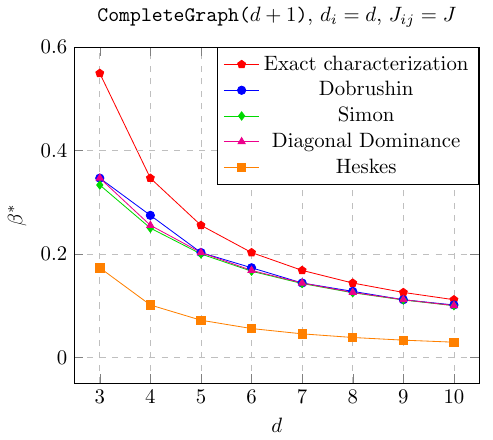}
\caption{Comparison of critical values of the inverse temperature predicted by different criteria from the literature on a complete graph with homogenous potentials and vanishing fields ($\Jij = J$ for all edges, $\theta_i=0$ and $\degi = d$ for all nodes), in dependence of the degree. }
\label{fig:critical_beta}
\end{figure}

\begin{figure}[]
\centering
\includegraphics[width=0.95\linewidth]{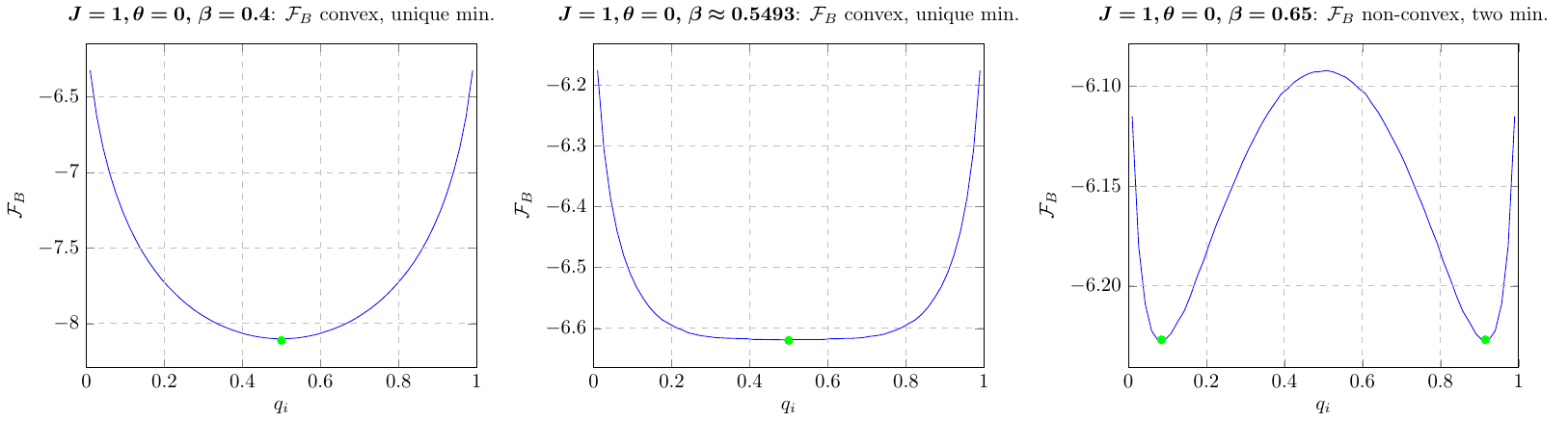}
\includegraphics[width=0.95\linewidth]{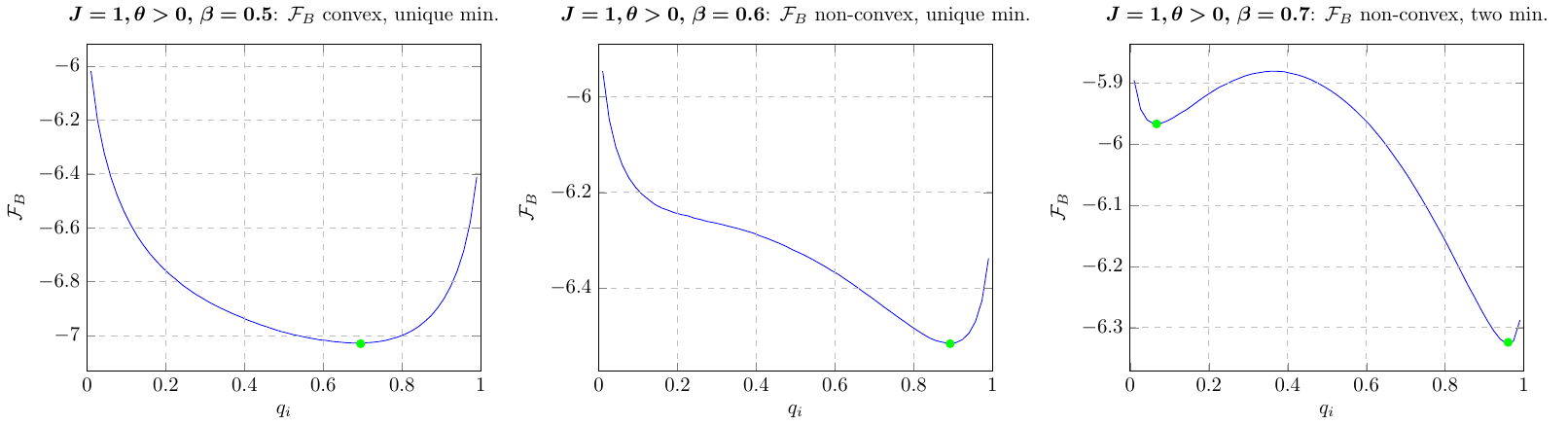}
\includegraphics[width=0.95\linewidth]{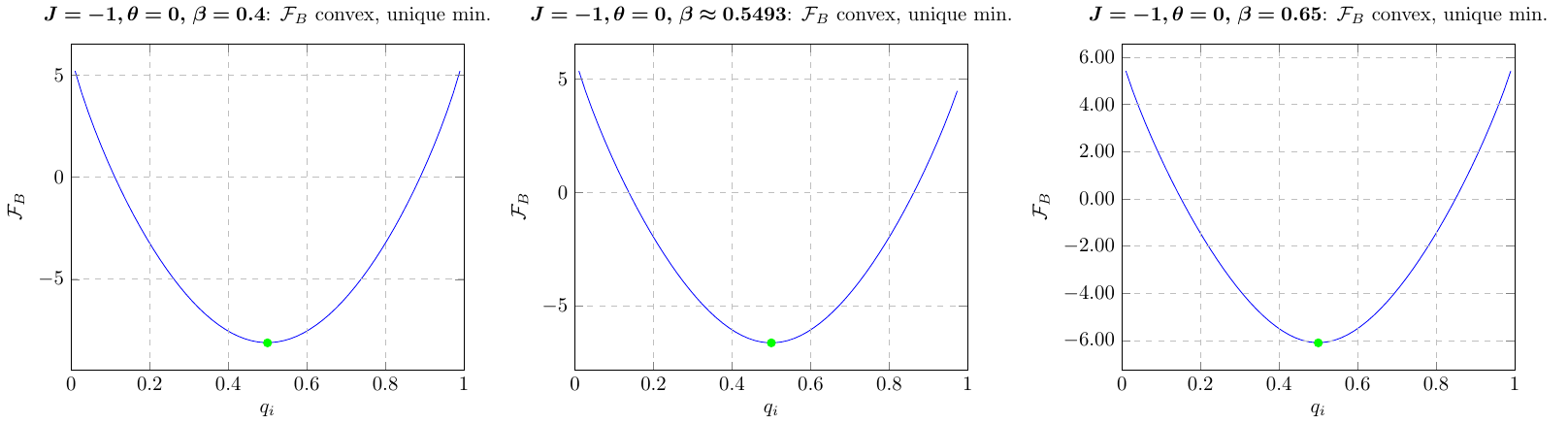}
\caption{Stage change behavior of the Bethe free energy in dependence of the inverse temperature $\beta$ and fields $\theta$ on a complete graph with $4$ nodes and homogenous potentials (i.e., $\Jij=J$ for all edges and $\theta_i=\theta$ for all nodes). These plots represent a projection of $\FB$ onto the main diagonal of $\Bbox$, i.e, points $\qvec=(q_1,q_2,q_3,q_4)$ with $q_1=\dots=q_4$. \\
First row: For $\theta=0$ and $J>0$ (a ferromagnetic model without fields), $\FB$ is convex and has a unique minimum in $\qi=0.5$ if $\beta$ is below a critical threshold of $\betacrit \approx 0.5493$. If $\beta$ increases beyond that threshold, $\FB$ becomes non-convex and at the same time two new minima arise (while the 'old' minimum disappears). \\
Second row: For $\theta >0$ and $J>0$ (a ferromagnetic model with fields), $\FB$ is convex and has a unique minimum if $\beta$ is sufficiently small. The location of the minimum is biased by the fields and thus not in $\qi=0.5$. If $\beta$ increases, $\FB$ reaches a point where it becomes non-convex but still has a unique minimum. If $\beta$ further increases, there arises a second Bethe minimum (with the up-to-then minimum still existing). This shows that the fields can induce a gap between convexity of $\FB$ and uniqueness of a Bethe minimum. \\
Third row: For $\theta=0$ and $J<0$ (an antiferromagnetic model without fields), $\FB$ preserves its convexity (and its unique minimum) at any temperature.}
\label{fig:Bethe_convexity_states}
\end{figure}

Further, we consider our second result based on the sum decomposition of the Bethe Hessian on the symmetric model. For all edges the inequalities in Theorem~\ref{thm:determinant_characterization_of_convexity} are identical and $\FB$ is convex on $\Bbox$ if
\begin{align*}
 \beta < \frac{1}{2|J|} \arccosh\Big(1 + \frac{2}{d (d-2)} \Big)
\end{align*}
is satisfied (for $d \geq 3$). We can use the trigonometric identity
\begin{align*}
 \arccosh(x) = \log\big(x + \sqrt{x-1} \sqrt{x+1} \big)
\end{align*}
and perform a few simplifications to show that this is equivalent to
\begin{align} \label{eq:threshold_sum_decomposition_complete_graph}
 \beta < \frac{1}{2|J|} \log\Big(\frac{d}{d-2} \Big).
\end{align}
\\ 
As already mentioned, the exact behavior for the symmetric model can be characterized. More precisely, for $\theta=0$, it has been shown that Gibbs measures become non-unique~\citep{georgii1988gibbs,tagamase2006gibbs} and LBP stops converging to a unique fixed point~\citep{mooij2005bethe}, if the inverse temperature passes the critical threshold of
\begin{align}
 \frac{1}{|J|} \arctanh\Big(\frac{1}{d-1} \Big).
\end{align}
For $d \geq 3$, this expression is the same as the critical threshold in inequality~\eqref{eq:threshold_sum_decomposition_complete_graph}. This follows  directly from the trigonometric identity
\begin{align*}
 \arctanh(x) = \frac{1}{2} \big(\log(1+x) - \log(1-x) \big).
\end{align*}
Thus, our result in Theorem~\ref{thm:determinant_characterization_of_convexity} is sharp in the sense, that it correctly predicts the value of $\beta$ at which a modified behavior occurs \footnote{More precisely, Gibbs measures become non-unique and LBP stops converging to a unique fixed point. However, $\FB$ might still be convex and have a unique minimum if the inverse temperature increases beyond the predicted thresholds. This can be observed in an antiferromagnetic model ($J<0$, Fig.~\ref{fig:Bethe_convexity_states}).}. Other results that are sharp on the symmetric model without fields are those of~\citet{ihler2005message},~\citet{mooij2007sufficient}, and~\citet{watanabe2009graphzeta}. Table~\ref{tab:critical_beta} and Fig.~\ref{fig:critical_beta} show a comparison between the critical thresholds that are predicted by the various criteria from the literature on the symmetric model without fields. If $\theta \neq 0$, none of these conditions are sharp any longer (with the results of Dobrushin and Mooij tending to more accurate predictions, as they consider the influence of the fields).

\begin{table}[h]
\centering
\caption{Estimates of a critical inverse temperature, predicted by various results from the literature on a complete graph with homogenous potentials and vanishing fields (i.e., $\Jij = J$ for all edges, $\theta_i=0$ and $\degi = d$ for all nodes). The conditions in the first row are sharp on this model (i.e., they correctly predict $\betacrit$ at which the model changes its behavior).} \label{tab:critical_beta}
\begin{tabular}{ |c|c|c| }

\hline
\\[-1em]
 Description & Reference & \normalsize $\betacrit$ \\ [4.5pt] \hline \hline
 \\[-1em]
 Exact characterization &  \citet{georgii1988gibbs},\citet{ihler2005message} & \normalsize $\frac{1}{|J|} \arctanh\Big(\frac{1}{d-1}\Big)$ \\ [4.5pt]
  & \citet{mooij2007sufficient},    &  \\ [4.5pt]
  & Theorem~\ref{thm:determinant_characterization_of_convexity} (this work) & \\ [1.5pt]
  \\[-1em] \hline
 Dobrushin & \citet{dobrushin1968uniqueness} & \normalsize $d$ odd: $\frac{1}{|J|} \arctanh\Big(\frac{1}{d}\Big)$ \\ [4.5pt]
 &  & \normalsize $d$ even: $\frac{1}{2|J|} \arctanh\Big(\frac{2}{d}\Big)$ \\ [4.5pt]  \hline
 \\[-1em]
 Simon & \citet{simon1979remark} & \normalsize $\frac{1} {|J| \, d}$ \\ [4.5pt] \hline 
 \\[-1em]
 Diagonal Dominance & Theorem~\ref{thm:polynomial_characterization_of_inf_Ri} (this work) & \normalsize $\frac{1}{4|J|}\log\Big(\frac{(d+1)^2}{(d-1)^2}\Big)$ \\ [4.5pt] \hline
 \\[-1em]
 Heskes & \citet{heskes2004uniqueness} & \normalsize $\frac{1}{4|J|} \log\Big(\frac{d-1}{d-2}\Big)$ \\ [4.5pt] \hline
 \end{tabular}
\end{table}

 After having compared various results and their power with respect to predicting properties of a model, we briefly discuss a few aspects of these properties themselves; specifically, we focus on the different stages of the Bethe free energy. As convexity of $\FB$ on the Bethe box $\Bbox$ implies uniqueness of a Bethe minimum (but not vice versa), there is generally a certain 'temperature gap' between these two properties. Any condition that predicts the convexity of $\FB$ cannot explain this gap, as it is caused by the presence of fields $\theta_i$ and increases with stronger fields. The fields can affect the location (and number) of Bethe minima, but not the curvature of $\FB$. If there are no fields, convexity of $\FB$ and uniqueness of a Bethe minimum are supposed to be equivalent. Although we do not have a formal proof of this statement, we would like to formulate it as a conjecture:
\begin{conj} \label{conj:equivalence_convexity_uniqueness}  {\ \\}
 Assume we have no external fields in our model (i.e., $\theta_i=0$ for all nodes $i$). Then $\FB$ is convex on $\Bbox$, if and only if it has a unique minimum.
\end{conj}
Figure~\ref{fig:Bethe_convexity_states} schematically illustrates the state change behavior of the Bethe free energy in dependence of the inverse temperature and the presence (or absence) of fields on a perfectly symmetric model. A detailed interpretation is given in the figure caption. Interestingly, $\FB$ preserves its convexity (and thus its unique minimum) in the antiferromagnetic model for any inverse temperature. We believe that this phenomenon is due to frustrations in the model, which are cycles in the graph with an odd number of antiferromagnetic edges~\citep{toulouse1977frustrations}. While frustrations can induce multiple Gibbs measures and prevent loopy belief propagation from converging~\citep{georgii1988gibbs,knoll2017stability,mooij2005bethe}, they may have 'stabilizing' effects on the convexity (and reliability) of the Bethe free energy. Without providing further details on this problem, we would like to formulate another conjecture:
\begin{conj} \label{conj:frustrated_cycles}  {\ \\}
 Frustrations help the Bethe free energy to preserve its convexity.
\end{conj}

The considerations made in this section are instructive and aim to explain the behavior of the Bethe free energy in dependence of the inverse temperature. We primarily analyzed special kind of models to understand its different 'stages', and discussed several criteria to predict changes between them. In the next and final step of our analysis, we put our focus on the quality of the Bethe approximation during these stages, regarding the accuracy of the estimates for the marginals and partition function. We will address these topics in Sec.~\ref{sec:Bethe_experiment} and present our experimental results on a broad variety of models.

\section{Convexity of the Bethe Free Energy: Experiments} \label{sec:Bethe_experiment}

In Sec.~\ref{subsec:Theoretical_Discussion}, we have exposed how the Bethe free energy passes through different stages if the inverse model temperature increases, and discussed possibilities to predict in which stage it currently is. We now focus on analyzing the reliability of the Bethe approximation, i.e., the accuracy of the marginal and partition function estimates that it provides. \\

The quality of the Bethe approximation has been analyzed, e.g., by~\citet{meshi2009convexifying,weller2014understanding,knoll2020JSTAT}, but rather at a qualitative level and without specifying its different stages. We extend this analysis by a quantitative component (i.e., concrete estimates for critical phase transition temperatures or, in other words, reliability thresholds), and the concept of convexity on the Bethe box $\Bbox$ that we have introduced in this work. Specifically, we analyze if the Bethe approximation is always reliable if $\FB$ is convex on $\Bbox$; if it is still reliable if it is non-convex but has a unique minimum; or if its current stage has no significant influence on the quality of the Bethe approximation. \\

In Sec.~\ref{subsec:Bethe_min_algo}, we show how we measure the quality of the Bethe free energy approximation and introduce our algorithm \texttt{BETHE-MIN} that we use for its minimization. In Sec.~\ref{subsec:experiments_ferro_general}, we explain our experimental setup and present all experimental results in detail.

\subsection{Minimization of the Bethe Free Energy} \label{subsec:Bethe_min_algo}
Let us recap the essential ideas of the Bethe approximation (introduced in Sec.~\ref{subsec:variational}) to clarify the concept of the subsequent experiments. In a finite graphical model, we intend to find approximations to the (generally non-accessible) partition function~\eqref{eq:partition_function}, and/or marginal distributions~\eqref{eq:singleton_marginals},~\eqref{eq:pairwise_marginals}. Arising from the Gibbs variational principle, the Bethe approximation provides estimates for these unknown quantities by making the approximations~\eqref{eq:Bethe_approx_partition} and~\eqref{eq:Bethe_approx_marginals}, i.e., by using its minimum on the set $\polytopeLocal$ of local probability constraints, the local polytope~\eqref{eq:local_polytope_reparameterized}. As we have shown in Sec.~\ref{subsec:Bethe_Properties}, we can reduce the dimension of this domain and instead minimize the Bethe free energy $\FB$ on a submanifold $\Bbox$ of $\polytopeLocal$, the Bethe box~\eqref{eq:Bethe_box}. In other words, to approximate the partition function and singleton marginals \footnote{More precisely, $\qi$ (the i-th component of $\qvec$) approximates the first entry of the marginal $p_i$, i.e., the probability that variable $X_i$ takes the value $+1$. The second entry (the probability that $X_i$ takes the value $-1$) can be estimated with $1-\qi$ (Table~\ref{tab:prob_table}).} we use
\begin{align}
	-\frac{1}{\beta} \log Z_{B} \coloneqq \min\limits_{\tilde{\qvec} \, \in \, \Bbox} \FB \, \, & \approx \, \, - \frac{1}{\beta} \log Z(\beta) \, , \label{eq:Bethe_approx_partition_rewritten} \\ 
	\{ \, q_i: \qvec =  \argmin\limits_{\tilde{\qvec} \in \Bbox}  \FB \, \, | \, \, i \in \setofnodes \,  \} \, \, & \approx \, \, \{ \, p_i \, \, | \, \, i \in \setofnodes \,   \}. \label{eq:Bethe_approx_marginals_rewritten}
\end{align}
The approximations to the pairwise marginals $\{ \, p_{ij} \, \, | \, \, (i,j) \in \setofedges \, \}$ can directly be computed from the elements in the set on the left-hand side of~\eqref{eq:Bethe_approx_marginals_rewritten}, according to the probability table in Sec.~\ref{subsec:Bethe_Properties} (Table \ref{tab:prob_table}), where $\xijopt(\qi,\qj)$ is computed from~\eqref{eq:xi_optimal} as a direct function of $\qi$ and $\qj$. \\

To analyze the quality of the Bethe approximation, we compare the exact partition function $Z(\beta)$, singleton marginals $p_i(X_i=\pm1)$, and pairwise marginals $p_{ij}(X_i=\pm1,X_j=\pm1)$ to the estimated quantities $Z_{B}$, $(\qi,1-\qi)$, and $(\xijopt, \qi-\xijopt, \qj-\xijopt,1+\xijopt - \qi - \qj)$, for all nodes $i$ and all edges $(i,j)$). For that purpose, we use the following error measures: 
\begin{itemize}
 \item For the partition function, we evaluate the absolute error between the negative logarithms of $Z(\beta)$ and $Z_{B}$, i.e., 
\begin{align} \label{eq:error_measure_partition}
 | \! - \log Z(\beta) + \log Z_{B} |.
\end{align}
\item For the singleton marginals, we evaluate the average (with respect to all nodes) $l^1$-error between the exact and estimated quantities, i.e.,
\begin{align} \label{eq:error_measure_sing_marg}
\begin{split}
 & \frac{1}{|\setofnodes|} \sum\limits_{i \in \setofnodes} \big( |p_i(X_i=+1) - \qi| + |\overbrace{p_i(X_i=-1)}^{= 1 \, - \, p_i(X_i=+1)} - (1-\qi)| \big) \\
 = \, &  \frac{2}{|\setofnodes|} \sum\limits_{i \in \setofnodes} \big( | p_i(X_i=+1) - \qi| \big).
 \end{split}
\end{align}
\item For the pairwise marginals, we evaluate the average (with respect to all edges) $l^1$-error between the exact and estimated quantities, i.e.,
\begin{align} \label{eq:error_measure_pw_marg}
\begin{split}
 \frac{1}{|\setofedges|} \sum\limits_{(i,j) \in \setofedges} \big( & |\, p_{ij}(X_i=+1,X_j=+1) - \xijopt \, | \\ 
 + & |\, p_{ij}(X_i=+1,X_j=-1) - (\qi - \xijopt) \, | \\
+ & |\, p_{ij}(X_i=-1,X_j=+1) - (\qj - \xijopt) \, | \\ 
+ &  |\, p_{ij}(X_i=-1,X_j=-1) - (1 + \xijopt - \qi - \qj) \, | \big).
 \end{split}
\end{align}
\end{itemize}

As we do generally not have access to the partition function and marginals (that we are actually trying to approximate), we evaluate the above error measures by performing our experiments in Sec.~\ref{subsec:experiments_ferro_general} on graphs that are small enough to allow for an exact computation of these quantities. For that purpose, we use the so-called junction tree algorithm~\citep{lauritzen1988junctiontree} that returns the correct solution, but whose computational complexity increases exponentially with the size of the largest clique \footnote{A clique is a subgraph where each node is connected to each other.} in the graph. \\

As discussed in previous sections, the errors induced by the Bethe approximation strongly depend on the model parameters (in particular the graph connectivity, potentials, and inverse temperature). We leave this problem to the experimental analysis in Sec.~\ref{subsec:experiments_ferro_general} as, we first need to address another important issue: the practical minimization of the Bethe free energy. Following the approxi\-mations~\eqref{eq:Bethe_approx_partition_rewritten} and~\eqref{eq:Bethe_approx_marginals_rewritten}, we need to compute its global minimum in the constrained set $\Bbox$. This can be a challenging task in practice, especially if we deal with large systems and numerical difficulties. Another prohibitive aspect is the potential existence of multiple local minima. In that case, there is no general technique that efficiently returns the global minimum, and we must usually rely on the quality of a (possibly sub-optimal) local minimum \footnote{For ferromagnetic models, \citet{weller2014a} have proposed an -- at least, in theory -- efficient approximation scheme to compute the partition function up to some $\epsilon$ in polynomial time (an FPTAS).}. Besides, there is no guarantee that the global minimum, if we manage to find it, induces a low error on the partition function or marginals. Finally, $\FB$ may exhibit particularly flat or steep regimes, implying extremely small or exploding gradients and slow convergence to a minimum (even if $\FB$ is convex on $\Bbox$). \\

For an efficient minimization of the Bethe free energy, various methods \footnote{Note that LBP can be seen as a minimization of $\FB$ too~\citep{heskes2003stable}. However, its convergence rate is often not satisfying -- still, several works have improved on it over time~\citep{aksenov2020relaxed,knoll2015scheduling,sutton2007dynamic}.} have been proposed: 
gradient-based algorithms combined with projection steps~\citep{welling2001belief,shin2012complexity}; a double-loop algorithm that decomposes $\FB$ into a convex and a concave part and iterates between these two components~\citep{yuille2002CCCP};
combinatorial optimization~\citep{weller2014a}; and convex optimization~\citep{weller2014understanding}. A drawback of these methods is that they do not exploit second-order properties of the Bethe free energy. \\ 

We propose \texttt{BETHE-MIN}, a projected quasi-Newton method that not only uses the information of the gradient but also of the Bethe Hessian. While parameter updates in the 'ordinary' Newton method require a costly inversion of the Hessian matrix, a quasi-Newton method uses an approximation to the inverse Hessian that is updated in each iteration as well. More precisely, if $\HB(\qvec^{(t)})$ is the Bethe Hessian evaluated in the current parameter vector $\qvec^{(t)}$ at time $t$ and $\textbf{B}^{(t)}$ is an approximation to its inverse, then we perform the update
\begin{align} \label{eq:quasi_newton_update}
\qvec^{(t+1)} = \qvec^{(t)} - \rho^{(t)} \cdot (\textbf{B}^{(t)})^T \, \nabla \FB(\qvec^{(t)}),
\end{align}
where $\rho^{(t)} > 0$ is the step size of the current iteration. 
To ensure that the next parameter vector $\qvec^{(t+1)}$ remains feasible (i.e., a point in the Bethe box $\Bbox$), we project it back into $\Bbox$ if $\rho^{(t)}$ is too large. More precisely, we reduce \footnote{Alternatively, one could use an orthogonal projection of $\qi^{(t+1)}$ into $\Bbox$ as, e.g., in \citet{kim2010quasinewton}. However, this might entail a significant change of the search direction.} $\rho^{(t)}$ until any component $\qi^{(t+1)}$ is a number in $(0,1)$ so that $\qvec^{(t+1)}$ lies in $\Bbox$. For faster convergence, we then optimize the choice of $\rho^{(t)}$ by applying a line search technique with respect to the so-called Wolfe conditions~\citep{wolfe1969ascent}. After each update of $\qvec^{(t)}$, we also need to update $\textbf{B}^{(t)}$. For that purpose, we use the relatively robust BFGS update rules for approximating the inverse Hessian~\citep{nocedal2006numerical}. The described steps are iterated, until we arrive at a stationary point $\qvec^{\ast}$ of $\FB$. All details on \texttt{BETHE-MIN} including pseudocode are provided in Appendix~\ref{sec:appendix_C}.

\subsection{Experimental Results} \label{subsec:experiments_ferro_general}

Through our experiments, we aim to evaluate the quality of the Bethe approximation in dependence of its stage (i.e.: convexity; non-convexity but uniqueness of a minimum; and multiple minima). The Bethe free energy is convex on the Bethe box $\Bbox$, if the inverse temperature is sufficiently small (Lemma~\ref{lemma:determinant_infinite_temperature}, Sec.~\ref{subsec:Bethe_Hessian_Decomposition}). Often $\FB$ first becomes non-convex and later develops multiple minima if $\beta$ increases; however, sometimes $\FB$ does not change its stage and remains convex at any temperature (e.g., in the antiferromagnetic model, Sec.~\ref{subsec:Theoretical_Discussion}). To predict in which stage $\FB$ currently is, we use some of the results discussed in Sec.~\ref{subsec:related} and Sec.~\ref{subsec:Theoretical_Discussion}. In particular, to predict convexity, we use our result derived from diagonal dominance \footnote{We observed in our experiments that the diagonal dominance criterion often provides slightly more accurate estimates of the critical $\beta$ at which $\FB$ becomes non-convex, than the sum decomposition criterion (Sec.~\ref{subsec:Bethe_Hessian_Decomposition}, Theorem~\ref{thm:determinant_characterization_of_convexity}). This is somewhat surprising, as Theorem~\ref{thm:determinant_characterization_of_convexity} is exact on the perfectly symmetric model without fields, while Theorem~\ref{thm:determinant_characterization_of_convexity} is not (Sec.~\ref{subsec:Theoretical_Discussion}).} of the Bethe Hessian (Sec.~\ref{subsec:Bethe_Diagonal_Dominance}, Theorem~\ref{thm:polynomial_characterization_of_inf_Ri}), and to predict uniqueness of a minimum, we use the result of~\citet{mooij2007sufficient} \footnote{More precisely, we use Corollary 4 from~\citet{mooij2007sufficient} with the specific choice of $m=5$.}. Since all of these conditions are only sufficient (but not necessary), we expect that the true thresholds of $\beta$ at which $\FB$ changes its stage are somewhat underestimated by the predicted values. \\

\begin{figure}[]
\centering
\subfloat{\includegraphics[width=0.9\linewidth]{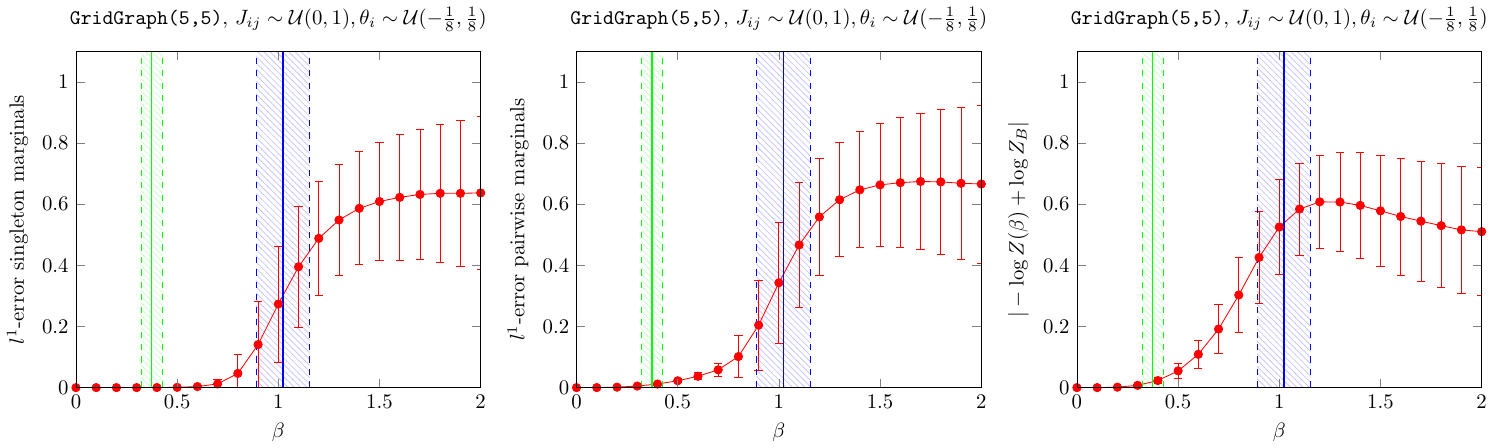}} \\
\subfloat{\includegraphics[width=0.9\linewidth]{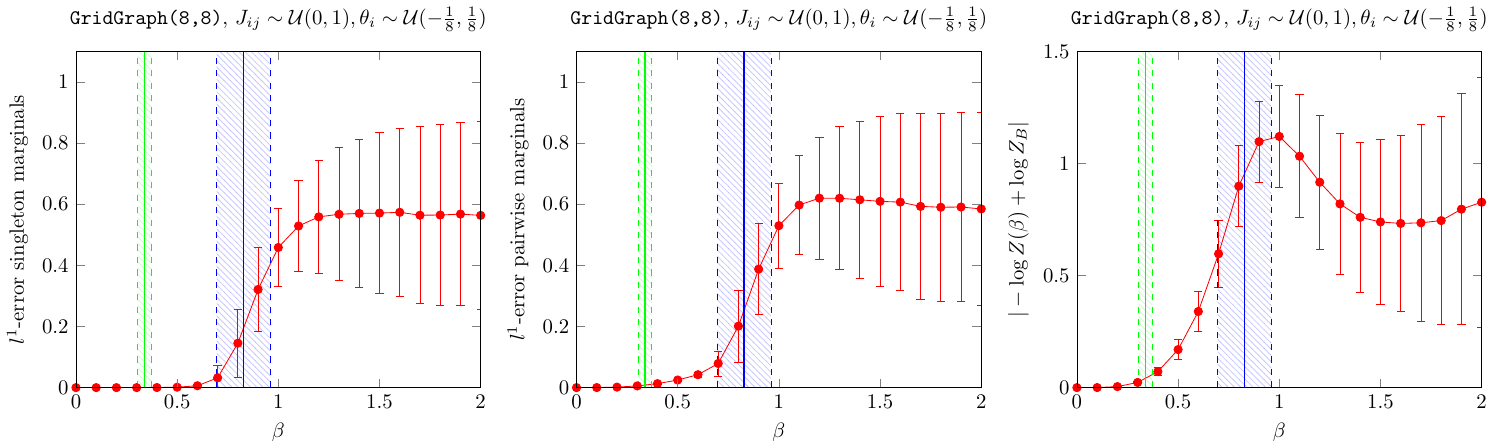}} \\
\subfloat{\includegraphics[width=0.915\linewidth]{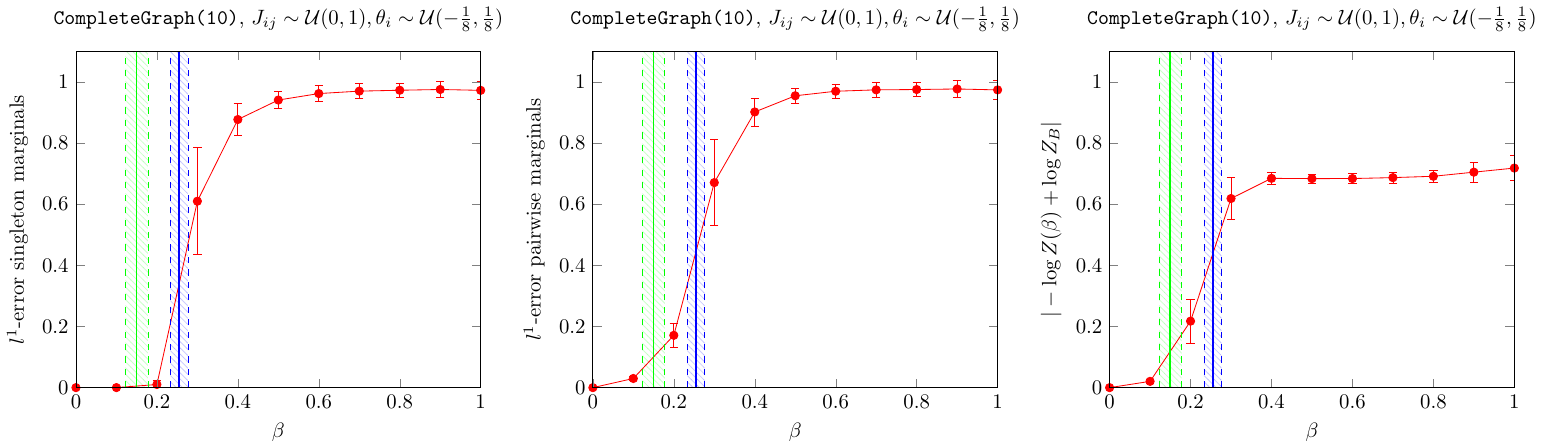}} \\
\subfloat{\includegraphics[width=0.92\linewidth]{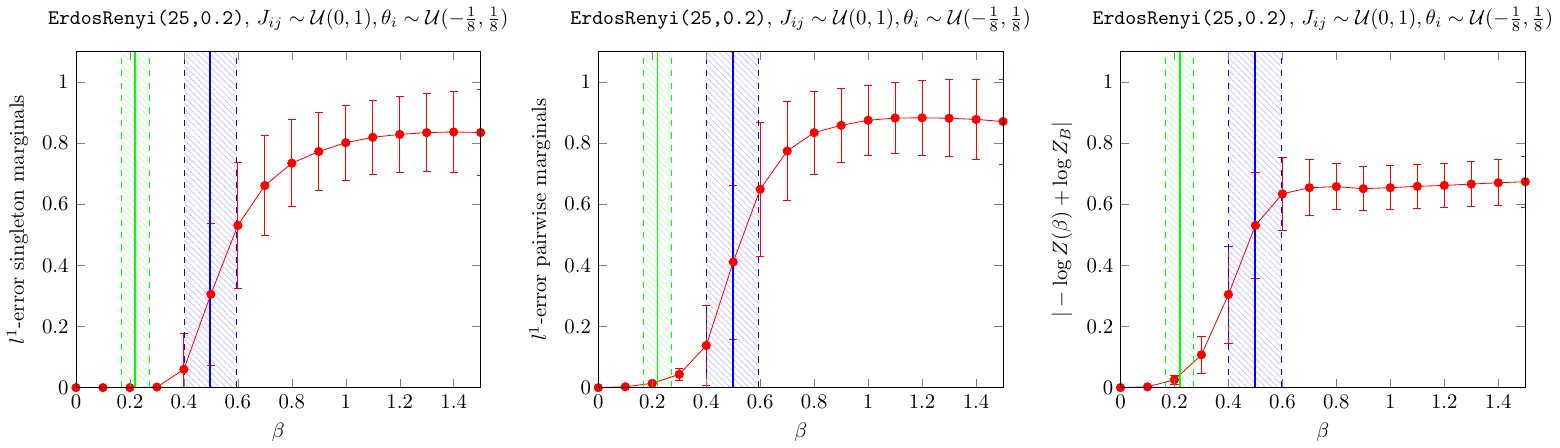}}
\caption{Ferromagnetic models. Errors (drawn in red) induced by the Bethe approximation on four different graph types, in dependence on the inverse temperature $\beta$. The critical point at which an increasing error becomes significant can be interpreted as a phase transition in the model. The green solid lines estimate the stage changes of $\FB$ from convex to non-convex; the blue solid lines estimate the stage changes of $\FB$ from a unique minimum to multiple minima. The dashed lines and error bars represent the corresponding standard deviations with respect to $200$ models, respectively. First column: singleton marginal error; second column: pairwise marginal error; third column: partition function error.}
\label{fig:phase_transitions_attractive}
\end{figure}

\begin{figure}[]
\centering
\subfloat{\includegraphics[width=0.9\linewidth]{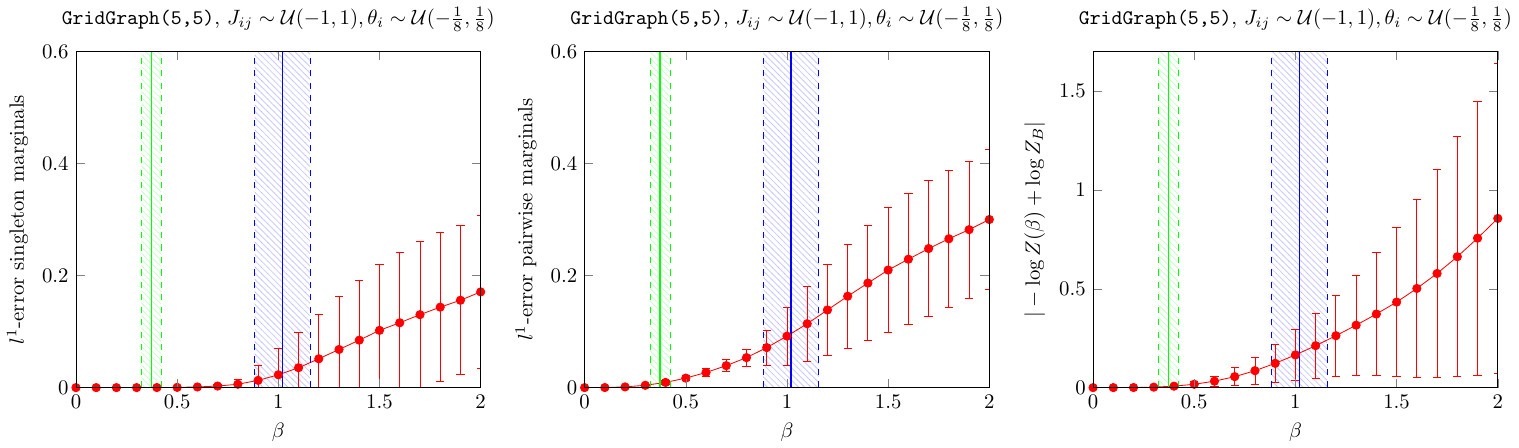}} \\
\subfloat{\includegraphics[width=0.9\linewidth]{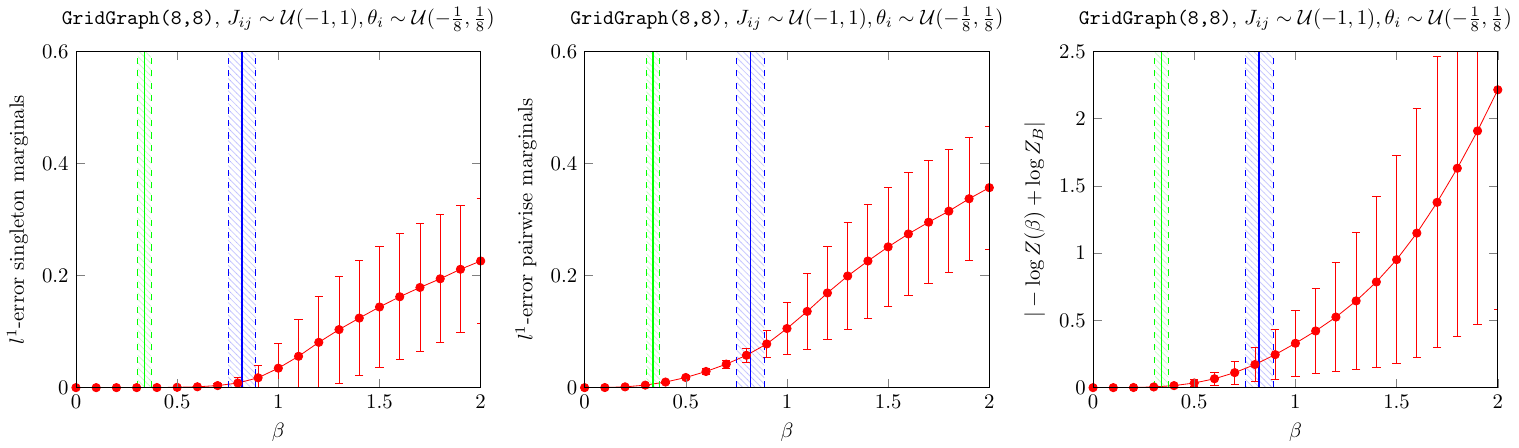}} \\
\subfloat{\includegraphics[width=0.915\linewidth]{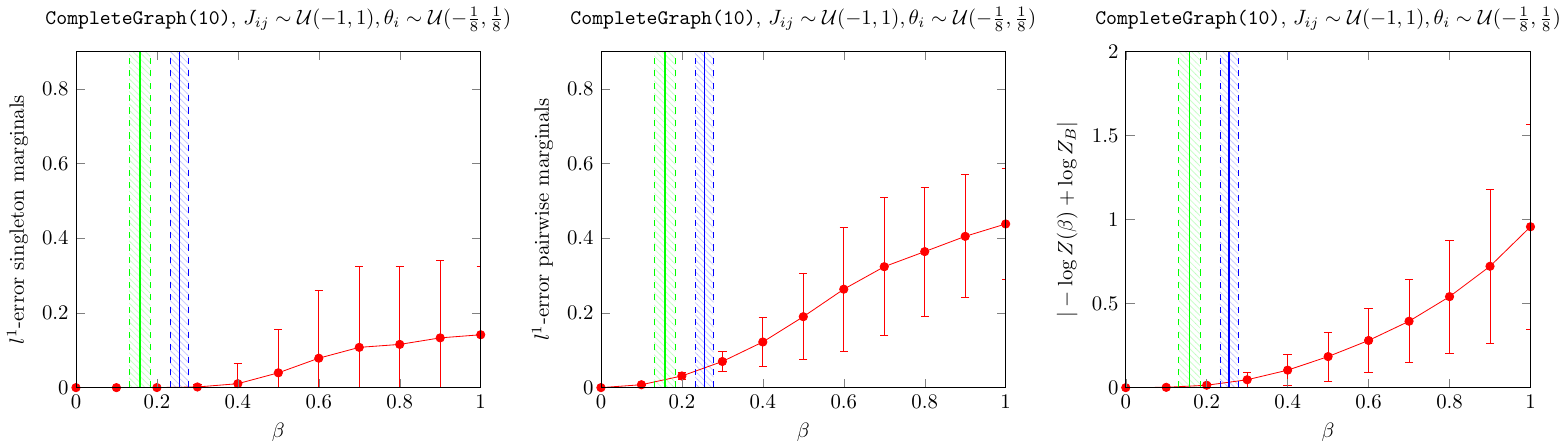}} \\
\subfloat{\includegraphics[width=0.92\linewidth]{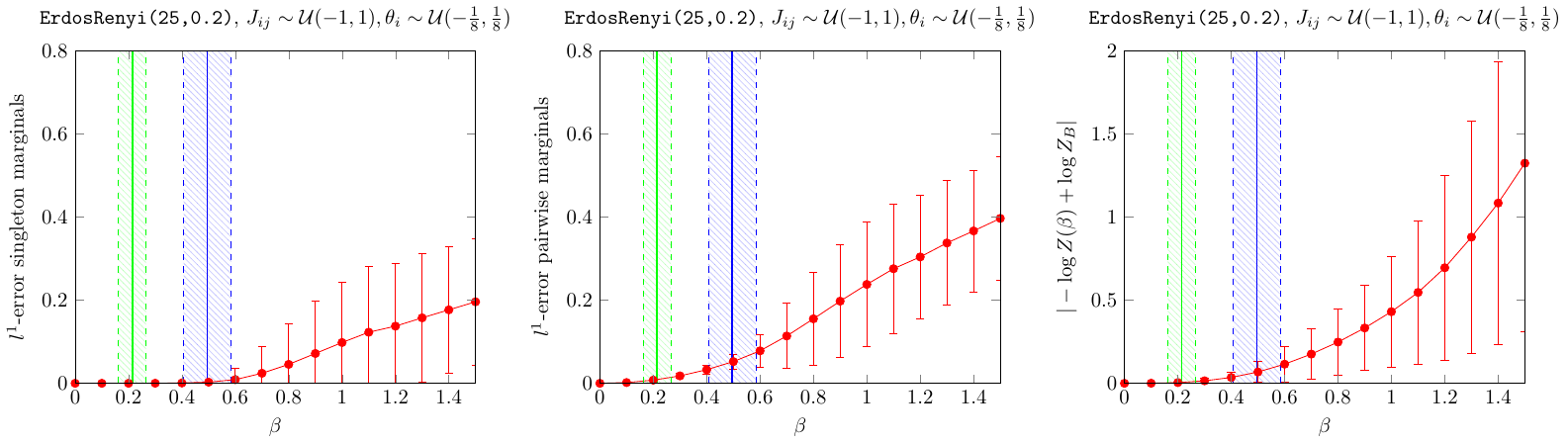}}
\caption{Spin glasses. Errors (drawn in red)  induced by the Bethe approximation on four different graph types, in dependence on the inverse temperature $\beta$. The critical point at which an increasing error becomes significant can be interpreted as a phase transition in the model. The green solid lines estimate the stage changes of $\FB$ from convex to non-convex; the blue solid lines estimate the stage changes of $\FB$ from a unique minimum to multiple minima. The dashed lines and error bars represent the corresponding standard deviations with respect to $200$ models, respectively. First column: singleton marginal error; second column: pairwise marginal error; third column: partition function error.}
\label{fig:phase_transitions_general}
\end{figure}

\begin{figure}[t]
\centering
\subfloat{\includegraphics[width=1\linewidth]{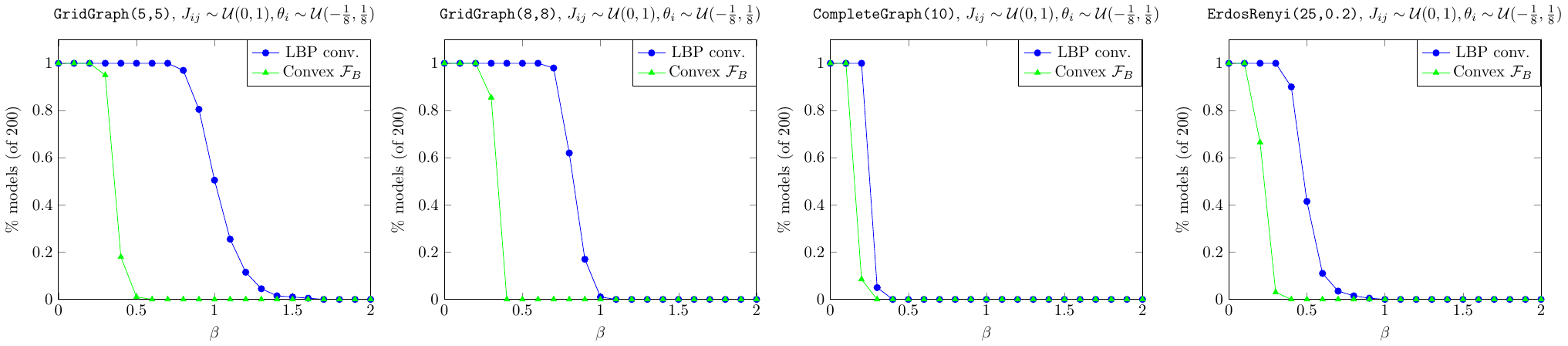}} \\
\subfloat{\includegraphics[width=1\linewidth]{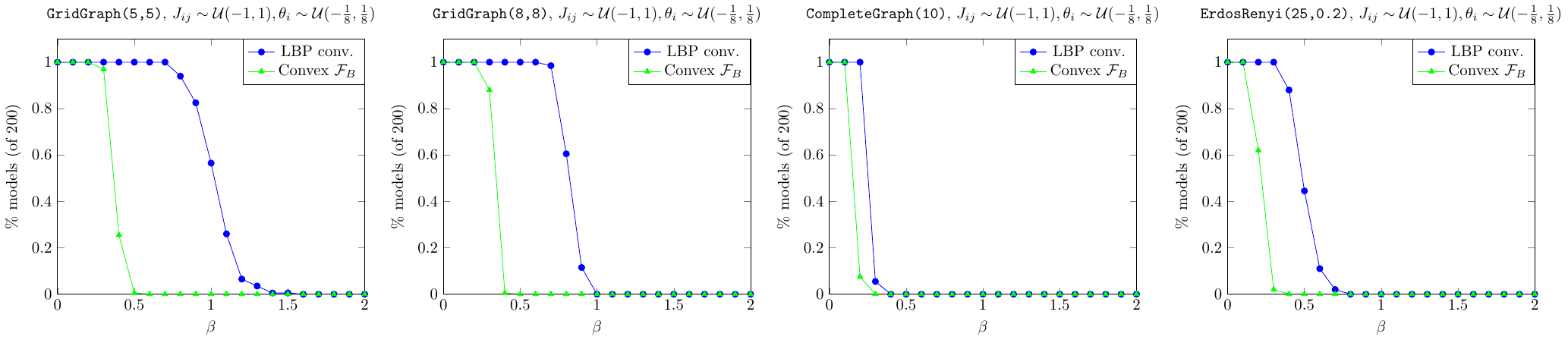}} \\
\caption{Percentage of models on which $\FB$ is predicted to be convex -- according to Theorem~\ref{thm:polynomial_characterization_of_inf_Ri} -- and to have a unique minimum (or, more precisely, convergence of LBP to a unique minimum) -- according to Corollary 4 in~\citet{mooij2007sufficient}. On graphs with a higher connectivity $\FB$ induces a smaller gap between these two properties. In the top row we consider ferromagnetic models, in the bottom row we consider spin glasses.}
\label{fig:percentage_convexity_vs_uniqueFP}
\end{figure}

\begin{figure}[t]
\centering
\includegraphics[width=0.5\linewidth]{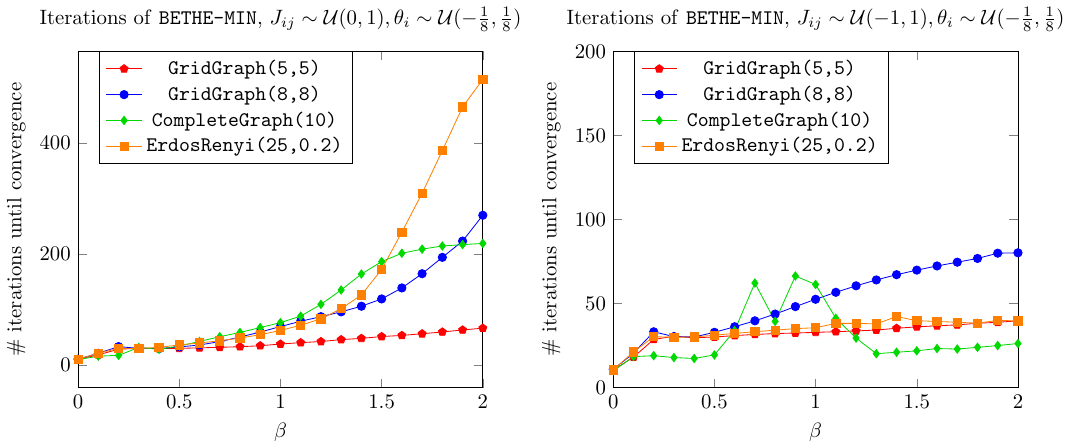}
\caption{Required iteration number until \texttt{BETHE-MIN} (Algorithm~\ref{algo:projected_quasi_Newton}) converges to a statio\-nary point of $\FB$. For ferromagnetic models (left-hand side), the iteration number increases as $\beta$ increases, with the steepest decrease for the Erdos-Renyi model. For spin glasses (right-hand side), the iteration number keeps at a lower level than for ferromagnetic models.}
\label{fig:iterations_of_BetheMin}
\end{figure}

Our experimental setup is as follows: We consider four different types of graphs: two lattice graphs (or square grid graphs) of size $5 \times 5$ and $8 \times 8$; a complete graph on $10$ nodes; and random graphs on $25$ nodes and an edge probability of $0.2$ for each pair of nodes~\citep{erdos1959random}. For each graph, we consider ferromagnetic models (i.e., with all interactions being ferromagnetic) and spin glasses (i.e., in which ferromagnetic and antiferromagnetic edges occur). For ferromagnetic models we uniformly sample the couplings $\Jij$ from the range $(0,1)$, for spin glasses we uniformly sample them from the range $(-1,1)$. Also, we assume that there exist fields $\theta_i$, that we uniformly sample from the range $(-\frac{1}{8},\frac{1}{8})$. For each type of graphical model (i.e., with a specific graph structure and either purely ferromagnetic or mixed -- i.e., both ferromagnetic and antiferromagnetic -- couplings), we create $200$ diffe\-rent instances; that is, altogether we create $4 \times (200 + 200)=1600$ different graphical models. For each individual model we increase the inverse temperature $\beta$ from $0$ to $2$ (in steps of $0.1$), and for each value of $\beta$ we use our algorithm \texttt{BETHE-MIN} (Algorithm~\ref{algo:projected_quasi_Newton}, Appendix~\ref{sec:appendix_C}) with $100$ random initializations of the parameter vector $\qvec^{(0)}$ to minimize the Bethe free energy and estimate the marginals and partition function associated to the specific model and current value of $\beta$. We compare the obtained estimates to the exact marginals and partition function -- computed by the junction tree algorithm~\citep{lauritzen1988junctiontree} -- and evaluate the errors with respect to the error measures defined in Sec.~\ref{subsec:Bethe_min_algo}. We average these errors over the $100$ runs of \texttt{BETHE-MIN} and afterwards over the $200$ different models corresponding to a specific graph type and potential configuration (with $\beta$ still fixed to some value between $0$ and $2$). The errors together with standard deviations are plotted in Fig.~\ref{fig:phase_transitions_attractive} (ferromagnetic models) and Fig.~\ref{fig:phase_transitions_general} (spin glasses) against the inverse temperature. These figures also show the estimated values of $\beta$ at which $\FB$ becomes non-convex -- in green, Theorem~\ref{thm:polynomial_characterization_of_inf_Ri} -- and develops multiple minima -- in blue,~\citet{mooij2007sufficient}). These values were first estimated for each model invividually and then averaged over the $200$ different models (together with standard deviations). Furthermore, Fig.~\ref{fig:percentage_convexity_vs_uniqueFP} shows the percentages of models, for which $\FB$ is predicted to be convex or to have a unique minimum at a specific value of $\beta$. It illustrates the 'gap' between convexity and uniqueness of a minimum (with respect to $\beta$), which becomes smaller with a higher connectivity in the graph. Fig.~\ref{fig:iterations_of_BetheMin} shows the number of required iterations until $\texttt{BETHE-MIN}$ converges to a stationary point. The iteration number increases as the inverse temperature increases, which is most likely due to a changed stage of the Bethe free energy (in particular, multiple minima) and the numerical difficulties that arise for regimes of high inverse temperature; however, this appears to be less distinct in spin glasses (right-hand side) than in ferromagnetic models (left-hand side). The general convergence rate of \texttt{BETHE-MIN} is higher than $99\%$ (with respect to all $1600$ sampled models) which is an astonishingly convincing result. \\

Generally, we observe in Fig.~\ref{fig:phase_transitions_attractive} and Fig.~\ref{fig:phase_transitions_general} that the quality of the Bethe approximation degrades (and later stabilizes at a high level), if $\beta$ passes a certain criti\-cal threshold (which is -- due to a steeper increase of the errors -- more distinct for ferromagnetic models, but also observable for spin glasses). We define this phenomenon to be a phase transition in the model, i.e., the point at which the Bethe approximation begins to lose its reliability. Following this interpretation, different error measures are associated to different phase transitions. Ideally, we would like to predict the critical $\beta$ at which a phase transition occurs (though some errors might still be tolerable after a phase transition has occurred). The main focus of our analysis is to show how the predicted Bethe stage changes can be used to estimate a phase transition and thus to quantify a reliabilty threshold for the Bethe approximation. \\

We first discuss our results on ferromagnetic models in more detail. As can be seen in Fig.~\ref{fig:phase_transitions_attractive}, phase transitions not only differ between the considered error measures but also depend on the graph structure. If the connectivity among nodes is high (such as in the complete graph), phase transitions already occur for smaller values of $\beta$; if the graphs are sparse (such as the grid graphs), they occur somewhat later and the subsequent degradation of the Bethe approximation accuracy happens more slowly. Also, the phase transition with respect to the partition function happens earlier, than the phase transition with respect to the marginal errors (with the singleton marginal phase transition happening as latest). \\

If we use the Bethe stage change temperature that is predicted by Theorem~\ref{thm:polynomial_characterization_of_inf_Ri} (i.e., from convex to non-convex, drawn by the green solid line) to estimate singleton marginal phase transitions, we observe that these seem to be slightly underestimate. Whether the reason for this is the gap between the predicted and the 'true' convexity of $\FB$, or that convexity is less relevant for explaining this error, is not clear. In any case, the Bethe approximation is accurate for the predicted $\beta$. The Bethe stage change temperature that is predicted by~\citet{mooij2007sufficient} (i.e., from a unique minimum to multiple minima, drawn by the solid blue line) overestimates the sigleton marginal phase transition. In particular, it predicts some $\beta$ at which the singleton marginal error is already considerable; i.e., the Bethe approximation is not any longer reliable at this point. The actual phase transition lies bet\-ween the values predicted by these two results. For the pairwise marginals, Theorem~\ref{thm:polynomial_characterization_of_inf_Ri} predicts the phase transition quite accurately (with a slight overestimation on the complete graph, where the associated error already increases very early).~\citet{mooij2007sufficient} again overestimate the value of the phase transition and predict a value of $\beta$ at which the associated error is already significant. For the partition function, the situation is roughly similar as for the pairwise marginals; generally, however, the absolute error on the partition function seems always to remain in an acceptable regime for ferromagnetic models. \\

Second, we discuss our results on spin glasses (Fig.~\ref{fig:phase_transitions_general}). The general development of all three types of errors behaves differently compared to ferromagnetic models. In particular, the increase of errors at a phase transition happens less abrupt and subsequently lasts over a longer period. Also, the errors on the marginals keep at a lower level as in the ferromagnetic case (however, the error on the partition function grows even stronger, the higher $\beta$ increases). As for ferromagnetic models, singleton marginal phase transitions occur later than those with respect to the pairwise marginals and partition function. For the singleton marginals, Theorem~\ref{thm:polynomial_characterization_of_inf_Ri} clearly underestimates phase transitions, while~\citet{mooij2007sufficient} provide accurate estimates (with a slight over-/ or underestimation depending on the graph type). The pairwise marginal phase transitions are quite accurately predicted by Theorem~\ref{thm:polynomial_characterization_of_inf_Ri}, while the value predicted~\citet{mooij2007sufficient} tend to an overestimation. For the partition function, the phase transition lies somewhere between the thresholds that are predicted by the two results. \\

Our experiments show that both convexity and uniqueness of a Bethe minimum (resp., the results to predict these stages) are useful concepts to infer about the reliability of the Bethe approximation. We summarize our experimental observations in the following statements:
\begin{itemize}
 \item The Bethe stage change from convexity to non-convexity (as predicted by Theorem~\ref{thm:polynomial_characterization_of_inf_Ri}) accurately predicts a phase transition with respect to the pairwise marginals and partition function in ferromagnetic models, and with respect to the pairwise marginals in spin glasses.
 \item The Bethe stage change from a unique minimum to multiple minima (as predicted by~\citet{mooij2007sufficient}) accurately predicts a phase transition with respect to the singleton marginals in and spin glasses.
 \item A phase transition with respect to the partition function in spin glasses lies between the two Bethe stage changes described above.
\end{itemize}

\section{CONCLUSION} \label{sec:conclusion}
In this work, we have considered the Bethe approximation in the context of probabilistic models on finite graphs.
We have refined the concept of a convex Bethe free energy by studying its behavior on a specific submanifold of its domain, the Bethe box.
In an extent theoretical part, we have derived two sufficient conditions for convexity of the Bethe free energy that are easi\-ly applicable in practice.
In the experimental part, we have specified different stages of the Bethe free energy (one of them its convexity) and analyzed its reliability with respect to its stage.
We have demonstrated with help of our theoretical results that convexity is a strong concept for verifying the reliability of the Bethe approximation in important problems of probabilistic inference.
We have further proposed an effective algorithm for minimizing the Bethe free energy that uses the information of its Hessian.
To the best of our knowledge, the analyzes and experiments in this work are novel and fundamentally improve the understanding and accessibility of the Bethe approximation.

\appendix

\section{Proofs of Sec.~\ref{subsec:Bethe_Diagonal_Dominance}} \label{sec:appendix_A}

{\bf Proof} [\normalfont Proof of Lemma~\ref{lemma:split_infima} ] {\ \\}
Without loss of generality, we may assume that we have only one pairwise term $\rijp(\qi,\qj)$ in the definition of $\Ri(\qvec)$ (i.e., one single ferromagnetic edge that is incident to node $i$). The statement involving multiple edges follows by induction over the number of edges; furthermore, Lemma~\ref{lemma:equivalence_rijp_rijm} shows \footnote{Note that Lemma~\ref{lemma:equivalence_rijp_rijm} follows independently from Lemma~\ref{lemma:split_infima}, although it appears chronologically later in this work. Thus we can utilize its statements at this point.} that $\rijm(\qi,\qj)$ (corresponding to an antiferromagnetic edge) is identical to $\rijp(\qi,\qj)$ (only with shifted symmetries.) Hence, we must prove that
\begin{align} \label{eq:WLOG_Ri}
\inf\limits_{(\qi,\qj) \in (0,1) \times (0,1)} \Big[ \ri(\qi) + \rijp(\qi,\qj) \Big] = \inf\limits_{\qi \in (0,1)} \Big[ \ri(\qi) + \Big(\inf\limits_{\qj \in (0,1)} \rijp(\qi,\qj) \Big) \Big].
\end{align}
To prove this, we show that the left-hand side in~\eqref{eq:WLOG_Ri} is as most as large as the right-hand side, and vice versa. \\
First, let $\hat{\qi}$ and $\hat{\qj}$ be fixed values in $(0,1)$. Then, by definition of the infimum,
\begin{align*}
 \inf\limits_{\qj \in (0,1)}  \Big[ \ri(\hat{\qi}) + \rijp(\hat{\qi},\qj) \Big] \leq \ri(\hat{\qi}) + \rijp(\hat{\qi},\hat{\qj})
\end{align*}
or, equivalently,
\begin{align*}
 \inf\limits_{\qj \in (0,1)}  \Big[ \ri(\hat{\qi}) + \rijp(\hat{\qi},\qj) \Big] - \ri(\hat{\qi}) \leq \rijp(\hat{\qi},\hat{\qj}).
\end{align*}
Since this holds for all $\hat{\qj} \in (0,1)$, we can also take the infimum with respect to $\qj$ of the expression on the right-hand side and obtain
\begin{align*}
 \inf\limits_{\qj \in (0,1)}  \Big[ \ri(\hat{\qi}) + \rijp(\hat{\qi},\qj) \Big] - \ri(\hat{\qi}) \leq  \inf\limits_{\qj \in (0,1)}   \rijp(\hat{\qi},\qj).
\end{align*}
Now add $\ri(\hat{\qi})$ again and take the infimum with respect to $\qi$ on both sides of the inequality, which results in
\begin{align*}
 \inf\limits_{(\qi,\qj) \in (0,1) \times (0,1)}  \Big[ \ri(\qi) + \rijp(\qi,\qj) \Big] \leq  \inf\limits_{\qi \in (0,1)} \Big[ \ri(\qi) + \Big(\inf\limits_{\qj \in (0,1)} \rijp(\qi,\qj) \Big) \Big]
\end{align*}
and proves the first of the two inequalities. \\
To prove the reverse inequality, let once again $\hat{\qi}$ and $\hat{\qj}$ be fixed in $(0,1)$ and observe that
\begin{align*}
 \ri(\hat{\qi}) + \rijp(\hat{\qi},\hat{\qj}) \geq \ri(\hat{\qi}) + \inf\limits_{\qj \in (0,1)} \rijp(\hat{\qi},\qj).
\end{align*}
This is valid for all $\hat{\qi} \in (0,1)$ and we can take the infimum with respect to $\qi$ of both sides of this inequality:
\begin{align*}
 \inf\limits_{\qi \in (0,1)} \Big[ \ri(\qi) + \rijp(\qi,\hat{\qj}) \Big] \geq \inf\limits_{\qi \in (0,1)} \Big[ \ri(\qi) + \inf\limits_{\qj \in (0,1)} \rijp(\qi,\qj) \Big]
\end{align*}
Finally, we recall that this inequality holds for all $\hat{\qj} \in (0,1)$ and we can once again take the infimum with respect to $\qj$ of the expression on the left-hand side, resulting in the reverse inequality
\begin{align*}
 \inf\limits_{(\qi,\qj) \in (0,1) \times (0,1)} \Big[ \ri(\qi) + \rijp(\qi,\qj) \Big] \geq \inf\limits_{\qi \in (0,1)} \Big[ \ri(\qi) + \inf\limits_{\qj \in (0,1)} \rijp(\qi,\qj) \Big].
\end{align*} \hfill\BlackBox
\\

{\bf Proof} [\normalfont Proof of Lemma~\ref{lemma:equivalence_rijp_rijm} ] {\ \\}
For this proof we require some additional notation: Recall from~\eqref{eq:Ri_rijp} and~\eqref{eq:Ri_rijm} that $\rijp(\qi,\qj)$ and $\rijm(\qi,\qj)$ implicitly depend on $\xijopt(\qi,\qj)$ -- defined in~\eqref{eq:xi_optimal} -- and $\Tij(\qi,\qj)$ -- defined in~\eqref{eq:Tij_optimal} -- which are functions of $\qi$ and $\qj$ (with $\Tij(\qi,\qj)$ itself depending on $\xijopt(\qi,\qj)$). Here we refine the definition of these two sub-functions according to the type of edge they represent: If $\xijopt(\qi,\qj)$ and thus $\Tij(\qi,\qj)$ are associated to a ferromagnetic edge (i.e., implicitly depend on $\Jij > 0$), we denote them by $\xijp(\qi,\qj)$ and $\Tijp(\qi,\qj)$; if they are associated to an antiferromagnetic edge (i.e., implicitly depend on $-\Jij < 0$), we denote them by $\xijm(\qi,\qj)$ and $\Tijm(\qi,\qj)$. In terms of this refined notation, we can also adapt the definition of $\rijp(q_i,q_j)$ and $\rijm(q_i,q_j)$ according to
\begin{align}
\rijp(\qi,\qj) & = \frac{\qj (1-\qj) - \xijp(\qi,\qj) + \qi \qj}{\Tijp(\qi,\qj)} \quad \text{and} \label{eq:defi_rijp} \\
\rijm(\qi,\qj) & = \frac{\qj (1-\qj) - \qi \qj +  \xijm(\qi,\qj)}{\Tijm(\qi,\qj)} \label{eq:defi_rijm}.
\end{align}
To prove~\eqref{eq:equivalence_rijp_rijm}, we first show the following symmetry properties:
\begin{align}
\begin{split}
 \xijp(\qi,\qj) & =  \xijp(\qj,\qi) \\ \xijm(\qi,\qj) & =  \xijm(\qj,\qi) \label{eq:self_symmetry_xij}
\end{split} \\
\begin{split}
 \xijm(\qi,\qj) - \qi \qj & =  -\xijp(\qi,1-\qj) + \qi (1-\qj) \\ & = -\xijp(1-\qi,\qj) + (1-\qi) \qj \label{eq:symmetry_xij}
\end{split} \\
\begin{split}
 \Tijm(\qi,\qj) & = \Tijp(1-\qj,\qi) \\ & = \Tijp(\qj,1-\qi) \\ & = \Tijp(\qi,1-\qj) \\ & = \Tijp(1-\qi,\qj) \label{eq:symmetry_Tij}
 \end{split}
\end{align}
The two equalities in~\eqref{eq:self_symmetry_xij} are clear from definition~\eqref{eq:xi_optimal}, where we have commutativity with respect to $\qi$ and $\qj$.  We thus begin with the first equality in~\eqref{eq:symmetry_xij}. Let us define $\aijp \coloneqq e^{4 \beta \Jij} - 1$ and $\aijm \coloneqq e^{4 \beta (-\Jij)} - 1$. Then, using~\eqref{eq:xi_optimal} for an antiferromagnetic edge, we have
\begin{align} \label{eq:xijm_expanded}
 \xijm(\qi,\qj) =  \frac{1}{2\aijm} \Big( 1+\aijm(\qi+\qj) - \sqrt{\big(1+\aijm(\qi+\qj)\big)^2 - 4 \aijm(1+\aijm) q_i q_j  } \, \, \Big).
\end{align}
If we express $\aijm$ in terms of $\aijp$, we observe that
\begin{align} \label{eq:xijm_term1}
 \aijm= \frac{\aijp}{-e^{4 \beta \Jij}} = \frac{-\aijp}{\aijp+1}, \quad \text{which implies}
 \quad \frac{1}{2\aijm} = \frac{-(\aijp+1)}{2\aijp},
 \end{align}
 \begin{align} \label{eq:xijm_term2}
 1+\aijm(\qi+\qj) = 1 - \frac{\aijp}{\aijp+1}(\qi+\qj), \quad \text{and}
\end{align}
\begin{align} \label{eq:xijm_term3}
\begin{split}
 \hspace{-0.4cm} & \, \sqrt{\big(1+\aijm(\qi+\qj)\big)^2 - 4 \aijm(1+\aijm) q_i q_j  } \\ 
 \hspace{-0.4cm} = & \, \sqrt{\big(1 - \frac{\aijp}{\aijp+1}(\qi+\qj)\big)^2 + 4 \frac{\aijp}{\aijp+1} \big( 1 - \frac{\aijp}{\aijp+1} \big) \qi \qj} \\
 \hspace{-0.4cm} = & \, \sqrt{1 - 2\frac{\aijp}{\aijp +1}(\qi+\qj) + \frac{(\aijp)^2}{(\aijp+1)^2}(\qi+\qj)^2 + 4\frac{\aijp}{(\aijp+1)^2} \qi\qj} \\
 \hspace{-0.4cm} = & \, \frac{1}{\aijp+1} \sqrt{(\aijp+1)^2 - 2\aijp(\aijp+1)(\qi+\qj) + (\aijp)^2(\qi+\qj)^2+4\aijp\qi\qj} \\
 \hspace{-0.4cm}= & \, \frac{1}{\aijp+1} \sqrt{\big(1 + \aijp (\qi + (1-\qj)\big)^2- 4 \aijp (1+\aijp)\qi(1-\qj)}.
 \end{split}
\end{align}
Now substitute~\eqref{eq:xijm_term1} -~\eqref{eq:xijm_term3} into~\eqref{eq:xijm_expanded} and subtract $\qi\qj$:
\begin{align*}
& \, \xijm(\qi,\qj) - \qi \qj \\ 
= & \, \frac{-1}{2\aijp} \Big( (\aijp+1) - \aijp(\qi+\qj)       \\ 
& - \sqrt{\big(1 + \aijp (\qi + (1-\qj)\big)^2- 4 \aijp (1+\aijp)\qi(1-\qj)} \, \Big) - \qi \qj \\
= & \, \frac{-1}{2\aijp} \Big(1 + \aijp \big((\qi + (1 - \qj)\big) \\ 
& - \sqrt{\big(1 + \aijp (\qi + (1-\qj)\big)^2- 4 \aijp (1+\aijp)\qi(1-\qj)} \, \Big) + \qi -  \qi \qj \\
= & \, - \xijp(\qi,1-\qj) + \qi (1-\qj),
\end{align*}
where we have inserted $\xijp$ according to its definition~\eqref{eq:xi_optimal} for a ferromagnetic edge. For the second equality in~\eqref{eq:symmetry_xij}, we can perform the same computation as above but with reversed role of $\qi$ and $\qj$. Next, we use the symmetry properties of $\xijp$ and $\xijm$ to prove~\eqref{eq:symmetry_Tij}, where we only show the first equality (the others follow analogously): 
\begin{align*}
 \Tijm(\qi,\qj) \stackrel{\eqref{eq:Tij_optimal}}{=} & \, \qi \qj (1-\qi) (1-\qj) - (\xijm(\qi,\qj) - \qi \qj)^2 \\
 \stackrel{\eqref{eq:symmetry_xij}}{=} & \, \qi \qj (1-\qi) (1-\qj) - (- \xijp(\qi,1-\qj) + \qi (1-\qj))^2 \\
 \stackrel{\eqref{eq:self_symmetry_xij}}{=} & \, (1-\qj) \qi \qj (1-\qi) - (\xijp(1-\qj,\qi) - (1-\qj) \qi)^2 \\
 \stackrel{\eqref{eq:Tij_optimal}}{=} & \, \Tijp(1-\qj,\qi)
\end{align*}
Finally, we use the derived properties to prove~\eqref{eq:equivalence_rijp_rijm}:
\begin{align*}
 \rijm(\qi,\qj) \stackrel{\eqref{eq:defi_rijm}}{=} & \, \frac{\qj (1-\qj) - \qi \qj +  \xijm(\qi,\qj)}{\Tijm(\qi,\qj)} \\
 \stackrel{\eqref{eq:symmetry_xij}}{=} & \, \frac{\qj (1-\qj) + \qi (1-\qj) -  \xijp(\qi,1-\qj)}{\Tijm(\qi,\qj)} \\
 \stackrel{\eqref{eq:symmetry_Tij}}{=} & \, \frac{\qj (1-\qj) -  \xijp(\qi,1-\qj) + \qi (1-\qj)}{\Tijp(\qi,1-\qj)} \\
 \stackrel{\eqref{eq:defi_rijp}}{=} & \, \rijp(\qi,1-\qj)
\end{align*}
An analogous computation shows the second equality in~\eqref{eq:equivalence_rijp_rijm}. \hfill\BlackBox
\\

{\bf Proof} [\normalfont Proof of Lemma~\ref{lemma:partial_derivatives_xij_Tij_rijp} ] {\ \\}
 The formulas~\eqref{eq:deri_1_xij_qi},~\eqref{eq:deri_1_xij_qj} for the partial derivatives of $\xijopt$ are derived in Lemma 13 in~\citep{weller2013bethebounds}. All other formulas~\eqref{eq:deri_1_Tij_qi} -~\eqref{eq:deri_2_Tij_qj2} follow from the elementary rules of differentiation, applied to the definitions~\eqref{eq:xi_optimal},~\eqref{eq:Tij_optimal},~\eqref{eq:Ri_rijp} of $\xijopt$, $\Tij$, $\rijp$. \hfill\BlackBox
\\

{\bf Proof} [\normalfont Proof of Theorem~\ref{thm:unique_stationary_rijp} ] {\ \\}
First, we multiply both sides in~\eqref{eq:stationary_equation_rijp} (using formula~\eqref{eq:deri_1_rijp_qj}) by $\Tij^2$ to remove the denominator. The resulting equation is
\begin{align}
 (1-2\qj-\frac{\partial \xijopt}{\partial q_{j}} +\qi)\Tij = (\qj(1-\qj)-\xijopt + \qi\qj)\frac{\partial \Tij}{\partial q_{j}}. \label{eq:stationary_equation_rijp_proof}
\end{align}
Let us consider both sides of this equation separately. We substitute the formulas for $\Tij$ from~\eqref{eq:Tij_optimal} and $\frac{\partial \xijopt}{\partial q_{j}}$ from~\eqref{eq:deri_1_xij_qj} into the left-hand side of~\eqref{eq:stationary_equation_rijp_proof}:
\begin{align*}
 & \big((1-2\qj-\frac{\partial \xijopt}{\partial q_{j}} +\qi \big) \Tij \\ 
 = \, & \big(1-2\qj-\frac{\aij (\qi - \xijopt) + \qi}{1 + \aij(\qi+\qj-2\xijopt)} +\qi\big) \big(\qi \qj (1-\qi) (1-\qj) - (\xijopt - \qi \qj)^2\big) \\
 = \, & \Big[\big(1+ \aij(\qi+\qj-2\xijopt)\big) (1-2\qj+\qi)  \big(\qi \qj (1-\qi) (1-\qj) - (\xijopt - \qi \qj)^2 \big) \\
 & - \big(\aij (\qi - \xijopt) + \qi\big) \big(\qi \qj (1-\qi) (1-\qj) - (\xijopt - \qi \qj)^2 \big) \Big] \\
 & \cdot \frac{1}{1+ \aij(\qi+\qj-2\xijopt)} \\
 = \, &  \big(2\qj-1+\aij(\xijopt-\qj-\qi^2+2\qj^2+\qi\qj+2\qi\xijopt-4\qj\xijopt) \big) \\ 
 & \cdot \big(\qi^2\qj+\qi\qj^2-2\qi\qj\xijopt-\qi\qj+(\xijopt)^2 \big) \, \frac{1}{1+ \aij(\qi+\qj-2\xijopt)}
 \end{align*}
We also substitute the analytical expression for $\frac{\partial \Tij}{\partial q_{j}}$ from~\eqref{eq:deri_1_Tij_qj} (with $\frac{\partial \xijopt}{\partial q_{j}}$ already substituted into $\frac{\partial \Tij}{\partial q_{j}}$) into the right-hand side of~\eqref{eq:stationary_equation_rijp_proof}:
 \begin{align*}
& (\qj(1-\qj)-\xijopt + \qi\qj)\frac{\partial \Tij}{\partial q_{j}} \\
= \, & (\qj(1-\qj)-\xijopt + \qi\qj) \\
& \cdot \big( (1-2\qj)\qi(1-\qi) - 2(\xijopt - \qi \qj) (\frac{\aij (\qi - \xijopt) + \qi}{1 + \aij(\qi+\qj-2\xijopt)} - \qi) \big) \\
= \, & \Big[\big(1+ \aij(\qi+\qj-2\xijopt)\big) \Big( \big(\qj(1-\qj)-\xijopt + \qi\qj \big) (1-2\qj)\qi(1-\qi) \\
&  +2\qi(\xijopt-\qi\qj) \big(\qj(1-\qj) - (\xijopt-\qi\qj)\big) \Big) \\
& - 2 (\xijopt - \qi\qj) \big(\aij (\qi - \xijopt) + \qi \big) \big(\qj (1-\qj) - \xijopt + \qi\qj \big) \Big]\\
& \cdot \frac{1}{1+ \aij(\qi+\qj-2\xijopt)}
\end{align*}
If we multiply the obtained expressions for both sides by $1+ \aij(\qi+\qj-2\xijopt)$ and simplify some terms, equation~\eqref{eq:stationary_equation_rijp_proof} becomes
\begin{align*}
 & \big(2\qj-1+\aij(\xijopt-\qj-\qi^2+2\qj^2+\qi\qj+2\qi\xijopt-4\qj\xijopt) \big) \\
 & \cdot \big(\qi^2\qj+\qi\qj^2-2\qi\qj\xijopt-\qi\qj+(\xijopt)^2 \big) \\
 & - \big(1+ \aij(\qi+\qj-2\xijopt)\big) \qi \big(\qj(1-\qj)-\xijopt+\qi\qj \big) \big(1-\qi+2(\xijopt-\qj) \big) \\
 & + 2 (\xijopt - \qi\qj) \big(\aij (\qi - \xijopt) + \qi \big) \big(\qj (1-\qj) - \xijopt + \qi\qj \big) = 0
\end{align*}
or, after expanding and grouping terms with the same power of $\xijopt$ together,
\begin{align*}
 & (\qi^2\qj^2 + 2\qi^2\qj^3-2\qi^3\qj^2 -\qi^2\qj+\qi^3\qj-\aij\qi^2\qj+\aij\qi^3\qj-\aij\qi^2\qj^2+4\aij\qi^2\qj^3) \\
 + & (\qi-\qi^2+\aij\qi^2-\aij\qi^3-4\qi\qj^2+2\qi^2\qj+4\aij\qi\qj-3\aij\qi\qj^2-2\aij\qi^2\qj \\
 & -4\aij\qi\qj^3-4\aij\qi^2\qj^2) \xijopt \\
 + & (2\qj-1-4\aij\qi-3\aij\qj+3\aij\qi^2+4\aij\qj^2+5\aij\qi\qj+4\aij\qi\qj^2) (\xijopt)^2 \\
 + & (3\aij-2\aij\qi-4\aij\qj) (\xijopt)^3 = 0.
\end{align*}
Now recall from~\eqref{eq:xi_optimal} that $\xijopt(\qi,\qj)$ is a function of $\qi$ and $\qj$, having the form $\xijopt(\qi,\qj)=\frac{1}{2\alpha_{ij}} \Big( \Qij - \sqrt{\Qij^2 - 4 \alpha_{ij}(1+\alpha_{ij}) q_i q_j  } \, \, \Big)$ with $\Qij = 1+\aij(\qi+\qj)$. For a shorter notation, let us denote the square root term in $\xijopt$ by $\sqrt{\dotsb}$. By inserting the analytical form of $\xijopt$ in the previous equation, we then get
\begin{align} \label{eq:algebraic_form_stationary_rijp}
 \tilde{A} + \tilde{B} \, (\Qij - \sqrt{\dotsb}) + \tilde{C} \, (\Qij - \sqrt{\dotsb})^2 + \tilde{D} \, (\Qij - \sqrt{\dotsb})^3 = 0
\end{align}
with the coefficients
\begin{align*}
& \tilde{A} \coloneqq (\qi^2\qj^2 + 2\qi^2\qj^3-2\qi^3\qj^2 -\qi^2\qj+\qi^3\qj-\aij\qi^2\qj+\aij\qi^3\qj-\aij\qi^2\qj^2+4\aij\qi^2\qj^3), \\
& \tilde{B} \coloneqq \frac{1}{2\alpha_{ij}} (\qi-\qi^2+\aij\qi^2-\aij\qi^3-4\qi\qj^2+2\qi^2\qj+4\aij\qi\qj-3\aij\qi\qj^2-2\aij\qi^2\qj \\
 & \quad \quad -4\aij\qi\qj^3-4\aij\qi^2\qj^2), \\
& \tilde{C} \coloneqq \frac{1}{4\alpha_{ij}^2}(2\qj-1-4\aij\qi-3\aij\qj+3\aij\qi^2+4\aij\qj^2+5\aij\qi\qj+4\aij\qi\qj^2), \quad \text{and} \\
& \tilde{D} \coloneqq \frac{1}{8\alpha_{ij}^3}(3\aij-2\aij\qi-4\aij\qj).
\end{align*}
Also, the power terms can be expanded according to
\begin{align}
 & (\Qij - \sqrt{\dotsb})^2 = 2\Qij^2 - 2\Qij\sqrt{\dotsb}-4\aij(1+\aij)\qi\qj \quad \text{and} \label{eq:power_term_squared} \\
 & (\Qij - \sqrt{\dotsb})^3 = 4\Big( \Qij^3 + \big(\aij(1+\aij)\qi\qj-\Qij^2 \big)\sqrt{\dotsb} - 3\Qij\aij(1+\aij)\qi\qj \Big). \label{eq:power_term_cubic}
\end{align}
Next, we substitute~\eqref{eq:power_term_squared} and~\eqref{eq:power_term_cubic} into~\eqref{eq:algebraic_form_stationary_rijp}, which we subsequently write as a radical equation:
\begin{align} \label{eq:radical_equation_stationary_rijp}
\begin{split}
 & \tilde{A}+ \Qij \tilde{B}+ 2 \big(\Qij^2-2\aij(1+\aij)\qi\qj\big)\tilde{C}+4\big(\Qij^3-3\Qij\aij(1+\aij)\qi\qj\big) \tilde{D} \\
= \, & \Big[\tilde{B} + 2\Qij\tilde{C}+ 4\big(\Qij^2-\aij(1+\aij)\qi\qj \big)\Big] \sqrt{\dotsb}
\end{split}
 \end{align}
To find all $\qj \in (0,1)$ that solve~\eqref{eq:radical_equation_stationary_rijp}, we distinguish between two cases regarding the coefficient $\tilde{B} + 2\Qij\tilde{C}+ 4\big(\Qij^2-\aij(1+\aij)\qi\qj \big)$  of the root term: \\

\underline{Case 1:} $\tilde{B} + 2\Qij\tilde{C}+ 4\big(\Qij^2-\aij(1+\aij)\qi\qj \big)  = 0$ \\

We treat this case by first computing the set of all $\qj \in (0,1)$ that satisfy this equation. For this purpose, we substitute the algebraic expressions for $\tilde{B}, \tilde{C}, \tilde{D}$ from above, and $\Qij=1+\aij(\qi+\qj)$ into this equation and, after a few simplifications, obtain
\begin{align*}
\big((2\aij\qi-\aij) \qj^2 +(2\aij\qi^2-4\aij\qi+\aij-1)\qj + \aij\qi(1-\qi)- \qi +1 \big) = 0.
\end{align*}
We recognize this as a quadratic equation in $\qj$, whose solutions are
\begin{align*}
 \qj^{(1)} = 1-\qi \quad \quad \text{and} \quad \quad  \qj^{(2)} = \frac{1+\aij\qi}{\aij(2\qi-1)}.
\end{align*}
While $\qj^{(1)} = 1-\qi$ is of course a feasible solution (i.e., lies in the interval $(0,1)$ whenever $\qj \in (0,1)$), we show that this is not the case for $\qj^{(2)}$: If $\qi < 0.5$, then $(2\qi-1) < 1$, and hence $\qj^{(2)} < 0$; if $\qi = 0.5$, then $\qj^{(2)}$ is not even well-defined; if $\qi > 0.5$, then $\qj^{(2)}$ is lower bounded by $\frac{1+\aij}{\aij}$, which is $>1$. In summary, $\qj^{(1)} = 1-\qi$ is the only candidate for being a feasible solution to equation~\eqref{eq:radical_equation_stationary_rijp}, under the assumption that we are in Case 1. It remains to check, if $\qj^{(1)}$ indeed solves equation~\eqref{eq:radical_equation_stationary_rijp} or if Case 1 does not admit a feasible solution of~\eqref{eq:radical_equation_stationary_rijp}. We thus insert $(1-\qi)$ for $\qj$ in all terms (including $\tilde{A},\tilde{B},\tilde{C},\tilde{D},\Qij$) of~\eqref{eq:radical_equation_stationary_rijp} and, considering that the right-hand side equals to zero, arrive at the equation
\begin{align*}
 & (\qi^2(1-\qi)^2 + 2\qi^2(1-\qi)^3-2\qi^3(1-\qi)^2 -\qi^2(1-\qi)+\qi^3(1-\qi) \\
 & -\aij\qi^2(1-\qi)+\aij\qi^3(1-\qi)-\aij\qi^2(1-\qi)^2+4\aij\qi^2(1-\qi)^3) \\
 & + (1+\aij) \frac{1}{2\alpha_{ij}} (\qi-\qi^2+\aij\qi^2-\aij\qi^3-4\qi(1-\qi)^2+2\qi^2(1-\qi) \\
 & +4\aij\qi(1-\qi)-3\aij\qi(1-\qi)^2-2\aij\qi^2(1-\qi) - 4\aij\qi(1-\qi)^3 \\
 & -4\aij\qi^2(1-\qi)^2) \\
 & + 2 \big((1+\aij)^2-2\aij(1+\aij)\qi(1-\qi)\big) \\
 & \cdot \frac{1}{4\alpha_{ij}^2}(2(1-\qi)-1-4\aij\qi-3\aij(1-\qi)+3\aij\qi^2+4\aij(1-\qi)^2 \\
 & +5\aij\qi(1-\qi)+4\aij\qi(1-\qi)^2)\\
 &+4\big((1+\aij)^3-3(1+\aij)\aij(1+\aij)\qi(1-\qi)\big) \\
 & \cdot \frac{1}{8\alpha_{ij}^3}(3\aij-2\aij\qi-4\aij(1-\qi)) = 0.
\end{align*}
After a series of elementary algebraic manipulations, all terms on the left-hand side of this equation cancel. This finally proves that $\qj=1-\qi$ is the unique solution of equation~\eqref{eq:radical_equation_stationary_rijp}, assuming that we are in Case 1. \\

 \underline{Case 2:} $\tilde{B} + 2\Qij\tilde{C}+ 4\big(\Qij^2-\aij(1+\aij)\qi\qj \big)  \neq 0$ \\
 
 Using the assumption of Case 2, we can isolate the radical in equation~\eqref{eq:radical_equation_stationary_rijp}, by dividing both sides by $\tilde{B} + 2\Qij\tilde{C}+ 4\big(\Qij^2-\aij(1+\aij)\qi\qj \big)$. If we then, to remove the root term, square both sides of the resulting equation and simplify the involved terms, we obtain the purely polynomial equation
 \begin{align*}
  \frac{1}{\aij^4} \Big[ & (\qi+\qj-1)^2(2\aij^2\qi^2\qj^2-2\aij^2\qi^2\qj+\aij^2\qi^2-2\aij^2\qi\qj^2 \\
  & + \aij^2\qj^2-4\aij\qi\qj+2\aij(\qi+\qj) + 1)^2 \Big] \\
  = \frac{1}{\aij^4} \Big[ & \Big( \big(1+\aij(\qi+\qj)\big)^2 - 4\aij(1+\aij)\qi\qj \Big) \\
  & \cdot(\qi+\qj-1)^2 \big(1+\aij(\qi+\qj-2\qi\qj \big)^2 \Big]
 \end{align*}
or, equivalently,
 \begin{align*}
 4 \aij^4 \qi^2 \qj^2 (1-\qi)^2 (1-\qj)^2 = 0.
 \end{align*}
 Since the solutions to this equation, $\qj^{(3)}=0$ and $\qj^{(4)}=1$, are boundary points of the open interval $(0,1)$ and thus not feasible, equation~\eqref{eq:radical_equation_stationary_rijp} does not admit any feasible solution in Case 2. \\
 
We conclude that $\qj^{\ast} \coloneqq \qj^{(1)} = 1-\qi$ is the unique solution to equation~\eqref{eq:radical_equation_stationary_rijp} and thus to equation $\frac{\partial \rijp}{\partial q_{j}} = 0$, which finally proves the Theorem. \hfill\BlackBox
\\

{\bf Proof} [\normalfont Proof of Lemma~\ref{lemma:rijp_critical_values} ] {\ \\}
\begin{enumerate}
 \item [(a)] We evaluate $\rijp$ in $(\qi,1-\qi)$ and get 
 \begin{align*}
  \rijp(\qi,1-\qi) & \stackrel{\eqref{eq:Ri_rijp}}{=} \frac{(1-\qi)\qi - \xijopt(\qi,1-\qi) + \qi(1-\qi)}{\Tij(\qi,1-\qi)} \\
  & \stackrel{\eqref{eq:Tij_optimal}}{=} \frac{2\qi(1-\qi) - \xijopt(\qi,1-\qi)}{\qi (1-\qi)(1-\qi)\qi - \big(\xijopt(\qi,1-\qi) - \qi(1-\qi)\big)^2} \\
  & \, = \frac{2\qi(1-\qi) - \xijopt(\qi,1-\qi)}{- \big(\xijopt(\qi,1-\qi)\big)^2  +2\qi(1-\qi)\xijopt(\qi,1-\qi)} \\
  & \, = \frac{1}{\xijopt(\qi,1-\qi)}.
 \end{align*}

 \item [(b)] To compute the limit values of $\rijp$, we require the limit values of $\xijopt$. They have been computed in~\citep{leisenberger2022fixing}, Lemma 2:
 \begin{align}
  & \lim\limits_{\qj \to 0} \xijopt(\qi,\qj) = 0 \label{eq:limit_values_xij_0} \\
  & \lim\limits_{\qj \to 1} \xijopt(\qi,\qj) = \qi \label{eq:limit_values_xij_1}
 \end{align}
With this we have
\begin{align*}
 \lim\limits_{\qj \to 0} \rijp(\qi,\qj) = \lim\limits_{\qj \to 0} \frac{\overbrace{\qj(1-\qj) - \xijopt(\qi,\qj) + \qi\qj}^{\to 0}}{\underbrace{\qi\qj(1-\qi)(1-\qj) -\big(\xijopt(\qi,\qj)-\qi\qj\big)^2}_{\to 0}},
\end{align*}
where both the numerator and the denominator tend to zero as $\qj$ approaches zero. We thus apply de L'H\^{o}spitals's rule and replace the above expression with the quotient of the derivatives:
\begin{align} \label{eq:limit_de_lHospital}
\lim\limits_{\qj \to 0} \rijp(\qi,\qj) = \lim\limits_{\qj \to 0} \frac{\frac{\partial}{\partial q_{j}}\big(\qj(1-\qj)- \xijopt(\qi,\qj) + \qi\qj\big)}{\frac{\partial}{\partial q_{j}} \Tij(\qi,\qj)}
\end{align}
Using the formulas from Lemma~\ref{lemma:partial_derivatives_xij_Tij_rijp}, (a) yields
\begin{align*}
 & \lim\limits_{\qj \to 0} \Big[\frac{\partial}{\partial q_{j}} \big(\qj(1-\qj)- \xijopt(\qi,\qj) + \qi\qj\big)\Big] \\
 = & \lim\limits_{\qj \to 0} \Big[1-2\qj- \frac{\partial \xijopt}{\partial q_{j}} + \qi\Big] \\
  \stackrel{\eqref{eq:deri_1_xij_qj}}{=} & \lim\limits_{\qj \to 0} \Big[1-2\qj-\frac{\aij (\qi - \xijopt) + \qi}{1 + \aij(\qi+\qj-2\xijopt)} +\qi \Big] \\
 \stackrel{\eqref{eq:limit_values_xij_0}}{=} & 1 - \frac{\qi(1+\aij)}{1+\aij\qi} + \qi \\
 = & \frac{1+\aij\qi^2}{1+\aij\qi}
\end{align*}
and
\begin{align*}
& \lim\limits_{\qj \to 0} \frac{\partial \Tij}{\partial q_{j}} \\
\stackrel{\eqref{eq:deri_1_Tij_qj}}{=} & \lim\limits_{\qj \to 0} \Big[(1-2\qj)\qi(1-\qi) - 2(\xijopt - \qi \qj) (\frac{\partial \xijopt}{\partial q_{j}} - \qi) \Big] \\
\stackrel{\eqref{eq:deri_1_xij_qj}}{=} & \lim\limits_{\qj \to 0} \Big[(1-2\qj)\qi(1-\qi) - 2(\xijopt - \qi \qj) \big(\frac{\aij (\qi - \xijopt) + \qi}{1 + \aij(\qi+\qj-2\xijopt)} - \qi\big) \Big] \\
\stackrel{\eqref{eq:limit_values_xij_0}}{=} & \qi(1-\qi) - 2\cdot 0 \cdot \big(\frac{\qi(1+\aij)}{1+\aij\qi} - \qi\big) \\
= & \qi(1-\qi).
\end{align*}
Then,~\eqref{eq:limit_de_lHospital} is further equal to
\begin{align*}
\frac{\frac{1+\aij\qi^2}{1+\aij\qi}}{\qi(1-\qi)} = \frac{1+\aij\qi^2}{(1+\aij\qi)\qi(1-\qi)},
\end{align*}
which proves~\eqref{eq:limit_values_rijp_zero}. An analogous calculation can be performed to compute the second limit value~\eqref{eq:limit_values_rijp_one}.
\end{enumerate} \hfill\BlackBox
\\

{\bf Proof} [\normalfont Proof of Theorem~\ref{thm:maximum_rijp} ] {\ \\}
First, we consider the relation between $r_{ij}^{(+,0)}(\qi)$ and $r_{ij}^{(+,1)}(\qi)$. We use the derived formulas from Lemma~\ref{lemma:rijp_critical_values}, (b) to compute their difference as
\begin{align*}
 & r_{ij}^{(+,1)}(\qi) - r_{ij}^{(+,0)}(\qi) \\
 = & \lim\limits_{\qj \to 1} \rijp(\qi,\qj) - \lim\limits_{\qj \to 0} \rijp(\qi,\qj) \\
 \stackrel{\eqref{eq:limit_values_rijp_zero},\eqref{eq:limit_values_rijp_one}}{=} & \frac{1+\aij (1-\qi)^2}{\big(1+\aij(1-\qi)\big)\qi(1-\qi)} - \frac{1+\aij \qi^2}{\big(1+\aij\qi\big)\qi(1-\qi)} \\
= & \frac{\overbrace{\aij^2(1-\qi)}^{>0}(1-2\qi)}{\underbrace{(1+\aij\qi)}_{>0}\underbrace{\big(1+\aij(1-\qi)\big)}_{>0}}.
\end{align*}
This expression is positive if $\qi \in (0,0.5)$, negative if $\qi \in (0.5,1)$, and zero if $\qi=0.5$. \\

Second, we consider the difference between $\rijp(\qi,1-\qi)$ and $r_{ij}^{(+,0)}(\qi)$:
\begin{align*}
 & \rijp(\qi,1-\qi) - r_{ij}^{(+,0)}(\qi) \\
= & \, \rijp(\qi,1-\qi) - \lim\limits_{\qj \to 0} \rijp(\qi,\qj) \\
\stackrel{\eqref{eq:rijp_value_in_stationary},\eqref{eq:limit_values_rijp_zero}}{=} & \frac{1}{\xijopt(\qi,1-\qi)} - \frac{1+\aij \qi^2}{(1+\aij\qi)\qi(1-\qi)} \\
\stackrel{\eqref{eq:xi_optimal}}{=} & \frac{2\aij}{1+\aij - \sqrt{(1+\aij)^2 - 4 \alpha_{ij}(1+\alpha_{ij}) q_i (1-\qi)  }} - \frac{1+\aij \qi^2}{(1+\aij\qi)\qi(1-\qi)}
\end{align*}
This expression is positive if and only if
\begin{align*}
 & 2\aij (1+\aij\qi) \qi(1-\qi) \\ > \, & (1+\aij\qi^2) \big(1+\aij-\sqrt{(1+\aij)^2 - 4 \alpha_{ij}(1+\alpha_{ij}) q_i (1-\qi)  }\big)
\end{align*}
or
\begin{align*}
 & (1+\aij \qi^2) \sqrt{(1+\aij)^2 - 4 \alpha_{ij}(1+\alpha_{ij}) q_i (1-\qi)  } \\ > \, & (1+\aij \qi^2)(1+\aij) - 2\aij (1+\aij\qi) \qi(1-\qi).
\end{align*}
If the expression on the right-hand side is negative, then the inequality is trivially satisfied (since all terms on the left-hand side are positive). If the expression on the right-hand side is nonnegative, we can square both sides of the inequality to remove the root term and obtain the equivalent statement
\begin{align*}
 & (1+\aij \qi^2)^2 \big((1+\aij)^2 - 4 \alpha_{ij}(1+\alpha_{ij}) q_i (1-\qi)  \big) \\ > \, & \big((1+\aij \qi^2)(1+\aij) - 2\aij (1+\aij\qi) \qi(1-\qi)\big)^2.
\end{align*}
After simplifying and rearranging these terms, we arrive at the inequality
\begin{align*}
 \aij \qi^6 -4\aij \qi^5 + 6\aij \qi^4 - 4\aij \qi^3 + \aij \qi^2 > 0
\end{align*}
which, in turn, is equivalent to
\begin{align*}
 & \aij \qi^2 (1-\qi)^4 > 0,
\end{align*}
a relation that is fulfilled unless $\qi \in \{0,1\}$.
In summary, we have shown that $\rijp(\qi,1-\qi)$ is greater than $r_{ij}^{(+,0)}(\qi)$, which proves another part of the Theorem. Moreover, the relation $\rijp(\qi,1-\qi) > r_{ij}^{(+,0)}(\qi)$ (for all $\qi \in (0,1)$) follows analogously to the previous calculation and will therefore be omitted. \\

Finally, we conclude that the unique stationary point $\qj^{\ast}=1-\qi$ is a maximum of $\rijp(\qi,\qj)$ with respect to $\qj \in (0,1)$, because its value in that point is greater than its limit values at the boundary points (otherwise $\rijp(\qi,\qj)$ would have another stationary point in that interval, which represents the line segment between $(\qi,0)$ and $(\qi,1)$ visualized in Figure~\ref{fig:sketch_rijp}). \hfill\BlackBox
\\

\section{Proofs of Sec.~\ref{subsec:Bethe_Hessian_Decomposition}} \label{sec:appendix_B}

{\bf Proof} [\normalfont Proof of Lemma~\ref{lemma:determinant_infinite_temperature} ] {\ \\}
 To compute the infinite temperature limit of $\HBijsubdet$, we require the limit values of $\xijopt$ and $\Tij$ as $\beta \to 0$. First, note that 
 \begin{align} \label{eq:aij_limit}
  \lim \limits_{\beta \to \,  0} \aij = \lim \limits_{\beta \to \,  0} \big[ e^{4 \beta \Jij} - 1 \big] = 0,
 \end{align}
where we have used the definition of $\aij$ from~\eqref{eq:xi_optimal}. Next, we use from~\citep{welling2001belief} that $\xijopt$ is the (positive) solution of the quadratic equation
\begin{align} \label{eq:quadratic_equation_xij}
  \big(1 + \aij(\qi+\qj) \big) \xijopt = \aij (\xijopt)^2 + (1+\aij) \qi\qj.
\end{align}
We take the limit on both sides of~\eqref{eq:quadratic_equation_xij} and obtain
\begin{align} \label{eq:quadratic_equation_xij_limit}
 \lim \limits_{\beta \to \,  0} \big[ \big(\underbrace{ 1 + \aij(\qi+\qj)}_{\to \,  1} \big) \xijopt \big] = \lim \limits_{\beta \to \,  0} \big[\underbrace{\aij(\xijopt)^2}_{\to 0}\big] + \lim \limits_{\beta \to \,  0} \big[\underbrace{(1+\aij) \qi\qj}_{\to \, \qi\qj}\big] 
\end{align}
which, after using~\eqref{eq:aij_limit}, becomes
\begin{align} \label{eq:xij_limit_beta}
 \lim \limits_{\beta \to \,  0} \xijopt = \qi \qi.
\end{align}
With this and the definition~\eqref{eq:Tij_optimal} of $\Tij$, we further obtain 
\begin{align} \label{eq:Tij_limit_beta}
 \lim \limits_{\beta \to \,  0} \Tij = \lim \limits_{\beta \to \,  0} \big[ \qi \qj (1-\qi)(1-\qj) - \underbrace{(\xijopt - \qi\qj)^2}_{\to \,  0} \big] = \qi \qj (1-\qi)(1-\qj).
\end{align}
Finally, we use~\eqref{eq:xij_limit_beta} and~\eqref{eq:Tij_limit_beta} to derive the desired result~\eqref{eq:determinant_infinite_temperature}:
\begin{align*}
 & \lim \limits_{\beta \to \,  0} \big[ (\HBijsubdet)(\qi,\qj) \big] \\
 \stackrel{\eqref{eq:HBijsub_determinant}}{=} \, &  \lim \limits_{\beta \to \,  0} \Big[ \frac{\degi-1}{\degi} \cdot \frac{\degj-1}{\degj} \cdot \frac{1}{\qi\qj(1-\qi)(1-\qj)} + \frac{1}{\Tij} \Big(1 - \frac{\degi-1}{\degi} - \frac{\degj-1}{\degj} \Big) \Big] \\
 \stackrel{\eqref{eq:Tij_limit_beta}}{=} \, & \frac{\degi-1}{\degi} \cdot \frac{\degj-1}{\degj} \cdot \frac{1}{\qi\qj(1-\qi)(1-\qj)} + \frac{\big(1 - \frac{\degi-1}{\degi} - \frac{\degj-1}{\degj}\big)}{\qi\qj(1-\qi)(1-\qj)} \\
 = \, \,  & \frac{1}{\degi \degj} \cdot \frac{1}{\qi \qj (1-\qi) (1-\qj)} > 0
\end{align*} \hfill\BlackBox
\\

{\bf Proof} [\normalfont Proof of Lemma~\ref{lemma:determinant_monotonically_decreasing} ] {\ \\}
 We use the formula for $\HBijsubdet$ from~\eqref{eq:HBijsub_determinant}, in which we see that the term
 \begin{align*}
  \frac{\degi-1}{\degi} \cdot \frac{\degj-1}{\degj} \cdot \frac{1}{\qi\qj(1-\qi)(1-\qj)}
 \end{align*}
is independent from $\beta$ and in the second term
 \begin{align}
  \frac{1}{\Tij} \Big(1 - \frac{\degi-1}{\degi} - \frac{\degj-1}{\degj} \Big)
 \end{align}
only $\Tij$ depends -- implicitly via $\xijopt$ -- on $\beta$. Let us therefore first compute the partial derivatives of $\xijopt$ and $\Tij$ (using the formulas~\eqref{eq:xi_optimal} and~\eqref{eq:Tij_optimal}) with respect to $\beta$, starting with $\xijopt$. As in the proof of Lemma~\ref{lemma:determinant_infinite_temperature}, we use from~\citep{welling2001belief} that $\xijopt$ is the (positive) solution of the quadratic equation
\begin{align} \label{eq:quadratic_equation_xij_twice}
  \big(1 + \aij(\qi+\qj) \big) \xijopt = \aij (\xijopt)^2 + (1+\aij) \qi\qj.
\end{align}
Note that also $\aij$ (from~\eqref{eq:xi_optimal}) depends on $\beta$. Its derivative is
\begin{align} \label{eq:derivative_aij_to_beta}
 \frac{\partial}{\partial \beta} \aij = \frac{\partial}{\partial \beta} \big(e^{4 \beta \Jij} - 1  \big) = 4\Jij e^{4 \beta \Jij},
\end{align}
which is positive for a ferromagnetic edge ($\Jij > 0$) and negative for an antiferromagnetic edge ($\Jij < 0$).
Now we compute the partial derivatives with respect to $\beta$ on both sides of~\eqref{eq:quadratic_equation_xij_twice} and obtain
\begin{align}
\begin{split}
 & \big( (\frac{\partial}{\partial \beta} \aij)(\qi+\qj) \big) \xijopt +  \big(1 + \aij(\qi+\qj) \big) \frac{\partial}{\partial \beta} \xijopt \\
 = \, & (\frac{\partial}{\partial \beta} \aij) (\xijopt)^2 + 2 \aij \xijopt \frac{\partial}{\partial \beta} \xijopt + (\frac{\partial}{\partial \beta} \aij) \qi \qj.
 \end{split}
\end{align}
Rearranging this equation and expressing $\frac{\partial}{\partial \beta} \xijopt$ through the other terms gives
\begin{align*} 
 \frac{\partial}{\partial \beta} \xijopt = \frac{(\frac{\partial}{\partial \beta} \aij) \big((\xijopt)^2 - (\qi+\qj) \xijopt + \qi\qj \big) }{\big(1+\aij(\qi+\qj) - 2 \aij \xijopt \big)}
\end{align*}
or, equivalently,
\begin{align} \label{eq:derivative_xij_to_beta}
 \frac{\partial}{\partial \beta} \xijopt = (\frac{\partial}{\partial \beta} \aij )\frac{(\qi-\xijopt)(\qj-\xijopt)}{1 + \aij(\qi+\qj-2\xijopt)}.
\end{align}
The terms $(\qi-\xijopt)$ and $(\qj-\xijopt)$ are elements from the joint probability table~\ref{tab:prob_table} and are therefore positive (and hence also the term $(\qi+\qj-2\xijopt)$). With~\eqref{eq:derivative_aij_to_beta}, it follows that also $\frac{\partial}{\partial \beta} \xijopt$ is positive if $\Jij > 0$, and negative if $\Jij < 0$. Next, we consider the partial derivative of $\Tij$ with respect to $\beta$:
\begin{align} \label{eq:derivative_Tij_to_beta}
\begin{split}
 \frac{\partial}{\partial \beta} \Tij = \frac{\partial}{\partial \beta} \big( \qi\qj(1-\qi)(1-\qj) - (\xijopt - \qi\qj)^2) = - 2(\xijopt - \qi\qj) (\frac{\partial}{\partial \beta} \xijopt)
 \end{split}
\end{align}
According to Lemma~\ref{lemma:Weller_bounds} from Sec.~\ref{subsec:Bethe_Diagonal_Dominance}, we know that $\xijopt > \qi \qj$ for $\Jij > 0$, and $\xijopt < \qi \qj$ for $\Jij < 0$. In each case (i.e., whether we consider a ferromagnetic or an antiferromagnetic edge) it follows together with~\eqref{eq:derivative_xij_to_beta} that $\frac{\partial}{\partial \beta} \Tij$ is negative (or, equivalently, that $\frac{\partial}{\partial \beta} \frac{1}{\Tij}$ is positive). Finally, note that
\begin{align*}
 \Big(1 - \frac{\degi-1}{\degi} - \frac{\degj-1}{\degj} \Big)
\end{align*}
is negative if we assume\footnote{As we will see later in this section, $\HBijsub$ is always positive definite if $\degi \leq 2$ or $\degj \leq 2$. Therefore, this is a meaningful assumption at this point.} that both $\degi, \degj$ are greater than $2$. Then the statement of Lemma~\ref{lemma:determinant_monotonically_decreasing} follows, because
\begin{align}
 \frac{\partial}{\partial \beta} \HBijsubdet \stackrel{\eqref{eq:HBijsub_determinant}}{=} \underbrace{\Big(1 - \frac{\degi-1}{\degi} - \frac{\degj-1}{\degj} \Big)}_{< 0} \underbrace{\frac{\partial}{\partial \beta} \frac{1}{\Tij}}_{> 0} < 0.
\end{align} \hfill\BlackBox
\\

{\bf Proof} [\normalfont Proof of Lemma~\ref{lemma:determinant_HBijsub_center} ] {\ \\}
We evaluate $\HBijsubdet$ in $(0.5,0.5)$, using the formula derived in~\eqref{eq:HBijsub_determinant}:
\begin{align} \label{eq:HBijsubdet_center}
 \HBijsubdet(0.5,0.5) = 16 \cdot \frac{\degi-1}{\degi} \cdot \frac{\degj-1}{\degj} + \frac{1}{\Tij} \Big(1 - \frac{\degi-1}{\degi} - \frac{\degj-1}{\degj} \Big)
\end{align}
Recall that $\Tij = \Tij(\qi,\qj)$ is a function of $\qi$ and $\qj$, which must be evaluated separately. For this we can use a result from~\citep{leisenberger2022fixing} (derived during the proof of Lemma 5), which says that
\begin{align} \label{eq:Tij_center}
 \Tij(0.5,0.5) = \frac{1}{16 \cosh^2(\beta \Jij)}.
 \end{align}
Substituting~\eqref{eq:Tij_center} in~\eqref{eq:HBijsubdet_center}, and performing a few manipulations yields
\begin{align} \label{eq:HBijsubdet_center_simplified}
\begin{split}
 \HBijsubdet(0.5,0.5) & = 16 \, \Big(\frac{\degi-1}{\degi} \cdot \frac{\degj-1}{\degj} - \frac{\degi \degj - \degi - \degj}{\degi \degj} \cosh^2(\beta \Jij) \Big) \\
 & = 16 \, \Big( \frac{\degi\degj-\degi-\degj+1 - \cosh^2(\beta \Jij) (\degi\degj - \degi - \degj)}{\degi \degj} \Big) \\
 & = 16 \, \Big( \frac{(\degi\degj-\degi-\degj) \big(1 - \cosh^2(\beta \Jij)\big) + 1 }{\degi \degj} \Big).
 \end{split}
\end{align}

With the trigonometric identity
\begin{align*} 
 1 - \cosh^2(\beta \Jij) = \frac{1}{2} \big(1 - \cosh(2 \beta \Jij \big)
\end{align*}
equation~\eqref{eq:HBijsubdet_center_simplified} is further equal to
\begin{flalign} \label{eq:HBijsubdet_center_further_simplified}
 & 16 \, \Big( \frac{\frac{1}{2} \cdot (\degi\degj-\degi-\degj) \big(1 - \cosh(2 \beta \Jij) \big) + 1 }{\degi \degj} \Big) \nonumber \\
 = \, & \, \frac{8}{\degi\degj} \cdot \Big((\degi\degj-\degi-\degj) \big(1 - \cosh(2 \beta \Jij) \big) + 2\Big).
\end{flalign}
So far, we have found a relatively simple expression for $\HBijsubdet(0.5,0.5)$. We now aim to determine its sign in dependence of $\beta$. To this end, let us make a couple of observations: First, note that the term $\frac{8}{\degi\degj}$ is only a positive scale factor which can be ignored. Second, we observe that the term
\begin{align} \label{eq:sign_HBijsubdet_center}
 (\degi\degj-\degi-\degj) \big(1 - \cosh(2 \beta \Jij) \big) + 2
\end{align}
is an even function of $\Jij$ (i.e., we may exchange $\Jij$ by $-\Jij$, without modifying the outcome), because the hyperbolic cosine is an even function of its argument. In other words, the sign of $\HBijsubdet(0.5,0.5)$ depends only on the absolute value of $\Jij$ but not on its sign -- we may thus replace $\Jij$ by $|\Jij|$ in~\eqref{eq:sign_HBijsubdet_center}. Third, the hyperbolic cosine is lower bounded by one (and equals this lower bound if and only if its argument is zero), which implies that
\begin{align*}
 1 - \cosh(2 \beta |\Jij|) < 0.
\end{align*}
Finally, the term $\degi\degj-\degi-\degj$ is always posititve under our assumption that both $\degi,\degj > 2$.
From these observations, we conclude that $\HBijsubdet(0.5,0.5)$ is positive if and only if
\begin{align*}
 (\degi\degj-\degi-\degj) \big(1 - \cosh(2 \beta \Jij) \big) >-2
\end{align*}
or, equivalently,
\begin{align*}
\beta <  \frac{1}{2 |\Jij|} \arccosh \big(1 + \frac{2}{\degi \degj - \degi - \degj} \big) \eqqcolon \betacrit.
\end{align*}
Exchanging the less-than sign by the equals sign or the greater-than sign proves the complete statement of the Lemma. \hfill\BlackBox
\\

{\bf Proof} [\normalfont Proof of Theorem~\ref{thm:HBijsubdet_at_betacrit} ] {\ \\}
 We already know from Lemma~\ref{lemma:determinant_HBijsub_center} that $(0.5,0.5)$ is a solution to the equation
 \begin{align} \label{eq:HBijsubdet_starting_equation}
  \HBijsubdet(\qi,\qj) = 0;
 \end{align}
now we must prove that it is the unique solution. Afterwards, to prove that $\HBijsubdet$ is positive for all $(\qi,\qj) \neq (0.5,0.5)$, we compute the Hessian matrix of $\HBijsubdet$ and show that it is positive definite. Since $\HBijsubdet$ is a continuous function of $(\qi,\qj)$, this implies the statement. \\

Concretely, the proof consists of three parts: In Part 1, we modify equation~\eqref{eq:HBijsubdet_starting_equation} to bring it into a simpler form. In Part 2, we show that $(0.5,0.5)$ is the unique solution to this modified equation, and, by construction, the unique solution to the original equation. In Part 3, we finally show that $\HBijsubdet$ positive for all other points in the domain. \\

\underline{Part 1:} In equation~\eqref{eq:HBijsubdet_starting_equation}, we substitute the expression
\begin{align} \label{eq:HBijsubdet_starting_equation_substituted}
 \frac{\degi-1}{\degi} \cdot \frac{\degj-1}{\degj} \cdot \frac{1}{\qi\qj(1-\qi)(1-\qj)} + \frac{1}{\Tij} \Big(1 - \frac{\degi-1}{\degi} - \frac{\degj-1}{\degj} \Big)
\end{align} 
for $\HBijsubdet$, which has been derived in~\eqref{eq:HBijsub_determinant}. Without loss of generality, we may assume that $\Jij > 0$ (i.e., we consider a ferromagnetic edge), because of the symmetry properties\footnote{More precisely, if we rotate $\HBijsubdet$ by 90 degrees in its domain around the center point, we obtain $\HBijsubdet$ which implicitly depends on a coupling with the same strength but an inverse sign. This transformation does not change the number of its roots, though. Hence, if $(0.5,0.5)$ is the unique root of $\HBijsubdet$ for $\Jij >0$, it is also for $\Jij < 0$.} of $\Tij$ derived in the proof of Lemma~\ref{lemma:equivalence_rijp_rijm} in Sec.~\ref{subsec:Bethe_Diagonal_Dominance}.
To abbreviate the notation, we define
\begin{align}
\begin{split} \label{eq:c1_c2_Q}
 & \ci \coloneqq \frac{\degi-1}{\degi} \cdot \frac{\degj-1}{\degj}, \\
 & \cii \coloneqq 1 - \frac{\degi-1}{\degi} - \frac{\degj-1}{\degj}, \quad \text{and} \\
 & \qprod \coloneqq \qi\qj(1-\qi)(1-\qj).
 \end{split}
\end{align}
Then, after multiplying by $\Tij$, equation~\eqref{eq:HBijsubdet_starting_equation_substituted} is equivalent to
\begin{align*}
 \ci \frac{\Tij}{\qprod} + \cii = 0
\end{align*}
or, using the definition~\eqref{eq:Tij_optimal} of $\Tij$,
\begin{align} \label{eq:HBijsubdet_starting_equation_substituted_new}
 \ci + \cii - \ci \frac{(\xijopt - \qi\qj)^2}{\qprod} = 0.
\end{align}
Since
\begin{align*} 
 \ci + \cii \stackrel{\eqref{eq:c1_c2_Q}}{=} \frac{(\degi-1)(\degj-1) + \degi \degj - (\degi - 1) \degj - \degi(\degj - 1)}{\degi \degj} = \frac{1}{\degi \degj},
\end{align*}
we can multiply equation~\eqref{eq:HBijsubdet_starting_equation_substituted_new} by $\frac{1}{\degi \degj \qprod}$ and obtain
\begin{align} \label{eq:HBijsubdet_starting_equation_simplified}
 \qprod - (\degi-1) (\degj-1) ((\xijopt)^2 - 2 \xijopt \qi \qj + \qi^2\qi^2) = 0.
\end{align}
Next, we consider $\xijopt$, using its definition~\eqref{eq:xi_optimal}. We have
\begin{align}
 \xijopt = \frac{1}{2 \aij} \Big(\Qij - \sqrt{\Qij^2 - 4 \alpha_{ij}(1+\alpha_{ij}) q_i q_j}\Big),
\end{align}
where $\Qij = 1 + \aij (\qi+\qj)$ with $\aij = e^{4 \betacrit \Jij} - 1$ depending on the critical $\betacrit$. For simplicity, let us denote the square root term by $\sqrt{\dotsb}$. Then
\begin{align*}
 (\xijopt)^2 = \frac{1}{4 \aij^2} \big(\Qij^2 - 2 \Qij \sqrt{\dotsb} + \Qij^2 - 4 \aij (\aij+1) \qi \qj \big)
\end{align*}
and, after inserting the equations for $\xijopt$ and $(\xijopt)^2$, equation~\eqref{eq:HBijsubdet_starting_equation_simplified} becomes
\begin{align*}
 \qprod - (\degi-1) (\degj-1) \Big[ & \frac{1}{4\aij^2} \big(\Qij^2 - 2 \Qij\sqrt{\dotsb} + \Qij^2 - 4\aij(\aij+1) \qi\qj\big) \\
 & - \frac{\qi\qi}{\aij}(\Qij-\sqrt{\dotsb})   + \qi^2\qj^2\Big] = 0.
\end{align*}
Further manipulations yield
\begin{align*}
 \qprod - (\degi-1) (\degj-1) \Big[(\Qij - \sqrt{\dotsb}) \big(\frac{\Qij}{2 \aij^2} - \frac{\qi\qj}{\aij} \big) - \frac{(\aij+1)}{\aij}\qi\qj + \qi^2\qj^2 \Big] = 0,
\end{align*}
where we can isolate the radical according to
\begin{align*}
 \frac{\qprod - (\degi-1)(\degj-1) \Big(\Qij \big(\frac{\Qij}{2 \aij^2} - \frac{\qi\qj}{\aij} \big) - \frac{(\aij+1)}{\aij}\qi\qj + \qi^2\qj^2\Big)}{\big(\frac{\Qij}{2 \aij^2} - \frac{\qi\qj}{\aij} \big) (\degi-1)(\degj-1)}   = (-1) \cdot \sqrt{\dotsb}.
\end{align*}
Now squaring both sides of this equation\footnote{Note that the squaring operation might induce additional solutions to the resulting equation, which are not solutions to the original equation. Each solution to the new equation, however, is also a solution to the original equation. In the later proof, we show that $(0.5,0.5)$ is the unique feasible solution to the new equation, and thus the only candidate for a solution to the original equation.} and manipulating the result gives
\begin{align*}
 & \qprod^2 - 2 \qprod (\degi-1)(\degj-1) \Big(\Qij \big(\frac{\Qij}{2 \aij^2} - \frac{\qi\qj}{\aij} \big) - \frac{(\aij+1)}{\aij}\qi\qj + \qi^2\qj^2 \Big) \\
 + \, & \frac{(\degi-1)^2(\degj-1)^2}{4\aij^4} \Big(\Qij^2 - 2\Qij \aij \qi\qj + 2\aij^2\qi^2\qj^2 - 2\aij^2\qi\qj - 2\aij\qi\qj \Big)^2 \\
 - \, & \frac{(\degi-1)^2(\degj-1)^2}{4\aij^4}\big(\Qij - 2\aij\qi\qj \big)^2 (\Qij^2 - 4\aij\qi\qj - 4\aij^2 \qi\qj) = 0,
\end{align*}
where we substitute $\Qij = 1 + \aij(\qi+\qj)$ and simplify the result:
\begin{align*}
\qprod^2 - \frac{2(\degi-1)(\degj-1) \qprod}{2 \aij^2} \big(\aij^2(2\qi^2\qj^2 - 2\qi^2\qj - 2\qi\qj^2 +\qi^2 + \qj^2) \\ 
+ \aij(2\qi+2\qj-4\qi\qj) + 1  \big) \\
+ \frac{(\degi-1)^2(\degj-1)^2}{4\aij^4} \big(\aij^2(2\qi^2\qj^2 - 2\qi^2\qj - 2\qi\qj^2 +\qi^2 + \qj^2) \\ 
+ \aij(2\qi+2\qj-4\qi\qj) + 1 \big)^2 \\
+ \frac{(\degi-1)^2(\degj-1)^2}{4\aij^4} \big(\aij^2(\qi^2 + \qj^2 - 2\qi\qj) + \aij(2 \qi + 2\qj - 4\qi\qj) +1 \big) \\
\cdot \big(1 + \aij(\qi+\qj) - 2\aij\qi\qj \big)^2 = 0
\end{align*}
This equation is the same as
\begin{align*}
 \qprod^2 - \frac{(\degi-1)(\degj-1) \qprod}{\aij^2} \big(\aij^2(2\qi^2\qj^2 - 2\qi^2\qj - 2\qi\qj^2 +\qi^2 + \qj^2) \\ 
+ \aij(2\qi+2\qj-4\qi\qj) + 1 \big) + (\degi-1)^2(\degj-1)^2 \qprod^2 = 0
\end{align*}
or
\begin{align*}
 \frac{\qprod}{\aij^2} \Big[\aij^2\qprod - (\degi-1)(\degj-1) \big(\aij^2(2\qi^2\qj^2 - 2\qi^2\qj - 2\qi\qj^2 +\qi^2 + \qj^2) \\ 
+ \aij(2\qi+2\qj-4\qi\qj) + 1 \big) + (\degi-1)^2(\degj-1)^2 \aij^2 \qprod \Big] = 0.
\end{align*}
Recall that we have defined $\qprod$ in~\eqref{eq:c1_c2_Q} as $\qi \qj (1-\qi)(1-\qj)$, which means that $\qi=0$, $\qi=1$, $\qj=0$, and $\qj=1$ are solutions to the above equation; however, they are not feasible (as the domain of $\HBijsubdet$ is the open set $(0,1) \times (0,1)$) and therefore not relevant for our further considerations. Yet, we can multiply the previous equation by $\frac{\aij^2}{\qprod}$, which leaves us with
\begin{align} \label{eq:HBijsubdet_equation_before_inserting_aij}
\begin{split}
 \aij^2\qprod - (\degi-1)(\degj-1) \big(\aij^2(2\qi^2\qj^2 - 2\qi^2\qj - 2\qi\qj^2 +\qi^2 + \qj^2) \\ 
+ 2\aij(\qi+\qj-2\qi\qj) + 1 \big) + (\degi-1)^2(\degj-1)^2 \aij^2 \qprod = 0.
\end{split}
\end{align}
Now we consider $\aij$, in whose definition $\aij = e^{4 \beta \Jij} - 1$ (from~\eqref{eq:xi_optimal}) we substitute the critical inverse temperature $\betacrit$ according to formula~\eqref{eq:critical_beta}:
\begin{align} \label{eq:aij_with_critical_beta}
 \aij = \exp \Big({\frac{2\Jij}{|\Jij|} \arccosh \big(1 + \frac{2}{\degi \degj - \degi - \degj} \big) } \Big) - 1
\end{align}
With the trigonometric identity\footnote{This trigonometric identity is valid for a positive argument of $\arccosh$, which is satisfied in our case as we assumed $\degi, \degj > 2$.}
\begin{align*}
\begin{split}
 & \arccosh \big(1 + \frac{2}{\degi \degj - \degi - \degj} \big) \\
 = & \log\Big(\sqrt{\big(1 + \frac{2}{\degi \degj - \degi - \degj}\big)^2 - 1 } + 1 + \frac{2}{\degi \degj - \degi - \degj} \Big)
 \end{split}
\end{align*}
it follows that~\eqref{eq:aij_with_critical_beta} is further equal to 
\begin{align*}
& \exp\Big[\log\Big(\sqrt{\big(1 + \frac{2}{\degi \degj - \degi - \degj}\big)^2 - 1 } + 1 + \frac{2}{\degi \degj - \degi - \degj} \Big)^2             \Big] -1 \nonumber \\
 = \, & \frac{(2 + 2 \sqrt{\degi \degj - \degi - \degj + 1} + \degi \degj - \degi - \degj)^2}{(\degi \degj - \degi - \degj)^2}  - 1 \\
 = \, & 4\, \Big( \frac{2(\degi-1)(\degj-1) + (2 + \degi\degj-\degi-\degj) \sqrt{(\degi-1)(\degj-1)}}{(\degi \degj - \degi - \degj)^2}\Big),
\end{align*}
where we have used our assumption $\Jij >0$ made the beginning of this proof (i.e., $\frac{\Jij}{|\Jij|} = 1$). If we define
\begin{align} \label{eq:definition_of_D}
D \coloneqq \degi \degj - \degi - \degj + 1 = (\degi - 1) (\degj - 1), 
\end{align}
we get
\begin{align} \label{eq:aij_with_critical_beta_simplified}
 \aij = 4 \, \Big( \frac{2D + (D+1)\sqrt{D}}{(D-1)^2} \Big).
\end{align}
We return to equation~\eqref{eq:HBijsubdet_equation_before_inserting_aij}, into which we insert expression~\eqref{eq:aij_with_critical_beta_simplified} for $\aij$:
\begin{align*}
 & \, \frac{16D}{(D-1)^4} \big(4D + 4(D+1)\sqrt{D} + (D+1)^2\big) \qprod \\
& \, - D \Big[\frac{16D}{(D-1)^4} \big(4D + 4(D+1)\sqrt{D} + (D+1)^2\big)(2\qi^2\qj^2 - 2\qi^2\qj - 2\qi\qj^2 +\qi^2 + \qj^2) \\
& \quad \quad \quad+ 8 \, \Big( \frac{2D + (D+1)\sqrt{D}}{(D-1)^2} \Big) (\qi+\qj-2\qi\qj) + 1 \Big] \\
& \, + \frac{16D^3}{(D-1)^4} \big(4D + 4(D+1)\sqrt{D} + (D+1)^2\big) \qprod = 0
\end{align*}
We multiply this equation by $\frac{(D-1)^4}{D}$, rearrange a few terms, and arrive at
\begin{align} \label{eq:HBijsubdet_final_equation}
 \begin{split}
 & 16 \big(4D + 4(D+1) \sqrt{D} + (D+1)^2 \big)(1+D^2) \qprod - (D-1)^4 \\
 & - 16 D \big(4D + 4(D+1) \sqrt{D} + (D+1)^2 \big) (2\qi^2\qj^2 - 2\qi^2\qj - 2\qi\qj^2 +\qi^2 + \qj^2) \\
 & - 8 (D-1)^2 \big(2D + (D+1) \sqrt{D} \big) (\qi + \qj - 2\qi\qj) = 0,
 \end{split}
\end{align}
which is the final form of our equation that we will analyze in the second part of this proof. \\

\underline{Part 2:}
In this part, we show that the center point $(0.5,0.5)$ is the unique solution to the modified equation~\eqref{eq:HBijsubdet_final_equation}. However,  instead of solving this equation directly, we apply an alternative strategy. Namely, we consider the left-hand side of~\eqref{eq:HBijsubdet_final_equation} as a funtion of $(\qi,\qj)$ and prove that it has a stationary point in $(0.5,0.5)$, in which it takes the value zero. Subsequently, we compute the limit values of this function at the boundary and show that they are all negative. We finally argue that this function is negative for all points $(\qi,\qj) \neq (0.5,0.5)$ in $(0,1) \times (0,1)$ and that $(0.5,0.5)$ must consequently be its unique global maximum (and thus the unique solution to equation~\eqref{eq:HBijsubdet_final_equation}). \\

Let us define the left-hand side of~\eqref{eq:HBijsubdet_final_equation} as
\begin{align} \label{eq:L_prime}
\begin{split}
 & \Lp(\qi,\qj) \coloneqq \\
 & 16 \big(4D + 4(D+1) \sqrt{D} + (D+1)^2 \big) (1+D^2) \qprod - (D-1)^4 \\
 & - 16 D \big(4D + 4(D+1) \sqrt{D} + (D+1)^2 \big) (2\qi^2\qj^2 - 2\qi^2\qj - 2\qi\qj^2 +\qi^2 + \qj^2) \\
 & - 8 (D-1)^2 \big(2D + (D+1) \sqrt{D} \big) (\qi + \qj - 2\qi\qj),
 \end{split}
\end{align}
considered as a function of $(\qi,\qj)$. To compute its stationary points (i.e., points where the gradient $\nabla \Lp(\qi,\qj)$ vanishes), we calculate its partial derivatives with respect to $\qi$ and $\qj$, and set them both equal to zero:
\begin{align} \label{eq:partial_derivative_Lp_qi}
\begin{split}
 & \frac{\partial}{\partial \qi} \Lp(\qi,\qj) \\
 = \, & 16 \big(4D + 4(D+1) \sqrt{D} + (D+1)^2\big) (1 + D^2) (1-2\qi) \qj (1- \qj) \\
 & - 32 D \big(4D + 4(D+1) \sqrt{D} + (D+1)^2 \big) (2\qi\qj^2 - 2\qi\qj - \qj^2 + \qi) \\
 & - 8 (D-1)^2 \big(2D + (D+1)\sqrt{D} \big) (1-2\qj) \stackrel{!}{=} 0
 \end{split}
\end{align}
\begin{align} \label{eq:partial_derivative_Lp_qj}
\begin{split}
 & \frac{\partial}{\partial \qj} \Lp(\qi,\qj) \\
 = \, & 16 \big(4D + 4(D+1) \sqrt{D} + (D+1)^2\big) (1 + D^2)  \qi (1- \qi) (1-2\qj) \\
 & - 32 D \big(4D + 4(D+1) \sqrt{D} + (D+1)^2 \big) (2\qi^2\qj - 2\qi\qj - \qi^2 + \qj) \\
 & - 8 (D-1)^2 \big(2D + (D+1)\sqrt{D} \big) (1-2\qi) \stackrel{!}{=} 0
 \end{split}
\end{align}
This equation system is hard to solve directly, and therefore we compute the difference between the two partial derivatives which we then set to zero:
\begin{align} \label{eq:difference_partial_derivatives_Lp}
\begin{split}
 & (\frac{\partial}{\partial \qi} \Lp - \frac{\partial}{\partial \qi} \Lp) (\qi,\qj) \\
 = \, & 16 \big(4D + 4(D+1) \sqrt{D} + (D+1)^2\big) (\qi-\qj)  \\
 & \cdot \Big[(1 + D^2) (-1 + \qi + \qj - 2\qi\qj) - 2 D (1 + \qi + \qj - 2\qi\qj) \Big] \\
 & - 16 (D-1)^2 \big(2D + (D+1)\sqrt{D} \big) (\qi-\qj) \stackrel{!}{=} 0
 \end{split}
\end{align}
Note that solutions to~\eqref{eq:difference_partial_derivatives_Lp} are the only potential solutions to the equation system~\eqref{eq:partial_derivative_Lp_qi} $+$~\eqref{eq:partial_derivative_Lp_qj}, and thus the only candidates for stationary points of $\Lp$ (while, on the other hand, there might be solutions to~\eqref{eq:difference_partial_derivatives_Lp} which are not stationary points of $\Lp$). We observe that in equation~\eqref{eq:difference_partial_derivatives_Lp}, the term $(\qi-\qj)$ occurs as a common factor in all terms which implies that all points $(\qi,\qj)$ with $\qi=\qj$ are solutions to~\eqref{eq:difference_partial_derivatives_Lp} (and thus candidates for stationary points of $\Lp$). We need to check if there are other solutions to~\eqref{eq:difference_partial_derivatives_Lp}, i.e., for which the remaining equation (without the factor $16 \cdot (\qi-\qj)$)
\begin{align} \label{eq:L_prime_prime_equation}
\begin{split}
 & \big(4D + 4(D+1) \sqrt{D} + (D+1)^2\big) \\
 & \cdot \Big[(1 + D^2) (-1 + \qi + \qj - 2\qi\qj) - 2 D (1 + \qi + \qj - 2\qi\qj) \Big] \\
 & - (D-1)^2 \big(2D + (D+1)\sqrt{D} \big) = 0
 \end{split}
\end{align}
 is satisfied. We will show that~\eqref{eq:L_prime_prime_equation} has no feasible solutions, by defining its left-hand side as another function
\begin{align} \label{eq:L_prime_prime}
\begin{split}
 & \Lpp(\qi,\qj) \coloneqq \\
  & \big(4D + 4(D+1) \sqrt{D} + (D+1)^2\big) \\
  & \cdot \Big[(1 + D^2) (-1 + \qi + \qj - 2\qi\qj) - 2 D (1 + \qi + \qj - 2\qi\qj) \Big] \\
 & - (D-1)^2 \big(2D + (D+1)\sqrt{D} \big),
 \end{split}
 \end{align}
and proving that $\Lpp$ is negative for all $(\qi,\qj)$ in $(0,1) \times (0,1)$. For that purpose, we first show that $(0.5,0.5)$ is the unique stationary point of $\Lpp$ which, however, is a saddle point (i.e., both infimum and supremum of $\Lpp$ are not taken by $\Lpp$ but only approached as limit values at the boundary); afterwards, we compute the limit values of $\Lpp$ at the boundary and show that they are all negative. From this, we conclude that $\Lpp$ must be negative on its entire domain. We therefore compute the partial derivatives of $\Lpp$ as
\begin{align} \label{eq:partial_derivative_Lpp_qi}
\begin{split}
 & \frac{\partial}{\partial \qi} \Lpp(\qi,\qj) \\
 = \, & \big(4D + 4(D+1) \sqrt{D} + (D+1)^2 \big) \cdot (1 - D)^2 \cdot (1 - 2\qj)
 \end{split}
\end{align}
and
\begin{align} \label{eq:partial_derivative_Lpp_qj}
\begin{split}
 & \frac{\partial}{\partial \qj} \Lpp(\qi,\qj) \\
 = \, & \big(4D + 4(D+1) \sqrt{D} + (D+1)^2 \big) \cdot (1 - D)^2 \cdot (1 - 2\qi).
 \end{split}
\end{align}
If we set both expressions to zero, we see that $(0.5,0.5)$ is the unique solution to the resulting equation system (since $\big(4D + 4(D+1) \sqrt{D} + (D+1)^2 \big) \cdot (1 - D)^2$ is positive, where $D$ has been defined in~\eqref{eq:definition_of_D}), and thus the only stationary point of $\Lpp$. The Hessian matrix of $\Lpp$ is
\begin{align*}
 \nabla^2 \Lpp(\qi,\qj)  = \underbrace{\big(4D + 4(D+1) \sqrt{D} + (D+1)^2 \big) \cdot (1 - D)^2}_{>0} \cdot 
 \begin{pmatrix}
  0 & -2\\
 -2  & 0
 \end{pmatrix},
\end{align*}
which is indefinite for all $(\qi,\qj)$ (because $\begin{pmatrix}  0 & -2 \\ -2  & 0 \end{pmatrix}$ has precisely one positive and one negative eigenvalue). In particular, $(0.5,0.5)$ (and any other point) is a saddle point of $\Lpp$. To characterize the behavior of $\Lpp$ as it approaches the boundary, we compute its corresponding limit values. Recall that $\Lpp$ is defined on $(0,1) \times (0,1)$ (which is an open subset of $\mathbb{R}^2$), and thus the boundary of its domain is precisely the union of the four line segments
 \begin{align} \label{eq:boundary_of_domain}
 \begin{split}
 & \{(q_i,q_j) \in \mathbb{R}^2 \, \vert \, q_i = 0, 0 \leq q_j \leq 1 \}  \, \, \cup  \{(q_i,q_j) \in \mathbb{R}^2 \, \vert \, q_i = 1, 0 \leq q_j \leq 1 \} \, \, \\  \cup \, & \{(q_i,q_j) \in \mathbb{R}^2 \, \vert \, 0 \leq q_i \leq 1, q_j = 0 \}  \, \, \cup \{(q_i,q_j) \in \mathbb{R}^2 \, \vert \, 0 \leq q_i \leq 1, q_j = 1 \},
 \end{split}
 \end{align}
i.e., the four edges (and corners) of the unit square. Hence, for $0 \leq k \leq 1$, the limit values of $\Lpp$ (defined in~\eqref{eq:L_prime_prime_equation}) can be computed as
\begin{align*}
 \lim_{\substack{q_i \to 0 \\ q_j \to k }} \Lpp(\qi,\qj)  = & \underbrace{\big(4D + 4(D+1) \sqrt{D} + (D+1)^2\big)}_{>0} \Big(\underbrace{(1+D^2)(-1+k)}_{\leq 0} \\
 & - \underbrace{2D(1+k))}_{\geq 0} \Big) - \underbrace{(D-1)^2 \big(2D +(D+1)\sqrt{D} \big)}_{>0} < 0,
\end{align*}
\begin{align*}
 \lim_{\substack{q_i \to 1 \\ q_j \to k }} \Lpp(\qi,\qj)  = & \underbrace{\big(4D + 4(D+1) \sqrt{D} + (D+1)^2\big)}_{>0} \Big(\underbrace{(1+D^2)(k-2k)}_{\leq 0} \\
 & - \underbrace{2D(2+k-2k))}_{\geq 0} \Big) - \underbrace{(D-1)^2 \big(2D +(D+1)\sqrt{D} \big)}_{>0} < 0,
\end{align*}
and analogously for $\lim_{\substack{q_i \to k \\ q_j \to 0 }} \Lpp(\qi,\qj)$ and $\lim_{\substack{q_i \to k \\ q_j \to 1 }} \Lpp(\qi,\qj)$. Hence, all limit values of $\Lpp$ at the boundary are negative. Since the only stationary point of $\Lpp$ is a saddle point, this implies that $\Lpp$ must be negative on its entire domain (otherwise, $\Lpp$ would attain a maximum somewhere). In particular, $\Lpp$ has no root in $(0,1) \times (0,1)$. With this, we have proven that equation~\eqref{eq:L_prime_prime_equation} has no feasible solution and that points $(\qi,\qj)$ with $\qi=\qj$ are consequently the only potential solutions to the equation system~\eqref{eq:partial_derivative_Lp_qi} $+$~\eqref{eq:partial_derivative_Lp_qj} (and thus the only candidates for stationary points of $\Lp$). With this knowledge, we return to equation~\eqref{eq:partial_derivative_Lp_qi}, into which\footnote{Note that we could also substitute $\qj=\qi$ into equation~\eqref{eq:partial_derivative_Lp_qj}, but due to symmetry this is equivalent.} we substitute $\qj=\qi$ to compute the actual stationary points of $\Lp$:
\begin{align} \label{eq:partial_derivative_Lp_qi_with_qj_inserted}
\begin{split}
 & \frac{\partial}{\partial \qi} \Lp(\qi,\qj = \qi) \\
 = \, & 16 \big(4D + 4(D+1) \sqrt{D} + (D+1)^2\big) \\
  & \cdot \Big[ (1 + D^2) \qi (1- \qi) (1-2\qi) - 2 D \qi (1- \qi) (1-2\qi) \Big] \\
 & - 8 (D-1)^2 \big(2D + (D+1)\sqrt{D} \big) (1-2\qi) \stackrel{!}{=} 0
 \end{split}
\end{align}
We observe that $(1-2\qi)$ is a common factor of all terms in~\eqref{eq:partial_derivative_Lp_qi_with_qj_inserted}. This means that $\qi=0.5$ is a solution to the above equation, that $(0.5,0.5)$ is a solution to the equation system~\eqref{eq:partial_derivative_Lp_qi} $+$~\eqref{eq:partial_derivative_Lp_qj}, and hence a stationary point of $\Lp$. To check whether equation~\eqref{eq:partial_derivative_Lp_qi_with_qj_inserted} has other solutions, we must compute the solutions to the remaining equation
\begin{align*} 
 \begin{split}
  & 2 \big(4D + 4(D+1) \sqrt{D} + (D+1)^2\big) (1-D)^2 \qi (1-\qi) \\
  & - (D-1^2) \big(2D + (D+1)\sqrt{D} \big) = 0,
 \end{split}
\end{align*}
which is equivalent to the quadratic equation
\begin{align} \label{eq:partial_derivative_Lp_quadratic equation}
 \qi^2 - \qi + \frac{\sqrt{D}}{2 (1+\sqrt{D})^2} = 0.
\end{align}
We solve equation~\eqref{eq:partial_derivative_Lp_quadratic equation} and obtain the solutions
\begin{align} \label{eq:solutions_quadratic equation}
 \qi^{(1)} = \frac{1}{2} - \sqrt{\frac{1}{4} - \frac{\sqrt{D}}{2(1+\sqrt{D})^2}} \quad \text{and} \quad \qi^{(2)} = \frac{1}{2} + \sqrt{\frac{1}{4} - \frac{\sqrt{D}}{2(1+\sqrt{D})^2}}.
\end{align}
Then the points $(\qi^{(1)},\qi^{(1)})$ and $(\qi^{(2)},\qi^{(2)})$ are also solutions to the equation system~\eqref{eq:partial_derivative_Lp_qi} $+$~\eqref{eq:partial_derivative_Lp_qj} and thus, in principle, stationary points of $\Lp$; yet, we have not shown if they are feasible at all (i.e., if $\qi^{(1)}$ and $\qi^{(2)}$ are in $(0,1)$). To address this issue, we observe that $\frac{\sqrt{D}}{2(1+\sqrt{D})^2}$ is monotonically decreasing as $D$ (defined in~\eqref{eq:definition_of_D}) increases; i.e., if $\degi$ or $\degj$ -- or both -- increases. As we have assumed $\degi > 2$ and $\degj>2$, it follows that $D\geq 4$ (if $\degi=\degj=3$), and hence $\frac{\sqrt{D}}{2(1+\sqrt{D})^2} \leq \frac{\sqrt{4}}{2(1 + \sqrt{4})^2}= \frac{1}{9}$.  Thus, $\qi^{(1)}$ and $\qi^{(2)}$ are bounded by
\begin{align} \label{eq:bound_one_qi}
\begin{split}
 & \qi^{(1)} \leq  \frac{1}{2} - \sqrt{\frac{1}{4} - \frac{1}{9}} = \frac{1}{2} - \frac{\sqrt{5}}{6} \quad  \text{and} \\
 & \qi^{(2)} \geq  \frac{1}{2} + \sqrt{\frac{1}{4} - \frac{1}{9}} = \frac{1}{2} + \frac{\sqrt{5}}{6}.
\end{split}
 \end{align}
On the other hand, if we consider the limit $D \to \infty$ (corresponding to infinitely many neighbors of node $i$ or node $j$), the term $\frac{\sqrt{D}}{2(1+\sqrt{D})^2}$ converges (from above) to zero. This induces a lower (resp. upper) bound on $\qi^{(1)}$ (resp. $\qi^{(2)}$) by observing that
\begin{align} \label{eq:bound_two_qi}
\begin{split}
 & \qi^{(1)} >  \frac{1}{2} - \sqrt{\frac{1}{4}} = \frac{1}{2} - \frac{1}{2} = 0 \quad  \text{and} \\
 & \qi^{(2)} <  \frac{1}{2} + \sqrt{\frac{1}{4}} = \frac{1}{2} + \frac{1}{2} = 1.
\end{split}
 \end{align}
The bounds~\eqref{eq:bound_one_qi} and~\eqref{eq:bound_two_qi} show that $\qi^{(1)}$ lies in the interval $(0,\frac{1}{2} - \frac{\sqrt{5}}{6})$ and $\qi^{(2)}$ lies in the interval $(\frac{1}{2} + \frac{\sqrt{5}}{6},1)$, which are both subsets of $(0,1)$. Hence, $(\qi^{(1)},\qj^{(1)}=\qi^{(1)})$ and $(\qi^{(2)},\qj^{(2)}=\qi^{(2)})$ are feasible stationary points of $\Lp$. In summary, $\Lp$ has precisely three stationary points in $(0,1) \times (0,1)$, which are the center point $(0.5,0.5)$ and the two symmetrical points $(\qi^{(1)},\qj^{(1)}=\qi^{(1)})$ and $(\qi^{(2)},\qj^{(2)}=\qi^{(2)})$ (with $\qi^{(1)}$ and $\qi^{(2)}$ defined in~\eqref{eq:solutions_quadratic equation}). \\

As the final step in this part, we need to determine which kind of stationary points they are, beginning with $(0.5,0.5)$. We therefore compute the Hessian matrix of $\Lp$ and evaluate it in $(0.5,0.5)$. We use the first-order derivatives of $\Lp$ derived in~\eqref{eq:partial_derivative_Lp_qi} and~\eqref{eq:partial_derivative_Lp_qj} to calculate its second-order partial derivatives with respect to $\qi$ and $\qj$:
\begin{align*}
  \begin{split}
   \frac{\partial^2 \Lp}{\partial \qi^2} = & -32 \big(4D + 4(D+1) \sqrt{D} + (D+1)^2\big) \Big[ (1+D^2) \qj (1-\qj) \\
   & + D (2\qj^2-2\qj+1) \Big] \\
   \frac{\partial^2 \Lp}{\partial \qi \partial \qj} = & \, \,  16 \big(4D + 4(D+1) \sqrt{D} + (D+1)^2\big) \Big[ (1+D^2) (1-2\qi) (1-2\qj) \\
   & - D (8\qi\qj - 4\qi - 4\qj) \Big] + 16 (D-1)^2 \big(2D + (D+1)\sqrt{D} \big) \\
   = & \frac{\partial^2 \Lp}{\partial \qj \partial \qi} \\
   \frac{\partial^2 \Lp}{\partial \qj^2} = & -32 \big(4D + 4(D+1) \sqrt{D} + (D+1)^2\big) \Big[ (1+D^2) \qi (1-\qi) \\
   & + D (2\qi^2-2\qi+1) \Big]
   \end{split}
 \end{align*}
Evaluating them in in the center point yields
\begin{flalign}
   \frac{\partial^2 \Lp}{\partial \qi^2} (0.5,0.5) = & -8 \big(4D + 4(D+1) \sqrt{D} + (D+1)^2\big) (D+1)^2, \label{eq:d2_Lp_dqi2} \\
   \frac{\partial^2 \Lp}{\partial \qi \partial \qj} (0.5,0.5) = & \, \, 32 D \big(4D + 4(D+1) \sqrt{D} + (D+1)^2\big) \label{eq:d2_Lp_dqi_qj} \\
   & + 16 (D-1)^2 \big(2D + (D+1)\sqrt{D} \big) \nonumber \\
   = & \frac{\partial^2 \Lp}{\partial \qj \partial \qi} (0.5,0.5), \quad \text{and} \label{eq:d2_Lp_dqj_qi} \\
   \frac{\partial^2 \Lp}{\partial \qj^2} (0.5,0.5) = & -8 \big(4D + 4(D+1) \sqrt{D} + (D+1)^2\big) (D+1)^2, \label{eq:d2_Lp_dqj2}
 \end{flalign}
with the main diagonal entries~\eqref{eq:d2_Lp_dqi2} and~\eqref{eq:d2_Lp_dqj2} of the Hessian matrix being negative, and the off-diagonal entries~\eqref{eq:d2_Lp_dqi_qj} and~\eqref{eq:d2_Lp_dqj_qi} being positive. We aim to apply the diagonal dominance criterion (Sec.~\ref{subsec:mathematical},~\eqref{eq:diagonal_dominance}) to show that the negative of the Hessian is positive definite (and thus, the Hessian itself is negative definite). I.e., for the first row of the Hessian, we compute the difference between the magnitude of the diagonal entry~\eqref{eq:d2_Lp_dqi2} and the magnitude of the off-diagonal entry~\eqref{eq:d2_Lp_dqi_qj} as
\begin{align*}
 & |\frac{\partial^2 \Lp}{\partial \qi^2} (0.5,0.5) | - |\frac{\partial^2 \Lp}{\partial \qi \partial \qj} (0.5,0.5) | \\
 = \, & 8 \big(4D + 4(D+1) \sqrt{D} + (D+1)^2\big) (D+1)^2 \\
  & - 32 D \big(4D + 4(D+1) \sqrt{D} + (D+1)^2\big) - 16 (D-1)^2 \big(2D + (D+1)\sqrt{D} \big) \\
  = \, & 8(D-1)^2 \big(4D + 4(D+1) \sqrt{D} + (D+1)^2\big) - 16 (D-1)^2 \big(2D + (D+1)\sqrt{D} \big) \\
  = \, & 8(D-1)^2 \big[(D+1)^2 + 2(D+1)\sqrt{D}  \big]
\end{align*}
which positive.For the second row of the Hessian, we can perform the same calculation. As the negatives of the diagonal entries~\eqref{eq:d2_Lp_dqi2} and~\eqref{eq:d2_Lp_dqj2} are positive, the negative of the Hessian satisfies the diagonal dominance criterion and is thus positive definite. We conclude that the Hessian of $\Lp$ is negative definite in $(0.5,0.5)$, and that the center point is a local maximum of $\Lp$. \\

To analyze of which kind the two remaining stationary points $(\qi^{(1)},\qj^{(1)}=\qi^{(1)})$ and $(\qi^{(2)},\qj^{(2)}=\qi^{(2)})$ are, we could perform a similar calculation; however, we choose a more elegant approach in which we argue that they can only be saddle points or minima of $\Lp$ (but definitely no maxima). Concretely, we argue as follows: Suppose, without loss of generality\footnote{The argumentation regarding the symmetrical point $(\qi^{(2)},\qj^{(2)}=\qi^{(2)})$ is the same.}, that $(\qi^{(1)},\qj^{(1)}=\qi^{(1)})$ is a local maximum of $\Lp$. Then there must be some point $(\hat{q}, \hat{q})$ located on the line segment connecting the two maxima $(0.5,0.5)$ and $(\qi^{(1)},\qj^{(1)}=\qi^{(1)})$, such that $(\hat{q}, \hat{q})$ is the minimum of all points between these two maxima (otherwise they would not be maxima); more precisely, the directional derivative of $\Lp$ along $(1,1)^T$ is zero in $(\hat{q}, \hat{q})$. On the other hand, $(\hat{q}, \hat{q})$ is also a minimum or a maximum with respect to the direction that is orthogonal to the main diagonal (i.e., the line segment that connects $(0,0)$ and $(1,1)$); in other words, the directional derivative of $\Lp$ along $(1,-1)^T$ is zero in $(\hat{q}, \hat{q})$. This follows from the fact that $\Lp$ is a symmetric function across any point that lies on the main diagonal with respect to that orthogonal direction, because $\qi$ and $\qj$ can be exchanged in the definition~\eqref{eq:L_prime} of $\Lp$ without changing its value. Hence, there exist two directions that are orthogonal to each other, for which the directional derivatives in $(\hat{q}, \hat{q})$ are zero. Consequently, $(\hat{q}, \hat{q})$ is another stationary point of $\Lp$. However, we know that this is not possible and thus a contradiction to our assumption that $(\qi^{(1)},\qj^{(1)}=\qi^{(1)})$ is a maximum of $\Lp$. In summary, this shows that $(0.5,0.5)$ is the unique local maximum of $\Lp$ in $(0,1) \times (0,1)$. \\

It remains to show that $\Lp$ takes the value zero in $(0.5,0.5)$, and that it approaches negative limit values at the boundary. From this, we can then conclude that $\Lp$ is negative for all $(\qi,\qj) \neq (0.5,0.5)$, which implies that $(0.5,0.5)$ is the unique solution to our starting equation~\eqref{eq:HBijsubdet_final_equation}. Let us therefore evaluate $\Lp$ in $(0.5,0.5)$ (using its definition~\eqref{eq:L_prime}):
\begin{align*}
 \Lp(0.5,0.5) = & \, \frac{16}{16} \big(4D + 4(D+1) \sqrt{D} + (D+1)^2 \big) (1+D^2)- (D-1)^4 \\
 & -\frac{16}{8} D \big(4D + 4(D+1) \sqrt{D} + (D+1)^2 \big) \\
 & - \frac{8}{2} (D-1)^2 \big(2D + (D+1) \sqrt{D} \big) \\
 = & \,  (1-D)^2 \big(4D + (D+1)^2 - 8D \big) - (D-1)^4 \\
 = & \,  (1-D)^2 (D-1)^2 - (D-1)^4 = 0
\end{align*}
Finally, we compute the limit values of $\Lp$ at the boundary~\eqref{eq:boundary_of_domain} as
\begin{align*}
 \lim_{\substack{q_i \to 0 \\ q_j \to k }} \Lp(\qi,\qj)  = & -\underbrace{(D-1)^4}_{>0} - \underbrace{16D \big(4D + 4(D+1) \sqrt{D} + (D+1)^2 \big) k^2}_{\geq 0} \\
 & - \underbrace{8 (D-1)^2 \big(2D + (D+1) \sqrt{D} \big) k}_{\geq 0}  < 0,
\end{align*}
\begin{align*}
 \lim_{\substack{q_i \to 1 \\ q_j \to k }} \Lp(\qi,\qj)  = & -\underbrace{(D-1)^4}_{>0} - \underbrace{16D \big(4D + 4(D+1) \sqrt{D} + (D+1)^2 \big) (1-k)^2}_{\geq 0} \\
 & - \underbrace{8 (D-1)^2 \big(2D + (D+1) \sqrt{D} \big) (1-k)}_{\geq 0}  < 0,
\end{align*}
and analogously for $\lim_{\substack{q_i \to k \\ q_j \to 0 }} \Lp(\qi,\qj)$ and $\lim_{\substack{q_i \to k \\ q_j \to 1 }} \Lp(\qi,\qj)$ (where $0 \leq k \leq 1$). As described above, we have proven that $(0.5,0.5)$ is the unique solution to the modifed equation~\eqref{eq:HBijsubdet_final_equation}, and thus the only candidate for solving the original equation~\eqref{eq:HBijsubdet_starting_equation}. But, as we already know from Lemma~\ref{lemma:determinant_HBijsub_center}, $(0.5,0.5)$ is actually a solution to~\eqref{eq:HBijsubdet_starting_equation}, and consequently it must be its unique solution. \\

\underline{Part 3:} 
  As we already know from the first two parts, $\HBijsubdet$ has precisely one root in its domain which is the center point $(0.5,0.5)$. In this third and final part of the proof, we verify the second statement of Theorem~\ref{thm:HBijsubdet_at_betacrit}, namely that $\HBijsubdet$ is positive for all $(\qi,\qj) \neq (0.5,0.5)$. To this end, we first multiply $\HBijsubdet$ by the positive term $\qprod \cdot \Tij$ to obtain a slightly modified version of $\HBijsubdet$, that we define as
  \begin{align} \label{eq:modification_H_of_HBijsubdet}
   \hspace{-0.2cm} H (\qi,\qj) \coloneqq \qprod \Tij \HBijsubdet \stackrel{\eqref{eq:HBijsub_determinant},\eqref{eq:c1_c2_Q}}{=} \qprod  \Tij \big(\ci \frac{1}{\qprod} + \cii \frac{1}{\Tij}\big)= \ci \Tij + \cii \qprod.
  \end{align}
The form~\eqref{eq:modification_H_of_HBijsubdet} has the advantage that $\qprod$ and $\Tij$ occur in the numerators instead of the denominators. At the same time, this modification does not change the sign of $\HBijsubdet$, which is positive in a point $(\qi,\qj)$ if and only if $H(\qi,\qj)$ is positive in $(\qi,\qj)$. In other words, to prove that $\HBijsubdet$ is positive for all $(\qi,\qj) \neq (0.5,0.5)$, we may equivalently prove this statement for $H(\qi,\qj)$. By construction, $H(\qi,\qj)$ has a unique root in $(0.5,0.5)$, and therefore it is sufficient to prove that $H(\qi,\qj)$ is positive definite in $(0.5,0.5)$. This implies that $H(\qi,\qj)$ is positive on its entire domain (otherwise it would have multiple roots). We thus calculate its second-order partial derivatives with respect to $\qi$ and $\qj$ as
\begin{align}
 & \frac{\partial^2 H}{\partial \qi^2} = \ci \frac{\partial^2 \Tij}{\partial \qi^2} - 2\cii \qj(1-\qj), \label{eq:second_order_derivatives_HBijsubdet_qi2} \\
 & \frac{\partial^2 H}{\partial \qi \partial \qi} = \ci \frac{\partial^2 \Tij}{\partial \qi \partial \qj} + \cii (1-2\qi) (1-2\qj), \label{eq:second_order_derivatives_HBijsubdet_qi_qj} \\
 & \frac{\partial^2 H}{\partial \qj^2} = \ci \frac{\partial^2 \Tij}{\partial \qj^2} - 2\cii \qi(1-\qi), \label{eq:second_order_derivatives_HBijsubdet_qj2}
\end{align}
with the second-order derivatives of $\Tij$ calculated in Lemma~\ref{lemma:partial_derivatives_xij_Tij_rijp}, (b). To evaluate~\eqref{eq:second_order_derivatives_HBijsubdet_qi2} -~\eqref{eq:second_order_derivatives_HBijsubdet_qj2} in $(0.5,0.5)$, we first need to evaluate the second-order derivatives of $\Tij$ and all involved sub-functions in $(0.5,0.5)$. From Lemma 1 in~\citep{leisenberger2022fixing}, we know the value of $\xijopt$ in the center point which is
\begin{align} \label{eq:xij_center}
 \xijopt(0.5,0.5) = \frac{\sigma(2\beta\Jij)}{2}.
 \end{align}
 Here we have used the logistic sigmoid function which is defined as
 \begin{align} \label{eq:logistic_sigmoid}
 \sigma(2\beta\Jij) \coloneqq \frac{1}{1+e^{-2\beta\Jij}}.
 \end{align}
 Earlier in this paper (in equation~\eqref{eq:Tij_center}) we have already provided the value for $\Tij$ in the center point (also as a result from~\citep{leisenberger2022fixing}) which is given by
\begin{align} \label{Tij_center_once_again}
 \Tij(0.5,0.5) = \frac{1}{16 \cosh^2(\beta \Jij)}.
 \end{align}
Next, we evaluate the first-order derivatives of $\xijopt$ and $\Tij$ in $(0.5,0.5)$, that have been derived in Lemma~\eqref{lemma:partial_derivatives_xij_Tij_rijp}, (a). Substituting~\eqref{eq:xij_center} and~\eqref{Tij_center_once_again} into the corresponding equations~\eqref{eq:deri_1_xij_qi} -~\eqref{eq:deri_1_Tij_qj} yields
\begin{align} \label{eq:first_order_derivatives_xij_center}
 \frac{\partial \xijopt}{\partial q_{i}}(0.5,0.5) = \frac{\partial \xijopt}{\partial q_{j}}(0.5,0.5) = \frac{\aij\big(\frac{1}{2}-\frac{\sigma(2\beta\Jij)}{2} \big) + 0.5}{1+\aij(1-\sigma(2\beta\Jij))}  \stackrel{\eqref{eq:logistic_sigmoid}}{=} \frac{1}{2}
\end{align}
and
\begin{align} \label{eq:first_order_derivatives_Tij_center}
 \frac{\partial \Tij}{\partial q_{i}}(0.5,0.5) = \frac{\partial \Tij}{\partial q_{j}}(0.5,0.5) = -2\big( \frac{\sigma(2\beta\Jij)}{2} - \frac{1}{4}\big)(\frac{1}{2} - \frac{1}{2}) = 0.
\end{align}
Subsequently, we evaluate the second-order derivatives of $\xijopt$ and $\Tij$ in $(0.5,0.5)$, that have been derived in Lemma~\eqref{lemma:partial_derivatives_xij_Tij_rijp}, (b). We substitute~\eqref{eq:xij_center} -~\eqref{eq:first_order_derivatives_Tij_center} into the corresponding equations~\eqref{eq:deri_2_xij_qi2} -~\eqref{eq:deri_2_Tij_qj2} and, using the definition~\eqref{eq:xi_optimal} of $\aij=e^{4\beta\Jij}-1$ and the trigonometrical identities
\begin{align} \label{eq:sinh_cosh_identities}
\begin{split}
 & \sinh(2\beta\Jij) = \frac{1}{2}(e^{2\beta\Jij}-e^{2\beta\Jij}),  \\
 & \cosh(2\beta\Jij) = \frac{1}{2} (e^{2\beta\Jij}+e^{2\beta\Jij}), \quad \text{and} \\
 & \tanh(2\beta\Jij) = \frac{\sinh(2\beta\Jij)}{\cosh(2\beta\Jij)},
 \end{split}
\end{align}
obtain
\begin{flalign} 
& \frac{\partial^2 \xijopt}{\partial \qi^2}(0.5,0.5) = \frac{\partial^2 \xijopt}{\partial \qi^2}(0.5,0.5) \nonumber \\
& = \frac{\aij(-\frac{1}{2}) \big(1+\aij(1-\sigma(2\beta\Jij))\big)}{\big(1+\aij(1-\sigma(2\beta\Jij))\big)^2} =- \sinh(2\beta\Jij), \label{eq:second_order_derivatives_xij_center_qi2} \\
& \frac{\partial^2 \xijopt}{\partial \qi \partial \qj}(0.5,0.5) = \frac{\partial^2 \xijopt}{\partial \qj \partial \qi}(0.5,0.5) \nonumber \\
& = \frac{(\frac{\aij}{2}+1)\big(1+\aij(1-\sigma(2\beta\Jij))\big)}{\big(1+\aij(1-\sigma(2\beta\Jij))\big)^2} = \cosh(2\beta\Jij), \label{eq:second_order_derivatives_xij_center_qi_qj}
\end{flalign}
and
\begin{flalign} 
\hspace{-0.5cm} & \frac{\partial^2 \Tij}{\partial \qi^2}(0.5,0.5) = \frac{\partial^2 \Tij}{\partial \qi^2}(0.5,0.5) \nonumber \\
\hspace{-0.5cm}  & = -\frac{1}{2} + 2\Big( \big(\frac{\sigma(2\beta\Jij)}{2} - \frac{1}{4} \big) (\sinh(2\beta\Jij)) \Big) = \frac{1}{2} \big(\cosh(2\beta\Jij)-2\big), \label{eq:second_order_derivatives_Tij_center_qi2} \\
\hspace{-0.5cm}  & \frac{\partial^2 \Tij}{\partial \qi \partial \qj}(0.5,0.5) = \frac{\partial^2 \Tij}{\partial \qj \partial \qi}(0.5,0.5) \nonumber \\
\hspace{-0.5cm}  & = -2 \Big(\big(\frac{\sigma(2\beta\Jij)}{2} - \frac{1}{4} \big) \big(\cosh(2\beta \Jij) -1 \big)   \Big)= -\sinh^2(\beta\Jij) \tanh(\beta\Jij). \label{eq:second_order_derivatives_Tij_center_qi_qj}
\end{flalign}
With the prepared second-order derivatives of $\Tij$, we are now ready to compute the Hessian matrix of the function $H(\qi,\qj)$ (defined in~\eqref{eq:modification_H_of_HBijsubdet}) in the center point. We thus insert the expressions from~\eqref{eq:second_order_derivatives_Tij_center_qi2} and~\eqref{eq:second_order_derivatives_Tij_center_qi_qj} into the equations for the second-order derivatives of $H(\qi,\qj)$ (computed in~\eqref{eq:second_order_derivatives_HBijsubdet_qi2} -~\eqref{eq:second_order_derivatives_HBijsubdet_qj2}) and get
\begin{align}
 & \frac{\partial^2 H}{\partial \qi^2}(0.5,0.5) = \frac{\ci}{2} \big(\cosh(2\beta\Jij)-2\big) - \frac{\cii}{2}, \label{eq:second_order_derivatives_HBijsubdet_qi2_center} \\
 & \frac{\partial^2 H}{\partial \qi \partial \qi}(0.5,0.5) = \ci \big(-\sinh^2(\beta\Jij) \tanh(\beta\Jij)\big), \quad \text{and} \label{eq:second_order_derivatives_HBijsubdet_qi_qj_center} \\
 & \frac{\partial^2 H}{\partial \qj^2}(0.5,0.5) = \frac{\ci}{2} \big(\cosh(2\beta\Jij)-2\big) - \frac{\cii}{2}. \label{eq:second_order_derivatives_HBijsubdet_qj2_center}
\end{align}
Now recall that we are considering $\HBijsubdet$ (and thus $H(\qi,\qj)$) for $\beta=\betacrit$, i.e., at the critical inverse temperature. We thus substitute $\betacrit$ (from~\eqref{eq:critical_beta} and with $D$ defined in~\eqref{eq:definition_of_D}) into~\eqref{eq:second_order_derivatives_HBijsubdet_qi2_center} -~\eqref{eq:second_order_derivatives_HBijsubdet_qj2_center}:
\begin{flalign}
 & \frac{\partial^2 H}{\partial \qi^2}(0.5,0.5) = \frac{\partial^2 H}{\partial \qj^2}(0.5,0.5)= \frac{1}{2} \big(\ci \frac{2}{D-1} -2 -\cii \big) \label{eq:second_order_derivatives_HBijsubdet_qi2_center_critial_beta} \\
 & \frac{\partial^2 H}{\partial \qi \partial \qj }(0.5,0.5) =  \frac{\partial^2 H}{\partial \qj \partial \qi}(0.5,0.5) = - \frac{\ci}{D-1} \frac{1}{\sqrt{D}} \label{eq:second_order_derivatives_HBijsubdet_qi_qj_center_critical_beta}
\end{flalign}
In the computation of~\eqref{eq:second_order_derivatives_HBijsubdet_qi_qj_center_critical_beta}, we have used the trigonometrical identities
\begin{align*}
 & \sinh(\betacrit \Jij) = \sinh\big(\frac{1}{2}\arccosh (1 + \frac{2}{D-1})\big) = \sqrt{\frac{1 + \frac{2}{D-1} - 1}{2}}  = \frac{1}{\sqrt{D-1}}, \\
 & \tanh(\betacrit \Jij) = \tanh\big(\frac{1}{2}\arccosh (1 + \frac{2}{D-1})\big) = \sqrt{\frac{1+\frac{2}{D-1}-1}{1+\frac{2}{D-1}+1}} = \frac{1}{\sqrt{D}}.
\end{align*}
To finally prove that the Hessian of $H(\qi,\qj)$ is positive definite in $(0.5,0.5)$, we apply the diagonal dominance criterion (Sec.~\ref{subsec:mathematical},~\eqref{eq:diagonal_dominance}). Note that the main diagonal entries~\eqref{eq:second_order_derivatives_HBijsubdet_qi2_center_critial_beta} are positive (because $|\ci| > |\cii|$, with $\ci,\cii$ being defined in~\eqref{eq:c1_c2_Q}) and the off-diagonal entries~\eqref{eq:second_order_derivatives_HBijsubdet_qi_qj_center_critical_beta} are negative (because $\ci > 0$). For both rows of the Hessian, the difference between the magnitude of the main diagonal entry and the magnitude of the off-diagonal entry is
\begin{align}
 \frac{1}{2} \big(\ci \frac{2}{D-1} -2 -\cii \big) - \frac{\ci}{D-1} \frac{1}{\sqrt{D}} = \frac{1 - \frac{1}{\sqrt{D}}}{2 \degi \degj (1 + \frac{1}{\sqrt{D}})} >0.
\end{align}
Hence, the Hessian of $H(\qi,\qj)$ is diagonally dominant and therefore positive definite in $(0.5,0.5)$. Then $H(\qi,\qj) > 0$ for all $(\qi,\qj) \neq (0.5,0.5)$ (as $(0.5,0.5)$ is its only root) and consequently this is also true for $\HBijsubdet(\qi,\qj)$. This completes the final part and thus the proof of Theorem~\ref{thm:HBijsubdet_at_betacrit}. \hfill\BlackBox
\\

\section{Algorithm \texttt{BETHE-MIN}} \label{sec:appendix_C}

Algorithm~\ref{algo:projected_quasi_Newton} describes our proposed method \texttt{BETHE-MIN} for minimizing the Bethe free energy that we have applied in our experiments in Sec.~\ref{subsec:experiments_ferro_general}. It requires an initialization $\qvec^{(0)}$ of the parameter vector which contains estimates of the singleton marginals. Until convergence, \texttt{BETHE-MIN} iterates through the following steps: Update of the search direction according a quasi-Newton scheme, i.e., with the inverse Hessian being replaced by an approximation; projecting the largest possible step back into $\Bbox$; optimizing the step size via an adaptive Wolfe line search (\texttt{WOLFE-LS}, described in Algorithm~\ref{algo:line_search_Wolfe}); updating the parameter vector according to~\eqref{eq:quasi_newton_update}; and updating the current approximation to the inverse Hessian (we use the BFGS update rules for approximating the inverse Hessian,~\citet{nocedal2006numerical}). For checking convergence, we compute the gradient norm with respect to the current parameter vector $\qvec^{(t)}$. After \texttt{BETHE-MIN} has converged, it returns a stationary point $\qvec^{\ast}$ (usually a minimum) of $\FB$. \\

\begin{algorithm}
\begin{algorithmic}
\caption{\texttt{BETHE-MIN} (Projected quasi-Newton for minimizing $\FB$)} \label{algo:projected_quasi_Newton}
\Require Parameter vector $\qvec^{(0)}$ in $\Bbox$
\Ensure Stationary point $\qvec^{\ast}$ of $\FB$
\State $\qvec^{(t)} \gets \qvec^{(0)}$
\State Initialize $\textbf{B}^{(t)}$ randomly as a matrix of size $|\setofnodes|\times|\setofnodes|$
\While{$\lVert \nabla \FB\big( \qvec^{(t)} \big)\rVert > \epsilon$} \Comment{Check for convergence}
\State $\qvec^{\text{old}} \gets \qvec^{(t)}$ \Comment{Store current parameter vector}
\State $\bm{d}^{(t)} \gets - \big(\textbf{B}^{(t)} \big)^T \, \nabla \FB\big(\qvec^{(t)} \big) $ \Comment{Update search direction}
\State $\qvec^{\pi} \gets \qvec^{(t)} + \bm{d}^{(t)}$
\While{$\qvec^{\pi} \notin \Bbox$} \Comment{Project $\qvec^{\pi}$ back into $\Bbox$}
\State $\rho^{\max} \gets 0.9 \cdot \rho^{\max}$
\State $\qvec^{\pi} \gets \qvec^{(t)} + \rho^{\max} \cdot \bm{d}^{(t)}$
\EndWhile
\State $\rho^{(t)} \gets \texttt{WOLFE-LS}(\qvec^{(t)},\qvec^{\pi})$ \Comment{Compute step size via Algorithm~\ref{algo:line_search_Wolfe}}
\State $\qvec^{(t)} \gets \qvec^{(t)} + \rho^{(t)} \cdot (\qvec^{\pi} - \qvec^{(t)})$ \Comment{Update parameter vector}
\State $\bm{s}^{(t)} \gets \qvec^{(t)} - \qvec^{\text{old}}$ 
\State $\bm{y}^{(t)} \gets \nabla \FB\big(\qvec^{(t)} \big) - \nabla \FB\big(\qvec^{\text{old}} \big)$ \Comment{BFGS update rules}
\State $\gamma^{(t)} \gets ( \bm{s}^{(t)})^T \bm{y}^{(t)}$ \Comment{\hspace{0.08cm} for approximating}
\State $\textbf{B}^{(t)} \gets \textbf{B}^{(t)} + \frac{1}{\big(\gamma^{(t)}\big)^2} \big(\gamma^{(t)} + (\bm{y}^{(t)})^T \textbf{B}^{(t)} \bm{y}^{(t)} \big)$ \Comment{\hspace{-0.12cm} the inverse Hessian} \\
\hspace{1.8cm} $+ \frac{1}{\gamma^{(t)}} \big(\textbf{B}^{(t)} \bm{y}^{(t)} ( \bm{s}^{(t)})^T + \bm{s}^{(t)} (\bm{y}^{(t)})^T \textbf{B}^{(t)} \big)$
\EndWhile
\State \Return $\qvec^{(t)}$
\end{algorithmic}
\end{algorithm}

The Wolfe conditions used in the adaptive line search stategy are as follows:

\begin{align}
  & \hspace{-0.3cm} \FB\big(\qvec^{\text{tail}} + \rho^{\text{W}} \cdot (\qvec^{\text{head}} - \qvec^{\text{tail}})\big) \, \leq \, \FB\big(\qvec^{\text{tail}}\big) + \tau_1 \cdot \rho^{\text{W}} \cdot \bm{d}^T \, \nabla \FB\big(\qvec^{\text{tail}}\big) \tag{W1} \label{eq:Wolfe-Powell_1} \\
  & \hspace{-0.3cm} \bm{d}^T \, \nabla \FB\big(\qvec^{\text{tail}} + \rho^{\text{W}} \cdot (\qvec^{\text{head}} - \qvec^{\text{tail}} )\big) \, \geq \, \tau_2 \cdot \bm{d}^T \, \nabla \FB\big(\qvec^{\text{tail}} \big) \tag{W2} \label{eq:Wolfe-Powell_2}
\end{align}

The parameter vectors $\qvec^{\text{tail}}$  and $\qvec^{\text{head}}$ specify the search interval. Condition~\eqref{eq:Wolfe-Powell_1} ensures that there is a sufficient decrease of the energy function after each iteration. This can be achieved by sufficiently reducing the step size $\rho^{\text{W}}$ to prevent $\FB$ from increasing again (which may happen if the step is too large). Conversely, condition~\eqref{eq:Wolfe-Powell_2} ensures that we do not stop moving along the search direction $\bm{d}$, as long as the energy function descends sufficiently steeply. Hence, the step size must not be chosen too small. Both conditions together guarantee that the step size is neither chosen too great nor too small. The parameters $\tau_1$ and $\tau_2$ control how 'strict' the individual conditions are. Following a recommendation of~\citet{nocedal2006numerical}, we have selected numerical values of $\tau_1=10^{-4}$ and $\tau_2=0.9$ for our experiments in Sec.~\ref{subsec:experiments_ferro_general}. The detailed procedure of computing a step size satisfying the Wolfe conditions is described in Algorithm~\ref{algo:line_search_Wolfe}. It is used as subroutine in Algorithm~\ref{algo:projected_quasi_Newton}.

\begin{algorithm}
\begin{algorithmic}
 \caption{\texttt{WOLFE-LS} (Adaptive line search with Wolfe conditions)} \label{algo:line_search_Wolfe}
\Require Endpoints $\qvec^{\text{tail}}$, $\qvec^{\text{head}}$ of search interval
\Ensure Step size $\rho^\text{W}$ satisfying the Wolfe conditions~\eqref{eq:Wolfe-Powell_1} and~\eqref{eq:Wolfe-Powell_2}
\State $\bm{d} \gets \qvec^{\text{head}} - \qvec^{\text{tail}}$ \Comment{Search direction}
\State Initialize step size $\rho^\text{W}$ randomly between zero and one
\While{\eqref{eq:Wolfe-Powell_1} is satisfied and~\eqref{eq:Wolfe-Powell_2} is not satisfied}
\State $\rho^\text{W} = 1.1 \cdot \rho^\text{W}$ \Comment{Expand interval of potential step sizes}
\EndWhile
\If{\eqref{eq:Wolfe-Powell_1} and~\eqref{eq:Wolfe-Powell_2} are satisfied}
\State \Return $\rho^{\text{W}}$ \Comment{STOP. Feasible step size has been found }
\Else
\State $l \gets 0$ \Comment{Define new search interval for feasible step size}
\State $r \gets \rho^\text{W}$ \Comment{Upper bound of new search interval}
\While{\eqref{eq:Wolfe-Powell_1} or~\eqref{eq:Wolfe-Powell_2} is not satisfied} \Comment{Contraction phase}
\State Pick $\rho^\text{W}$ randomly from interval $(l,r)$
\If{\eqref{eq:Wolfe-Powell_1} is not satisfied}
\State $r \gets \rho^\text{W}$ \Comment{Upper bound of search interval is reduced}
\Else
\State $l \gets \rho^\text{W}$ \Comment{Lower bound of search interval is increased}
\EndIf
\EndWhile
\EndIf
\State \Return $\rho^\text{W}$
\end{algorithmic}
\end{algorithm}





\newpage


\vskip 0.2in
\bibliography{JMLR2024}

\end{document}